\documentclass[pdflatex,sn-mathphys-num]{sn-jnl}

\usepackage{graphicx}%
\usepackage{multirow}%
\usepackage{amsmath,amssymb,amsfonts}%
\usepackage{amsthm}%
\usepackage{mathrsfs}%
\usepackage[title]{appendix}%
\usepackage{xcolor}%
\usepackage{textcomp}%
\usepackage{manyfoot}%
\usepackage{booktabs}%
\usepackage{tabularx}%
\usepackage{longtable}
\usepackage{vcell}
\usepackage{array}
\usepackage{ragged2e}
\usepackage{pdflscape}
\usepackage{algorithm}%
\usepackage{algorithmicx}%
\usepackage{algpseudocode}%
\usepackage{listings}%
\usepackage{soul}
\usepackage{awesomebox}
\usepackage{subcaption}
\usepackage{rotating}

\usepackage{multirow}
\usepackage{multicol}
\usepackage{enumitem}
\usepackage{siunitx} 
\usepackage{comment}

\usepackage{caption}
\theoremstyle{thmstyleone}%
\theoremstyle{thmstyletwo}%

\theoremstyle{thmstylethree}%

\begin{document}

\title[Where can AI be used?]{%
\centering
Where can AI be used?
\\Insights from a deep ontology of work activities}


\author[1,2]{\fnm{Alice} \sur{Cai}}\email{cai@mit.edu}
\equalcont{These authors contributed equally to this work.}

\author[1,2]{\fnm{Iman} \sur{YeckehZaare}}\email{oneman@mit.edu}
\equalcont{These authors contributed equally to this work.}

\author[1,3]{\fnm{Shuo} \sur{Sun}}\email{shuo.sun@smart.mit.edu}
\equalcont{These authors contributed equally to this work.}

\author[1,3]{\fnm{Vasiliki} \sur{Charisi}}\email{vasiliki.charisi@smart.mit.edu}
\equalcont{These authors contributed equally to this work.}

\author[1,3]{\fnm{Xinru} \sur{Wang}}\email{xinru.wang@smart.mit.edu}

\author[1,2]{\fnm{Aiman} \sur{Imran}}\email{aimanim@mit.edu}

\author[1,2]{\fnm{Robert} \sur{Laubacher}}\email{rjl@mit.edu}

\author[1,3]{\fnm{Alok} \sur{Prakash}}\email{alok.prakash@smart.mit.edu}

\author*[1,2]{\fnm{Thomas W.} \sur{Malone}}\email{malone@mit.edu}

\affil*[1]{\orgdiv{Center for Collective Intelligence}, \orgname{Massachusetts Institute of Technology}, \orgaddress{\street{100 Main Street}, \city{Cambridge}, \postcode{02142}, \state{Massachusetts}, \country{USA}}}

\affil[2]{\orgdiv{Sloan School of Management}, \orgname{Massachusetts Institute of Technology}, \orgaddress{\street{100 Main Street}, \city{Cambridge}, \postcode{02142}, \state{Massachusetts}, \country{USA}}}

\affil*[3]{\orgdiv{Mens, Manus, and Machina}, \orgname{Singapore-MIT Alliance for Research and Technology}, \orgaddress{\street{1 CREATE Way}, \city{Singapore}, \postcode{138602}, \country{Singapore}}}


\abstract{
Artificial intelligence (AI) is poised to profoundly reshape how work is executed and organized, but we do not yet have deep frameworks for understanding where AI can be used. Here we provide a comprehensive ontology of work activities that can help systematically analyze and predict uses of AI. To do this, we disaggregate and then substantially reorganize the approximately 20K activities in the US Department of Labor’s widely used O*NET occupational database. Next, we use this framework to classify descriptions of 13,275 AI software applications and a worldwide tally of 20.8 million robotic systems. Finally, we use the data about both these kinds of AI to generate graphical displays of how the estimated units and market values of all worldwide AI systems used today are distributed across the work activities that these systems help perform. We find a highly uneven distribution of AI market value across activities, with the top 1.6\% of activities accounting for over 60\% of AI market value. Most of the market value is used in information-based activities (72\%), especially creating information (36\%), and only 12\% is used in physical activities. Interactive activities include both information-based and physical activities and account for 48\% of AI market value, much of which (26\%) involves transferring information. These results can be viewed as rough predictions of the \textit{AI applicability} for all the different work activities down to very low levels of detail. Thus, we believe this systematic framework can help predict at a detailed level where today's AI systems can and cannot be used and how future AI capabilities may change this.

}




\maketitle

\section{Introduction}





Many people today believe that AI will lead to vast changes\textemdash either positive or negative\textemdash in the ways work is done. But to think systematically about these issues, one of the most basic questions is: \textit{Where can AI be used?} Where, for example, is it both technically and economically feasible to use AI for performing specific work tasks today? And how might these answers change in the future?

A common way for researchers to answer questions like these is by systematically identifying work activities and then considering which of these activities can be done by AI (see \cite{autor_skill_2003,autor_task_2013,Frey_and_Osborne_2013,brynjolfsson_what_2018} and Appendix \ref{sec:related_work}). For example, much of this research uses the work activities identified in O*NET \cite{ONET_Resource_Center_2016}, the U.S. Department of Labor’s comprehensive occupational database. This database contains detailed descriptions of approximately 20{,}000 activities from over 900 occupations in the U.S. economy.


Although O*NET is very comprehensive and has been widely used to help job seekers find suitable jobs, it has limitations as a conceptual tool for analyzing where AI can be used. For example, due to the compound nature of its basic tasks, it is often difficult to determine which element of a compound task is relevant in a specific situation. 



To help address these problems, we created an early version of a \textit{deep ontology of work activities}. By \textit{work activities}, we mean activities like those commonly performed by human workers, and now, increasingly, also by computers, robots, and other types of machines. By \textit{ontology}, we mean a formal specification of the key types of entities in a subject area and their properties and relationships (see \cite{gruber_translation_1993,guarino_formal_1998,noauthor_ontology_2025,smith_ontology_2003} and Section~\ref{sec:background:ontology}). And by \textit{deep}, we mean that the structure of the ontology includes multi-leveled ``family trees" of entities that have deep similarities with each other. 

To create an ontology with these properties, we started with all the O*NET tasks, disaggregated them into what we call \textit{atomic activities}, and created more abstract \textit{generic activities} as needed to arrange the atomic activities into family trees of similar activities. Our current version of this ontology contains 39,603 total activities, including 1,113 generic activities, 15,989 atomic activities, and 20,950 original O*NET tasks. 

To use this ontology as a framework for understanding where AI can be used, we acquired two large datasets about AI uses: (1) a database with descriptions of 13,275 AI software applications collected on a website called ``There’s an AI for That\textsuperscript{®}'' (TAAFT)~\cite{nedelcu_theres_2026}, and (2) data on 20.8 million robots of different types as reported by the International Federation of Robotics (IFR)~\cite{IFR2025Service, IFR2025Industrial}.


We then classified all of these AI systems into the activities in our ontology that they performed or helped perform. Next, we created graphical displays of the intensity of AI usage in each activity. We interpreted this intensity of usage as an indicator of the \textit{AI applicability} for each activity. And, finally, using the concept of \textit{inheritance} that is common in ontologies, we assumed that the AI applicability of an activity was often ``inherited" by its more specialized descendants in the same branch of the family tree, potentially all the way down to very detailed activities. 

This, then, provides a systematic way of predicting\textemdash at a detailed level\textemdash where AI can be used. The data we already have in the ontology can be used to help predict uses that are possible today. And by hypothesizing future changes in AI applicability for abstract activities, the ontology can also be used to predict possible AI uses in the future.

\section{Results}
\label{sec:result}

\subsection{A deep ontology of work activities}
\label{sec:results:ontology}

\begin{figure}[htbp]
    \centering
    \includegraphics[width=1\textwidth]{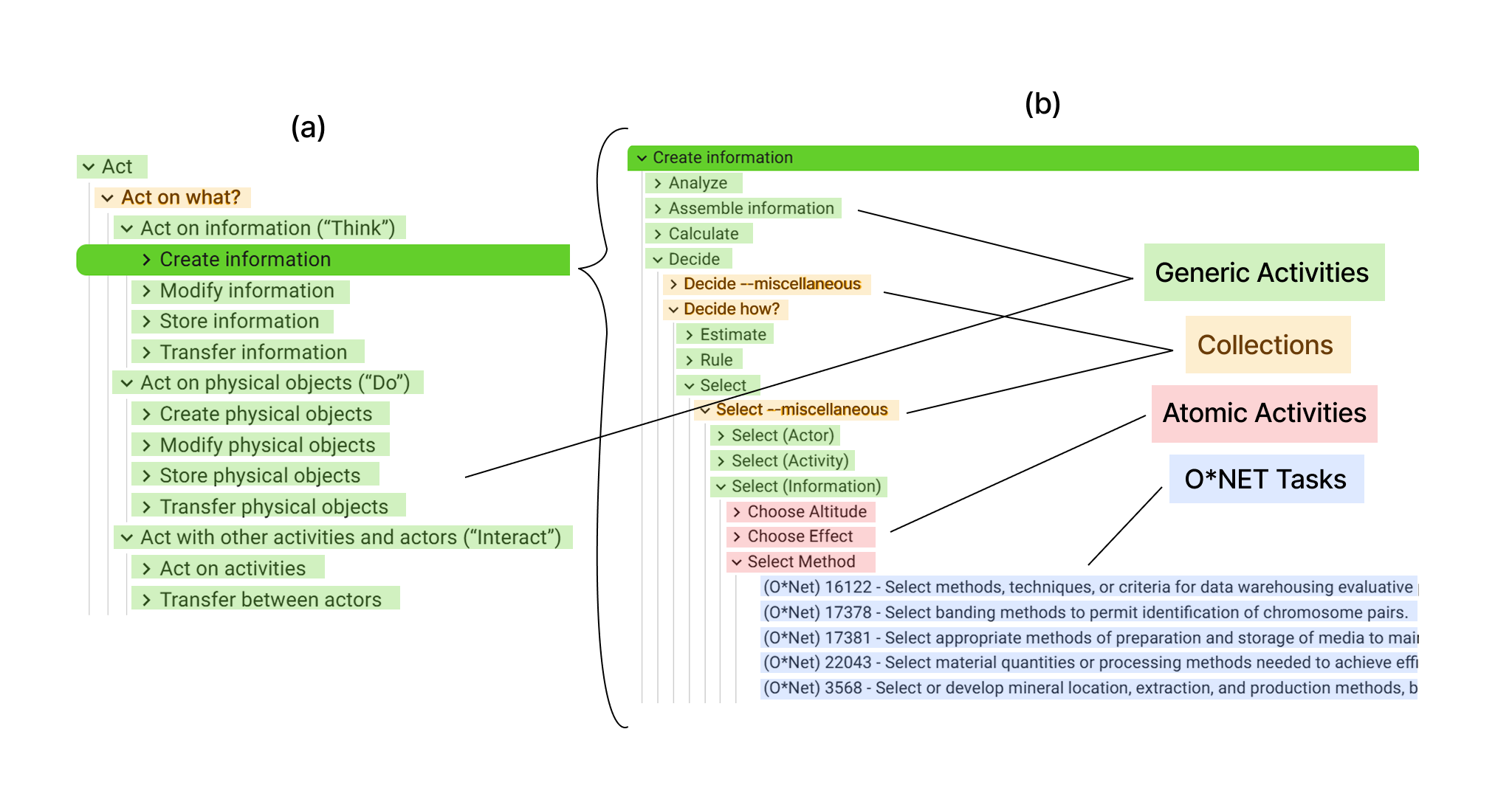}
    \caption{
        \textbf{Excerpt of the ontology structure.}
        (a) The top 4 levels of the ontology, (b) a sample of the next 5 levels beneath ``Create information,'' including generic activities, such as ``Decide" and ``Select," and leading all the way down to atomic activities, such as ``Select method." At the last level are O*NET tasks, such as ``Select banding methods to permit identification of chromosome pairs." Collections are groupings of specializations that often differ on a dimension such as ``Decide how?" or ``Decide what?"
    }
    \label{fig-ontologyexcerpt}
\end{figure}

Using the methods for automated classification and human editing described in Sections \ref{sec:background:ontology} to \ref{sec:method:platform}, our first result was a comprehensive ontology of work activities including approximately 40,000 total activities. These activities are arranged in a hierarchy with a median depth of 9 levels, and \autoref{fig-ontologyexcerpt} shows a sample of how the different types of activities are organized. 

At the far left is the most general type of activity of all, ``Act,'' followed by a \textit{collection} of more specialized (``child") activities that are all answers to the question ``Act on what?''  These three generic activities and their nicknames are: ``Act on information (`Think'),'' ``Act on physical objects (`Do'),'' and ``Act with other activities or actors (`Interact').''  

Indented beneath these three major categories of activities are further specializations (``children") of each. For example, ``Act on information'' has children such as ``Create information,'' ``Modify information,'' and ``Transfer information.'' To the right is an expansion of one of these activities, ``Create information,'' showing further children and collections. At the next to last level of the hierarchy are the \textit{atomic activities}, which include both verbs and objects in their names. Finally, at the far right are the compound O*NET tasks in which these atomic activities are included. The O*NET tasks, however, are not \textit{children} of the atomic activities; they are just the places in O*NET where the atomic activities are used.


\paragraph{Inheritance of AI applicability}

As noted above, our ontology takes advantage of the concept of inheritance. This concept formalizes the common sense notion that if something is true of one kind of object, it is usually true of more specialized types of that kind of object. For example, the things that are true of mammals are true of dogs, and the things that are true of dogs are true of Dalmatian puppies. 

\begin{figure}[h]
    \centering
    \includegraphics[width=1\textwidth]{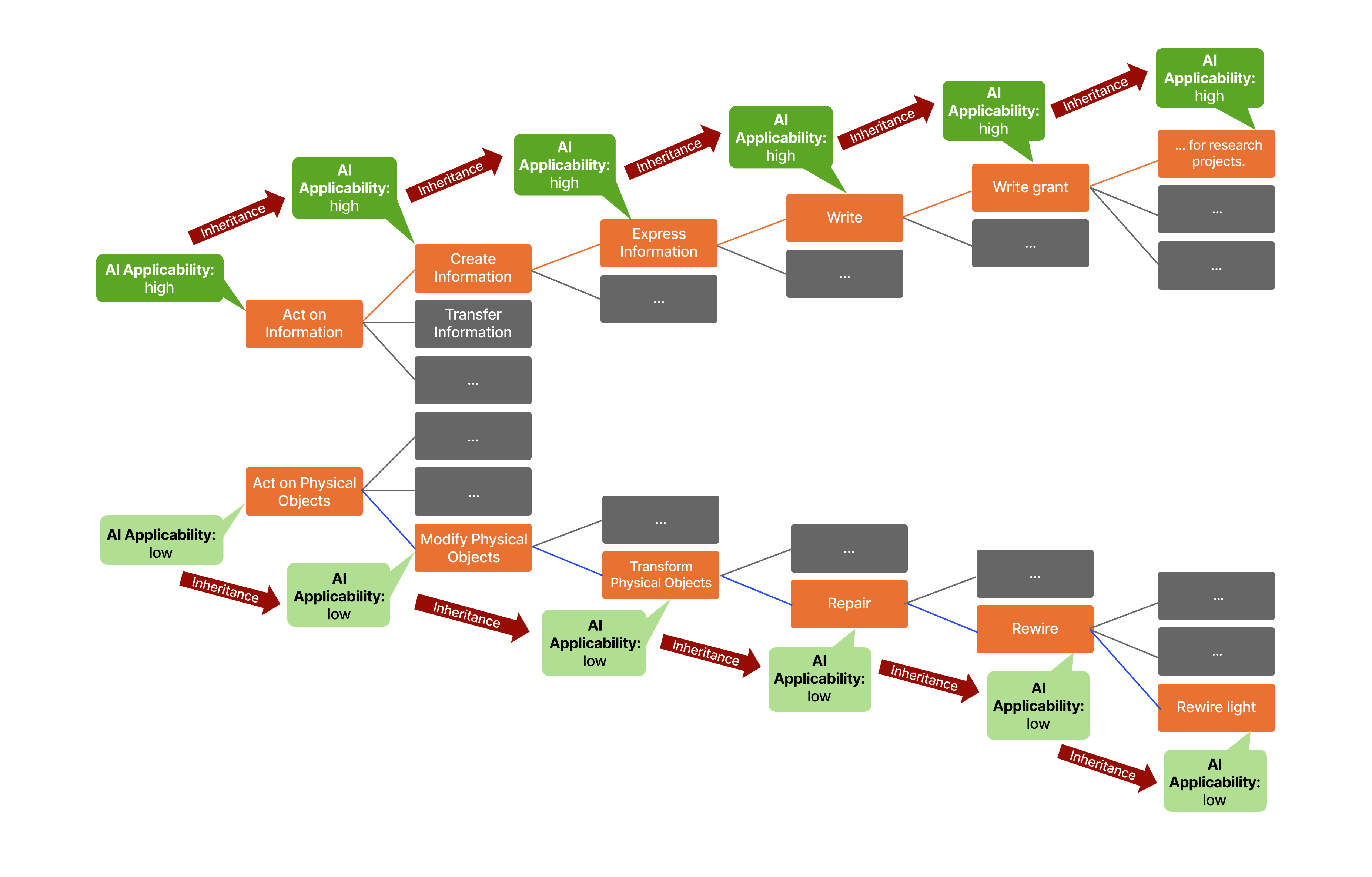}
    \caption{
        \textbf{Demonstration of \textit{inheritance} of the property \textit{AI Applicability}.} This stylized example illustrates how different values of the property of ``AI applicability" can be inherited down different branches of the ontology from very abstract types of activity like ``Create information" to very specific activities like ``Write grant application for research projects." 
    }
    \label{fig-inheritance}
\end{figure}

In our case, we focus on inheriting the property of \textit{AI applicability}, that is, whether an activity can be performed using AI. And we hypothesize that if an activity can be performed using AI, then the more specialized types of that activity (its ``descendants") can probably also be performed using AI. \autoref{fig-inheritance} illustrates this principle with a stylized example in which AI applicability is shown as high for the descendants of ``Act on information" and low for the descendants of ``Act on physical objects."

This means that if we discover that AI is applicable to a parent activity (e.g., ``Write''), we have principled grounds for hypothesizing that it may be applicable to the child activities (e.g., ``Write report,'' ``Write email'') even before verifying this empirically. 

Of course, this is not always true, and the concept of inheritance also allows the value of a property to be \textit{overridden} at any point in the hierarchy. In this case, the new value that overrides an inherited value is inherited by further descendants below that point. For instance, even though most mammals reproduce by live birth, duck-billed platypuses lay eggs. In the same way, even though AI is highly applicable to the activity of writing, there are some types of writing (such as writing deeply personal love letters) where it would be much less applicable. 


The concept of inheritance also means that one way to estimate the AI applicability of an abstract activity is to assume that if most of its descendants have high AI applicability, then it probably does, too. In most of the results below, we take advantage of this assumption by estimating the AI applicability of an activity as the sum of both (a) the \textit{direct} uses of AI systems in performing the activity, and (b) the \textit{aggregated} uses of AI in all the descendants of the activity. 


\subsection{AI software applications}
\label{sec:taaft}

To use this ontology in understanding where AI can be used, we first obtained a database with descriptions of 13,275 AI software applications collected on a website called ``There’s an AI for That\textsuperscript{®}'' (TAAFT)~\cite{nedelcu_theres_2026}. These applications span a wide range of tasks, from text generation and data analysis to image generation and workflows (see Section \ref{ai_software_applications}). Then, using an automated classification pipeline (Section \ref{sec:method:pipeline:automated}), we classified each application into the most specific ontology activity that included that application's primary activity. 


\subsubsection{Where are AI software applications used today?}
\label{sec:taaft:current}

\begin{figure}[ht]
    \centering
    \includegraphics[width=0.9\linewidth]{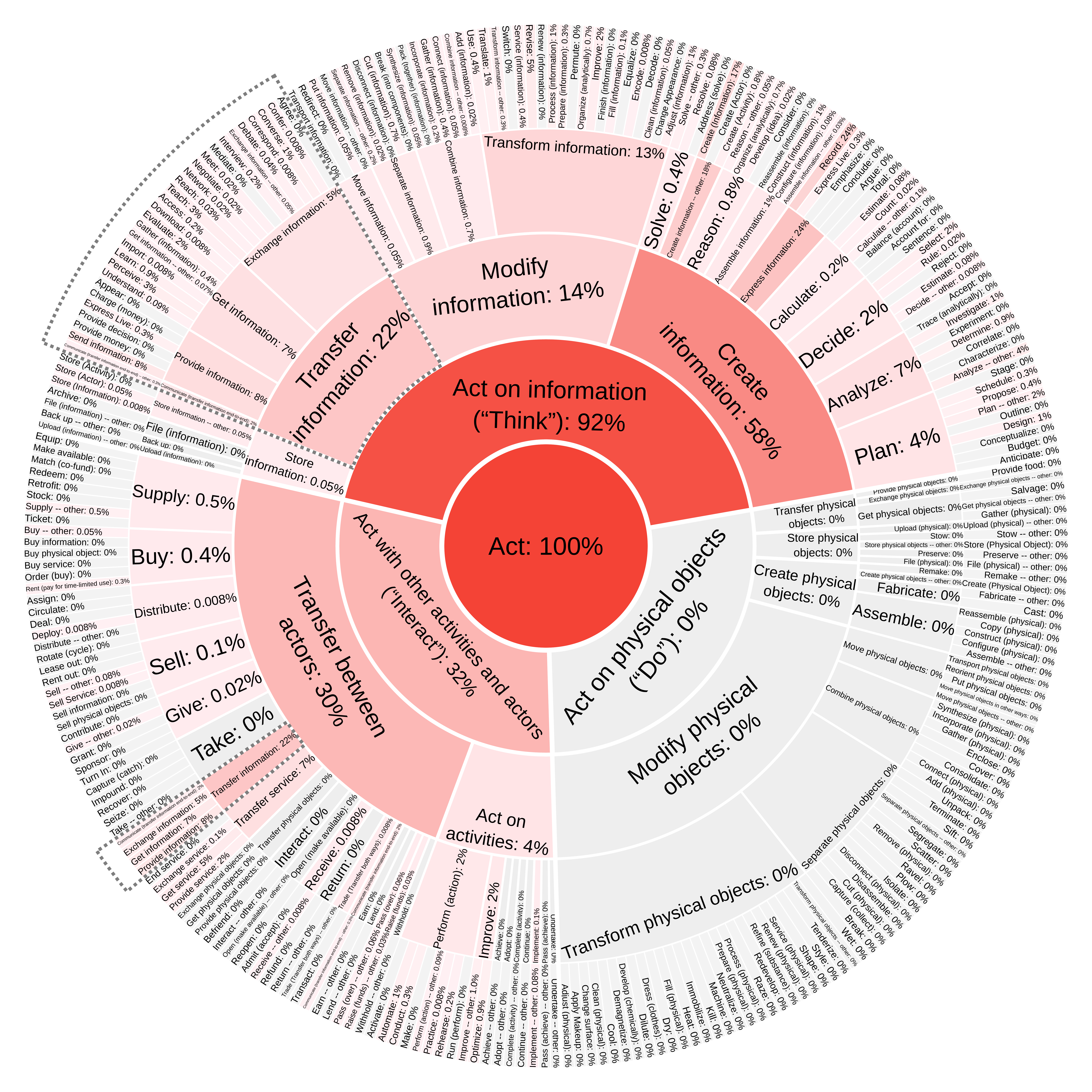}
    \caption{
        \textbf{Where are AI software applications used today?}
        Sunburst diagram illustrating the concentrations of AI software applications across activity types. Color intensity represents the proportion of AI applications associated with each activity (including its specializations in the outer rings) relative to the total number of AI applications in the dataset. We view this as a rough indicator of \textit{AI applicability}. (Percentages in a ring sum to more than 100\% when activities are classified in multiple places, but these double counts are removed at subsequent higher levels.) \textbf{Note}: For this and all other sunburst diagrams in the paper, readers are encouraged to zoom in to examine fine-grained specializations.
    }
    \label{fig:taaft_2025}
\end{figure}


The resulting distribution of AI software applications is shown in the sunburst diagram in~\autoref{fig:taaft_2025}. Instead of the outline format shown in \autoref{fig-ontologyexcerpt}, this diagram shows ``Act,'' at the center, and successive outer rings correspond to increasingly specialized types of activity. For legibility, only the first five levels of the ontology are shown here, but Appendices~\ref{ap:taaft:full_ontology} and \ref{ap:taaft:full_ontology_direct} show all fourteen levels of the complete ontology.

For each ontology activity, the percentage value shown represents the percentage of AI applications classified under that activity, including all of its specializations in the outer rings. The
color intensity of each segment reflects this percentage of AI usage, and we interpret these AI concentrations as a rough indicator of the \textit{AI applicability} for each type of activity.

The diagram also highlights an important structural feature of the ontology: some activities have multiple generalizations (or ``parents," see ``multiple inheritance'' in Section~\ref{sec:background:multiple}). For example, the activity ``Transfer information'' is a child of both ``Think'' and ``Interact,'' so it appears in two locations in the diagram. Both of these instances are indicated in the diagram by gray dashed outlines (in the lower left and upper left quadrants). In these cases, the percentage of AI applications in a given ring may sum to more than 100\%, but this double-counting is corrected at the higher levels where the multiple branches come together. (See Appendix~\ref{ap:taaft:multiple_inheritance} for more examples). 

\paragraph{Is AI software used most for thinking, doing, or interacting?} 
At the second level of the ontology, all of the AI software applications fall under the high-level categories ``Act on information (`Think')'' and ``Act with other activities and actors (`Interact')''. In contrast, \textit{no} AI software applications are classified under ``Act on physical objects (`Do'),'' because the TAAFT dataset includes only digital applications. As we will see, in Sections~\ref{sec:robotics} and~\ref{sec:comparison}, however,  robotic systems complement AI software in this dimension.  

An examination of the darkest-shaded regions in~\autoref{fig:taaft_2025} indicates that the vast majority of AI applications (approximately 92\%) are concentrated within the high-level category, ``Think.'' Within this category, activities related to ``Create information'' constitute the largest share (58\%), followed by ``Transfer information'' (22\%, multiply-inherited), and ``Modify information'' (14\%). Together, these three activity types dominate the current AI software application landscape, reflecting the central role of information-related activities in contemporary AI software uses.

The ``Interact'' category accounts for 32\% of AI applications. However, the majority of this share is attributable to the multiply-inherited activity ``Transfer information'' (22\%), which is a specialized type of both ``Think'' and ``Interact.'' This overlap suggests that many interactive AI software applications also heavily feature information-related processes.

Overall, 92\% of applications are mapped to only 6.8\% of activities, all of which are under ``Think'' in the ontology. This highly skewed distribution suggests that the use of contemporary AI software tools is very concentrated in only a few types of activities. This has at least two possible implications. First, it suggests specific places to look for applications of today's AI software tools. And, second, it suggests opportunities for further development of new types of AI software for the activities that \textit{aren't} heavily populated yet.  

\begin{figure}[ht]
    \centering
    \includegraphics[width=0.99\linewidth]{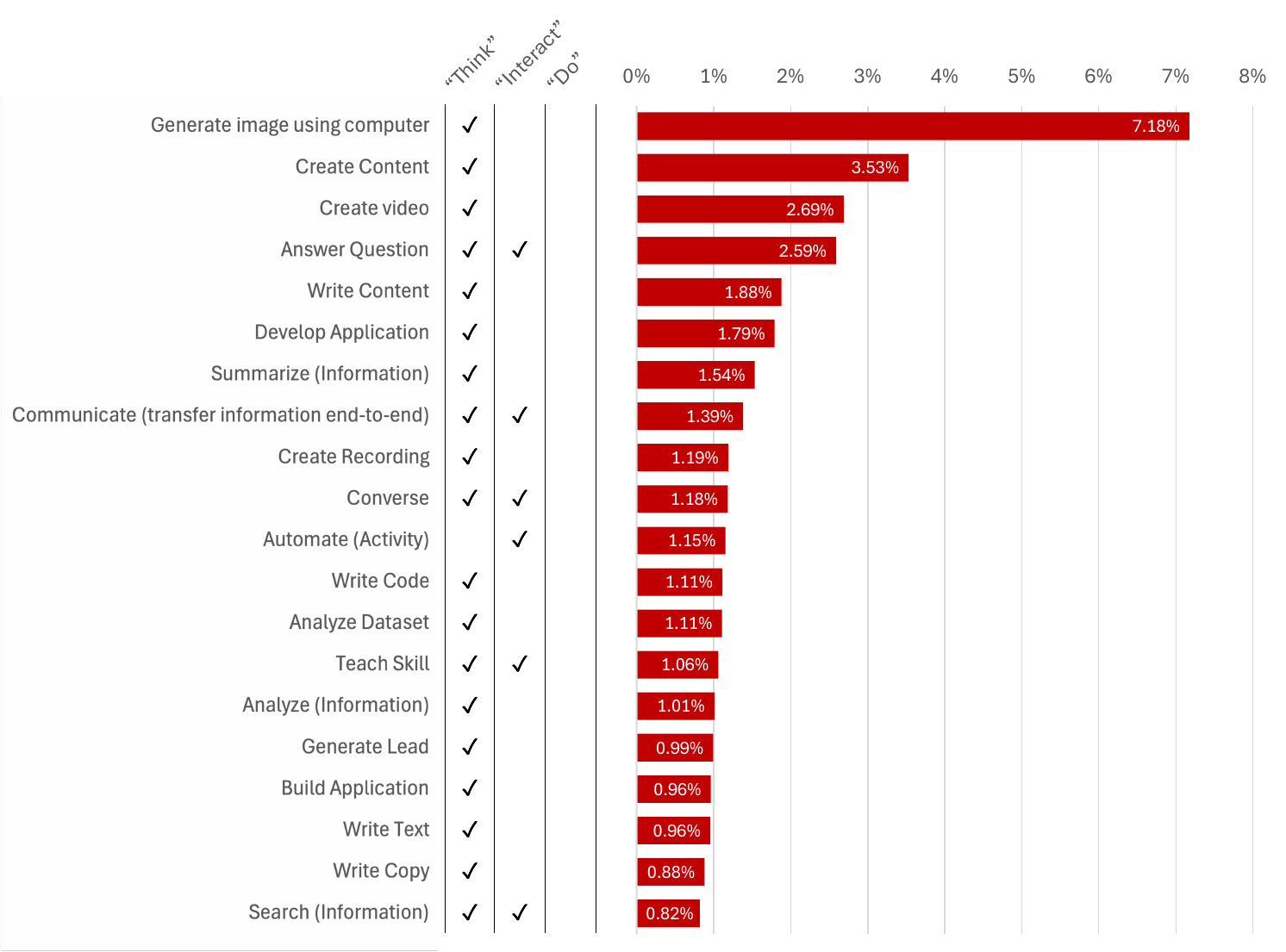}
    \caption{
        \textbf{The 20 activities with the most AI software applications.} These 20 activities alone account for more than 35\% of all AI software applications.  
    }
    \label{fig:taaft_popular_activities}
\end{figure}

\paragraph{In what activities is AI software used most?}

\autoref{fig:taaft_popular_activities} shows the individual activities into which the most AI software applications were classified directly. Several clear patterns emerge in this figure. First, the highest-ranked activities are overwhelmingly centered on content generation and information manipulation, such as ``Generate image using computer,'' ``Create content,'' ``Create video,'' ``Write content,'' ``Develop application,'' ``Summarize (Information),'' and ``Create recording.'' These activities under ``Think'' account for a substantial fraction of all AI applications, reflecting the dominant role of generative capabilities in the current AI ecosystem. 


\begin{figure}[t]
    \centering
    \includegraphics[width=1.0\linewidth]{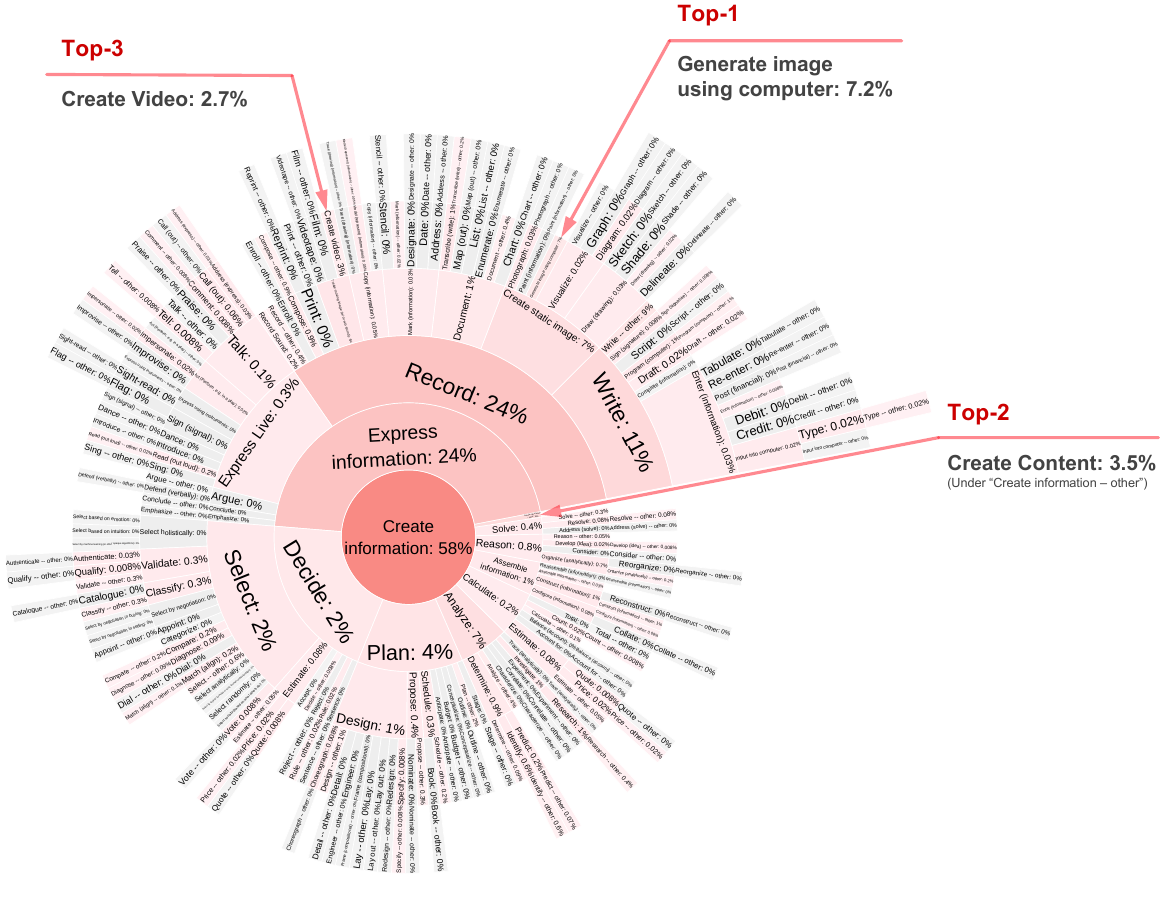}
    \caption{\textbf{Distribution of AI applications across activities within the ``Create information'' branch of the ontology}. Consistent with the term ``generative AI," most current AI applications (58\%) involve various forms of creating information, including images, videos, and other kinds of content.}
    \label{fig:taaft_create_infomation}
\end{figure}

Beyond pure generation, activities such as ``Answer question,'' ``Communicate,'' ``Converse,'' and ``Teach skill,'' also appear prominently. These activities are multiply inherited from both ``Think'' and ``Interact.'' Their strong presence indicates that many widely adopted AI software applications operate at the intersection of information processing and interpersonal interaction, rather than within purely informational or purely social domains. Meanwhile, activities such as ``Automate (Activity),'' that are classified only as ``Interact'' tend to rank lower. 

In general, the distribution of AI software applications across activities is highly skewed. The top 20 activities, representing only about 0.1\% of all activities in the ontology, collectively account for more than 35.0\% of all AI software applications. This pronounced concentration confirms that current AI adoption is dominated by a small subset of highly tractable and commercially valuable activities, while the vast majority of potential activities in the ontology remain sparsely populated. As discussed further in Section~\ref{sec:discussion:opportunities}, this highlights substantial opportunities for future expansion and diversification of AI use.

To investigate in more detail the activities with the highest concentration of AI software applications,~\autoref{fig:taaft_create_infomation} shows a zoomed-in sunburst view of the ``Create information'' branch of the ontology, which includes 58\% of all AI applications. 

The three activities with the most AI software applications, ``Generate image using computer,'' ``Create Content,'' and ``Create video,'' are all present as specializations of this branch. Note, by the way, that even though their AI software intensity is relatively high, the angular width of these activities in the sunburst diagram is small because they have relatively few specialized types of ontology activities classified under them.

\begin{figure}[ht]
    \centering
    \includegraphics[width=1.0\linewidth]{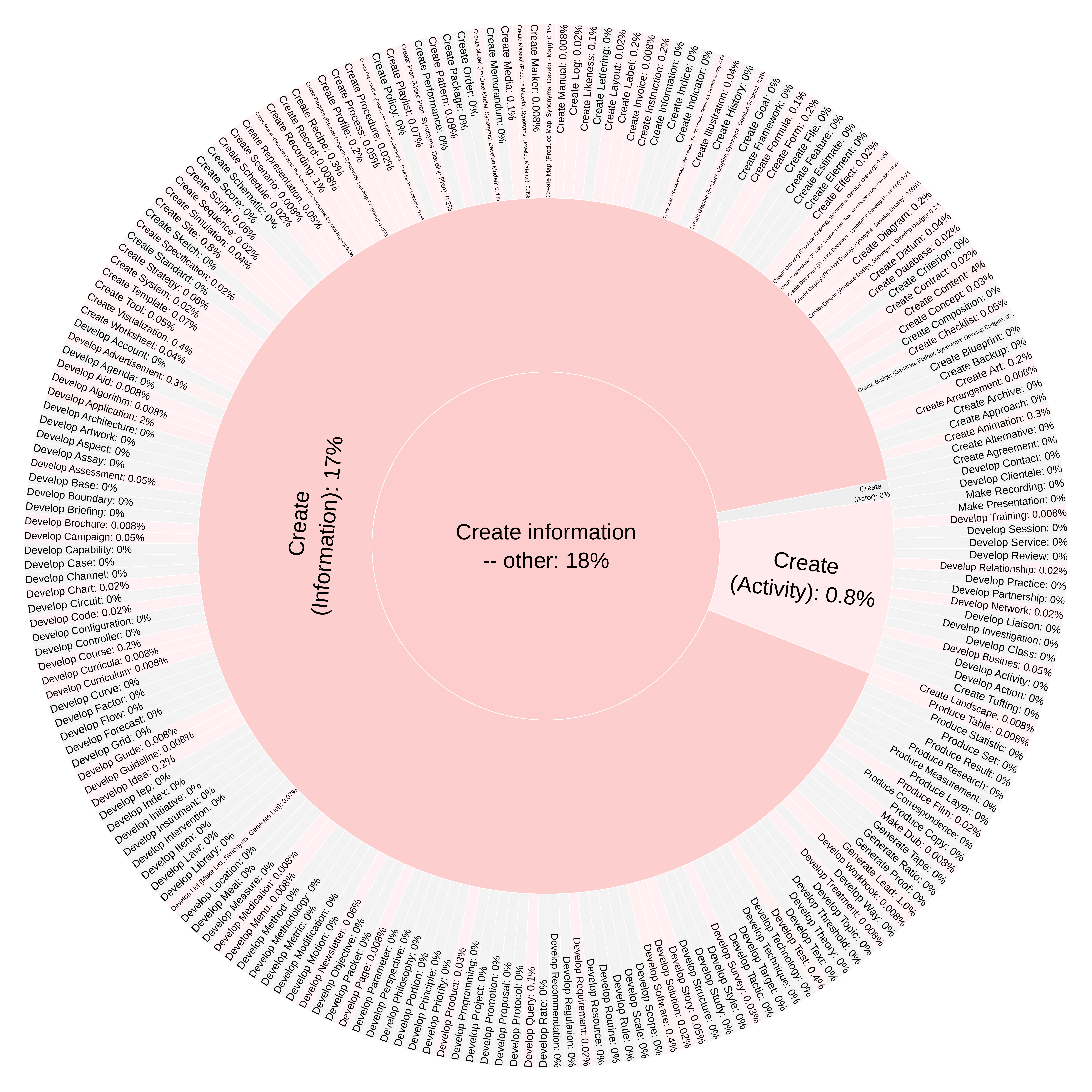}
    \caption{\textbf{Distribution of AI applications across activities within the ``Create Information – other'' branch of the ontology}. The diagram includes detailed examples of activities both with and without AI applications, and as discussed in Section~\ref{sec:discussion:opportunities},  this provides a kind of checklist for different types of opportunities. (Readers are encouraged to zoom in to see detailed activity names in the diagram.)}
    \label{fig:taaft_create_infomation_other}
\end{figure}

At this finer level of granularity, AI applications targeting informational activities are further concentrated in a small number of specialized subcategories. In particular, ``Express information'' accounts for 24\% of applications, ``Create information -- other'' (which includes more specialized activities that feature the verb ``Create'' and its synonyms along with specific objects, such as ``Create budget,'' ``Create content,'' and others) accounts for 18\%, and ``Analyze'' accounts for 7\%. 

\autoref{fig:taaft_create_infomation_other} shows a further zoomed-in view of the ``Create information -- other'' collection, which comprises 222 atomic activities located at the outermost level of the ontology. Among these activities, only 99 currently have AI software applications, while the remaining 123 have no observed AI applications. The 18\% of AI applications associated with this collection are thinly distributed across the covered activities, with the majority of atomic activities receiving less than 0.1\% of applications each. Appendix~\ref{ap:taaft:atomic_activities} provides illustrative examples of the five of these atomic activities with the highest AI concentration, along with the O*NET tasks that include these atomic activities. And Section~\ref{sec:discussion:opportunities} discusses how frameworks like those in~\autoref{fig:taaft_create_infomation_other} and Appendix~\ref{ap:taaft:atomic_activities} can help systematically analyze and identify possible new opportunities for using AI.

\paragraph{In what activities is AI software not being used?}

While current AI software applications are densely clustered around a small set of activities, a substantial number of activities remain sparsely represented or entirely unpopulated by these applications. Excluding activities under the ``Do" category that exhibit zero counts (shown in gray), several ``Think" and ``Interact" activities show no recorded AI applications. These include, among others, ``Administer (treat),'' ``Collaborate (Actor),'' ``Comply,'' ``Attend (go to),'' ``Confer (Actor),'' and ``Analyze (Physical Object),'' each accounting for 0\% of observed AI applications (as shown in the full sunburst diagram in Appendix~\ref{ap:taaft:full_ontology}). 

Notably, some of these underrepresented activities correspond to forms of work that are either highly context-dependent, socially embedded, or physically situated, which may partially explain their limited AI adoption. In other words, these activities may require high levels of the human capabilities included in the EPOCH framework (Empathy, Presence, Opinion, Creativity, and Hope)~\cite{loaiza_epoch_2024}. 

In addition, several activities that involve authoritative, directive, or passive roles, such as ``Order (authoritative),'' ``Authorize,'' ``Assign,'' and ``Receive,'' are also absent from the current AI application landscape. These activities may be less amenable to automation or may offer comparatively limited value when delegated to AI systems, given prevailing human preferences, accountability concerns, and organizational norms. Taken together, the absence of AI applications in these areas highlights important gaps and boundaries in current AI capabilities and adoption. These gaps point to potential opportunities for future research, engineering, and entrepreneurial activity. 

Viewed through the lens of our ontology, the distribution of AI applications reveals a strong emphasis on information-centric and cognitively intensive activities, including content generation, modification, and analysis. At the same time, this analysis highlights a pronounced imbalance in AI deployment, with certain regions of the activity space appearing highly saturated while others remain comparatively underexplored. Underrepresented activities are likely constrained either by current technical limitations, limited availability of relevant data, or insufficient economic incentives for development. 

In many ways, the \textit{negative} examples of activities without much AI usage so far may be as interesting as the positive ones, and we elaborate further on these possibilities in Section~\ref{sec:discussion:opportunities}.

\subsubsection{How have uses of AI software applications evolved?}

\begin{figure}[ht]
    \centering
    \includegraphics[width=1.0\linewidth]{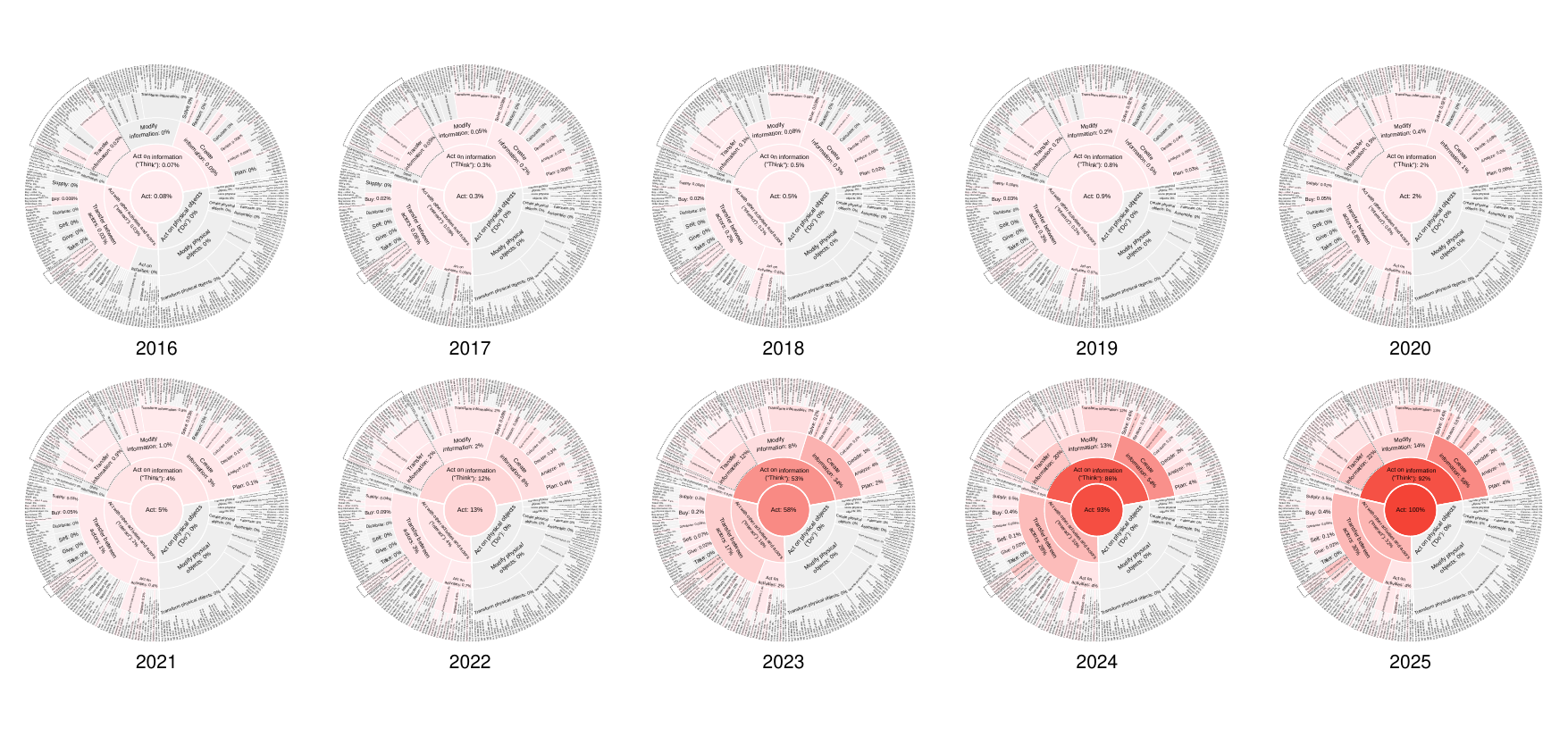}
    \caption{
        \textbf{Sunburst diagrams illustrating the changes in usage intensity of AI software applications from 2016 to 2025}. 
        Each diagram represents the distribution of AI applications across ontology activities for a given year. Color intensity encodes the percentage of applications relative to the 2025 distribution. Color shading is thus fixed and directly comparable across years, enabling longitudinal comparison of category growth and concentration. This suggests substantial growth in the number of applications over time but relatively little broadening in the types of activities for which applications are available. (See larger diagrams for each year in Appendix~\ref{ap:taaft:sunbursts}.)
    }
    \label{fig:taaft_combined}
\end{figure}

In addition to the cross-sectional analysis just described, we also conducted a longitudinal study of how  uses of AI software applications have evolved over time. Using the launch dates of AI applications recorded in the dataset, we analyzed changes in both the intensity and functional distribution of AI software applications over the past decade.

\paragraph{Emergence of new specializations}
\autoref{fig:taaft_combined} reveals a clear and accelerating expansion of AI applications from 2016 to 2025, accompanied by increasing breadth of coverage within the ontology. In the early period (2016--2022), AI applications are relatively sparse and distributed across a limited number of activities, with shallow penetration into specialized branches. 

A pronounced inflection point emerges after 2022, with the rise of generative AI, when growth becomes substantially more rapid. This shift is reflected both in increased color intensity and increased spread into deeper layers of the ontology. By 2023--2025, the diagrams exhibit markedly broader and darker regions, indicating a sharp rise in the number of AI applications across many branches. The use of consistent color scaling across years further highlights that recent growth is driven primarily by intensification within already dominant informational and cognitive activity categories, rather than by uniform expansion into previously underrepresented functional areas (year-specific visualizations with independent color scaling are provided in Appendix~\ref{ap:taaft:sunbursts}.)

\begin{figure}[h]
    \centering
    \includegraphics[width=1.0\linewidth]{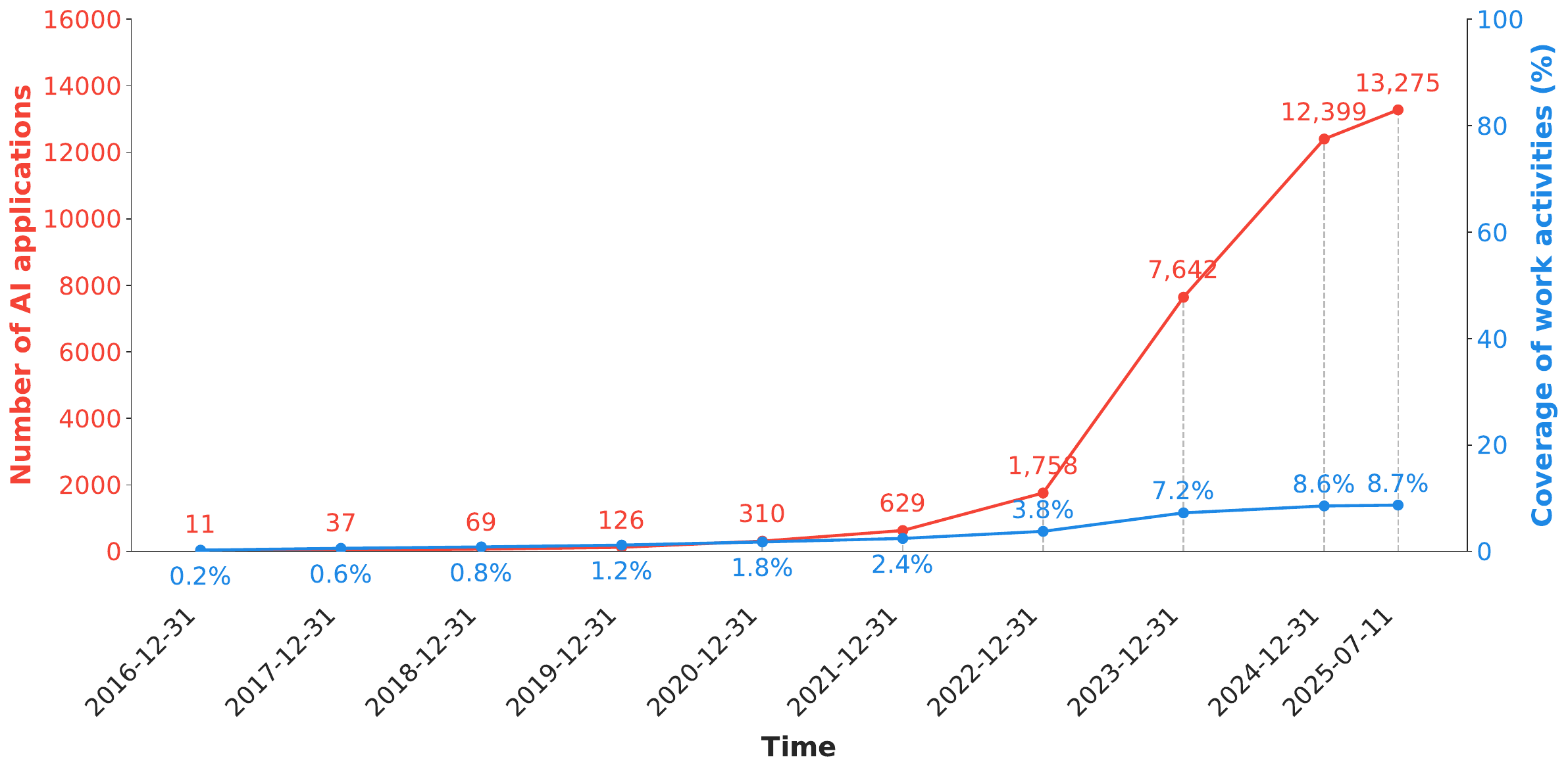}
    \caption{
        \textbf{Growth in AI software applications from 2016 to 2025.} Significant growth in the number of applications began in 2022 with the emergence of generative AI, but it appeared to slow in 2024-25. The breadth of activities for which AI applications were available also increased over time, but at a much slower rate than the number of applications.
    }
    \label{fig:taaft_trend_combined}
\end{figure}

\paragraph{Growth in scale versus functional coverage}
\autoref{fig:taaft_trend_combined} shows the temporal growth in the number of AI applications and the coverage of work activities from 2016 to 2025. Coverage of work activities is defined as the proportion of activities in the ontology for which there is at least one AI application, relative to the total number of activities in the ontology.



In 2016, the ecosystem was still nascent, with only 11 recorded AI applications covering 0.2\% of activities. Growth during the early years (2017--2021) was gradual in both scale and coverage, reaching 629 applications and 2.4\% coverage by 2021. A clear inflection point appears after 2022, with the number of applications rising sharply to 7,642 in 2023 and surging further to 12,399 in 2024. Over this two-year period, the number of AI applications increased by a factor of 6.0 times, while coverage expanded by only a factor of 1.2 times.

Taken together, these trends suggest that despite the rapid growth in the number of AI applications, the current AI ecosystem exhibits limited functional diversity and generalizability. Development remains concentrated within a relatively narrow subset of task categories, indicating that advances in AI capabilities have thus far translated into depth in a few activities rather than breadth across the space of human activities.

\subsection{Robotic systems}
\label{sec:robotics}

To understand the uses of robotic systems, we used the annual \textit{World Robotics} reports~\cite{IFR2025Service, IFR2025Industrial} published by the International Federation of Robotics (IFR). These reports comprise the world’s most comprehensive dataset on robotic systems, including 2024 data on 20.8 million robots used for purposes such as cleaning, laser cutting, welding, and preparing food. (See Section \ref{robotic_systems} for more details.)


\subsubsection{Where are robotic systems used today?}
\label{robotic_systems}

\begin{figure}[htbp]
    \centering
    \includegraphics[width=0.9\linewidth]{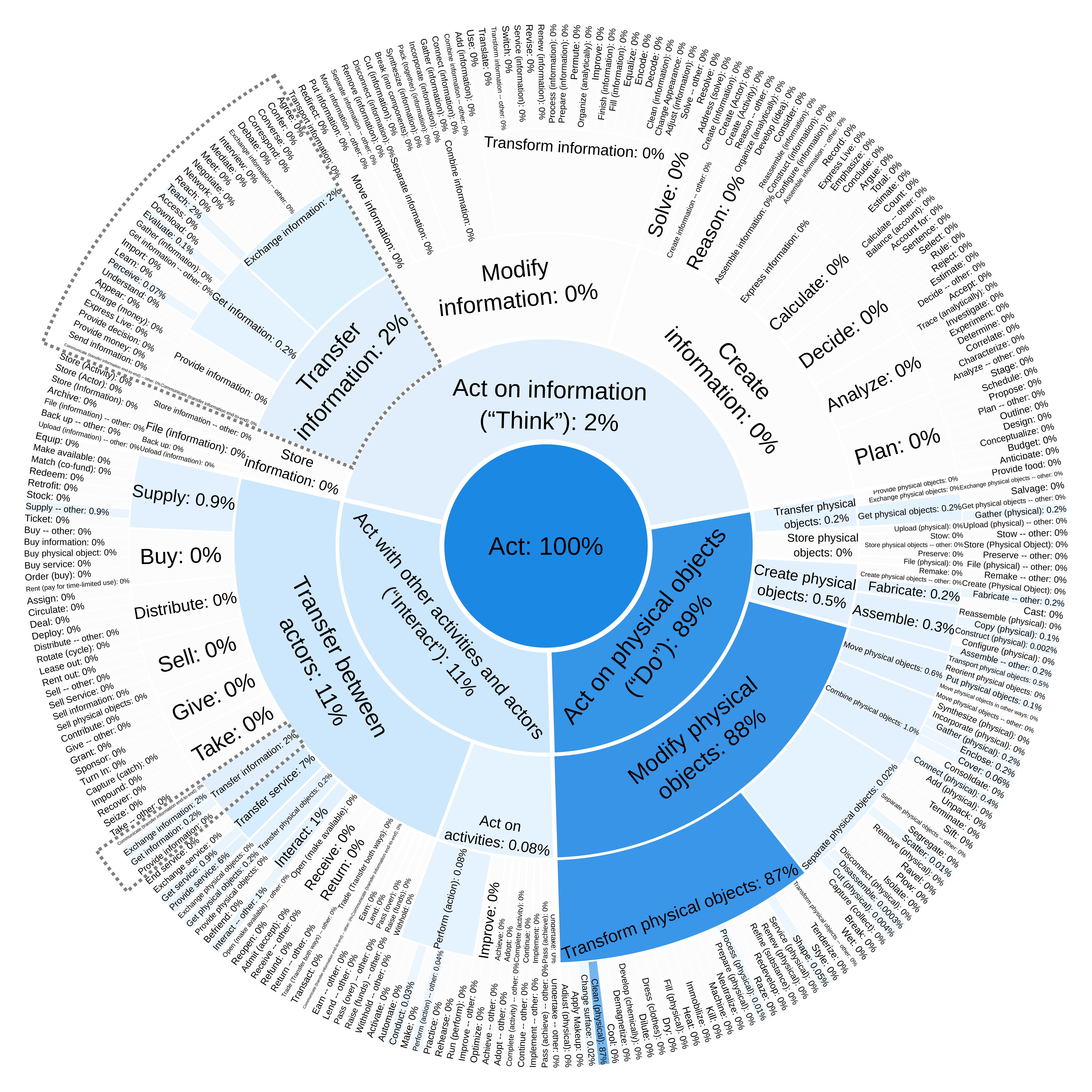}
    \caption{
        \textbf{Where are robotic systems used today?}
        Sunburst diagram showing the distribution of robotic systems deployed in 2024. Percentages and color intensity indicate the proportion of robotic systems associated with each activity relative to the total number of robotic systems deployed in that year (Percentages in a ring sum to more than 100\% when activities are classified in multiple places, but these double counts are removed at subsequent higher levels.). 
    }
    \label{robot_inst_2024}
\end{figure}

Using the manual classification methods described in Section~\ref{sec:method:pipeline:manual}, we classified the 20.8 million robotic systems from the IFR dataset into the activities in our ontology. 
\autoref{robot_inst_2024} shows these results in a sunburst diagram. As before, the percentage values and color intensities indicate the share of robotic systems in that activity relative to the total number of robotic systems, with each deployed unit weighted equally. Also, as before, we interpret these measures of robotic system concentrations as a rough indicator of the \textit{AI applicability} for each type of activity.


\paragraph{Are robotic systems used most for thinking, doing, or interacting?} 

Not surprisingly, the overwhelming majority of robot deployments (89\%) fall under the ``Act on physical objects (`Do')'' category. This finding is consistent with the defining characteristic of robots: their embodiment and operation within the physical environment. The most common activity in the ``Do" branch by far is ``Clean floor,'' due, in large part, to the fact that 76.7\% of all 2024 robots deployments were low cost floor-cleaning robots. 11\% of robotic system deployments are classified under ``Act with other activities and actors (`Interact').'' And 2\% of these systems are classified under both  the ``Think'' and ``Interact'' ontology branches because they teach people or provide navigational guidance to people in public places, activities that involved both cognitive and interactive tasks.

Because the category ``Clean floor'' contains an exceptionally large number of robots, it exerts a disproportionate influence on the aggregate results. To reduce the influence of this factor, we re-examined the distribution of robotic activities after excluding this category. ~\autoref{exclude_clean_floor_robot_inst_2024} shows that, once floor-cleaning robots are removed, the activity distribution becomes nearly balanced between ``Do" (55\%) and ``Interact" (45\%). 


\paragraph{In what activities are robots used most?}

\begin{figure}[ht]
    \centering
    \includegraphics[width=0.99\linewidth]{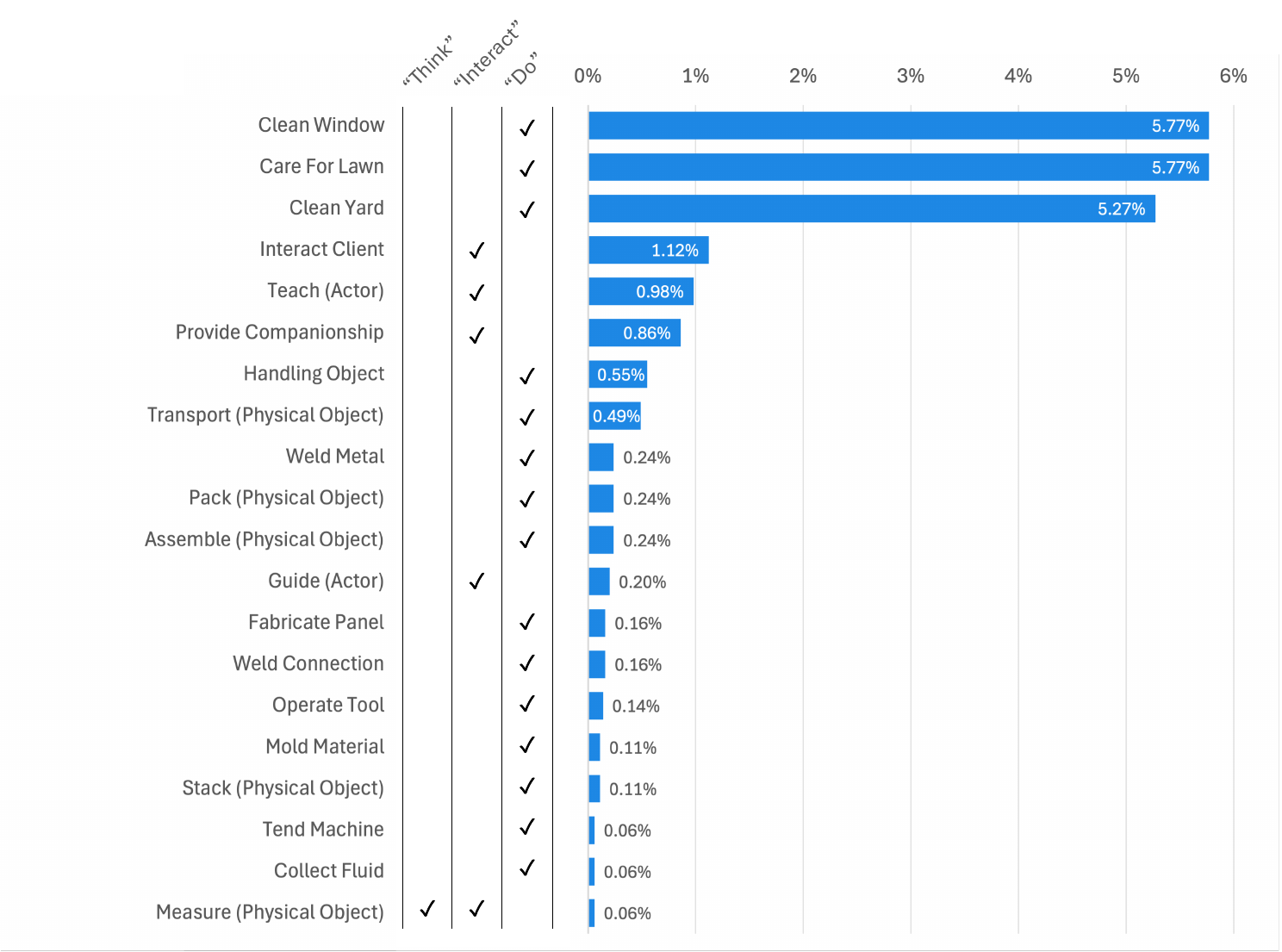}
    \caption{
        \textbf{Top 20 robotic activities in 2024} (excluding ``Clean floor'' robots, which account for 76.7\% of total installations).
    }
    \label{fig:robot_popular_activities}
\end{figure}

\autoref{fig:robot_popular_activities} shows the 20 most prevalent activities performed by robots, after ``Clean floor" is excluded. The most popular activities outside of ``Clean floor'' which accounts for 76.7\% of total installations are ``Clean windows'' (5.8\%), ``Care for lawn'' (5.8\%), and ``Clean yard'' (5.3\%). All of the top four activities are types of ``Do," and they are all performed by service robots in less-structured environments potentially among humans.

Interestingly, the next most popular activities are ``Interact client,'' ``Teach (Actor),'' and ``Provide companionship.''  These activities all fall under ``Interact," and collectively, they account for almost 3\% of robot installations. These activities are physically situated and socially embedded and require advanced systems of social intelligence. In other words, although some of those activities, such as ``Teach,'' are commonly performed by AI software applications, social robots attempt to improve the effectiveness of the AI systems with the use of additional modalities, physical social presence, and affective engagement, which are important for activities such as teaching.

Lastly, 12 out of the 14 remaining activities on the list are, again, classified under ``Do." They include ``Transport (Physical Object)''\textemdash primarily due to scale deployment of robots in warehouses\textemdash ``Weld metal,'' and ``Fabricate panel.'' These activities are all performed by robots in well-structured environments.

\subsubsection{How have uses of robotic systems evolved?}

As with AI software applications, we conducted a longitudinal study of how  uses of robotic systems have evolved over time. In this case, however, the IFR data available about \textit{numbers} of robots deployed for each type of activity was not sampled in a way to make it directly comparable across years \cite{IFR2025Service}. Instead, only the cross-sectional \textit{proportions} of robots in different activities were comparable across years.  

\begin{figure}[h]
    \centering 
    \includegraphics[width=1.0\linewidth]{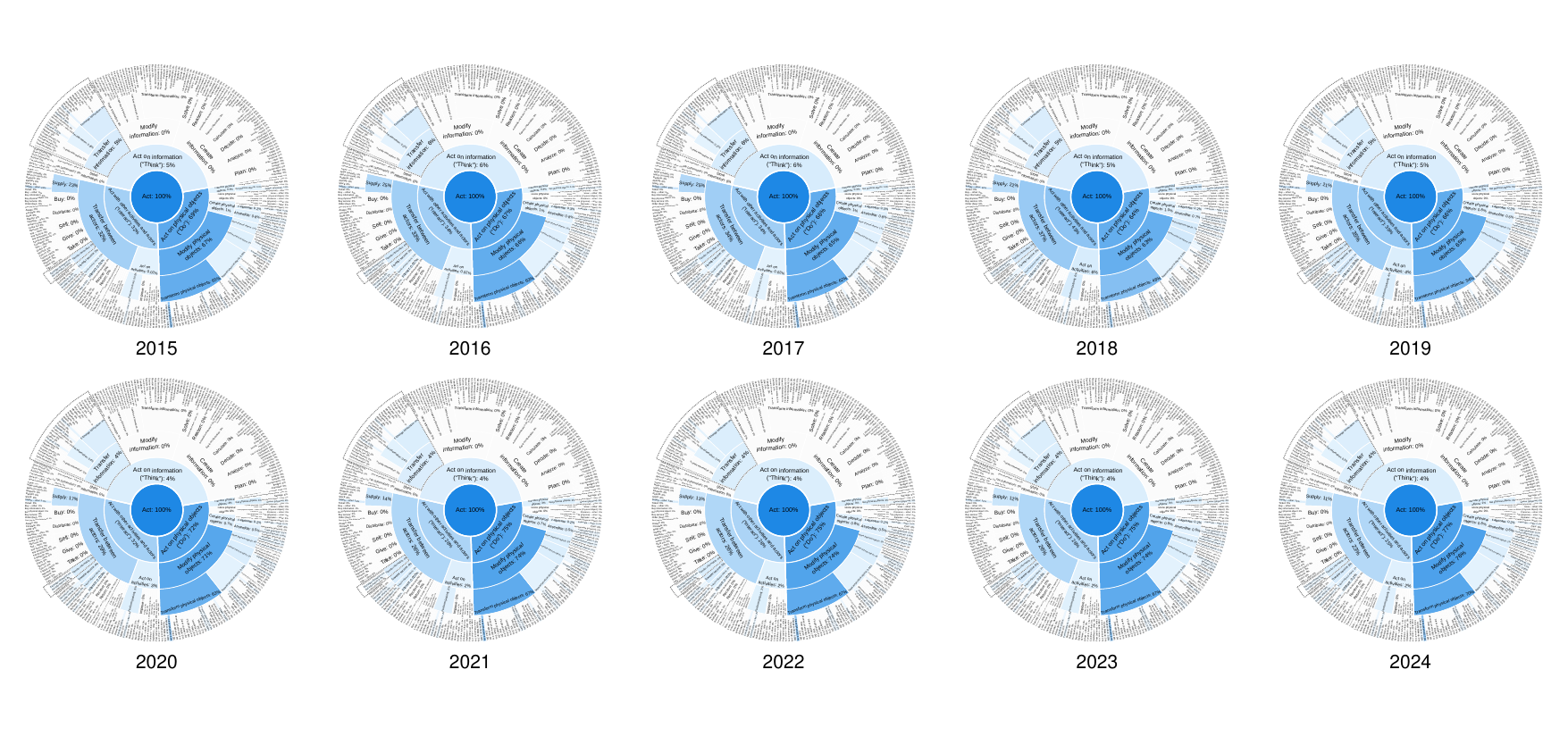}
    \caption{
        \textbf{Sunburst diagrams illustrating the changes in distribution of robotic systems across activities from 2015 to 2024}.
        Each diagram shows how installations of robotic systems are distributed across ontology activities for a given year. Color intensity encodes the proportion of robotic systems deployed in an activity, relative to the total robotic systems deployed that year. In general, this figure suggests that there may be less broadening over time of robot uses across different types of activities than of AI software applications. (See larger diagrams for each year in Appendix~\ref{ap:robot:sunburst}.)
    }
    \label{robot_long_2015_2024_distribution}
\end{figure}

Accordingly, \autoref{robot_long_2015_2024_distribution} shows the changes between 2015 and 2024 in the proportional \textit{distributions} of robot installations across the different types of activities defined in our ontology. Fortunately, the IFR uses a relatively consistent taxonomy for different types of robots over time, and when necessary to ensure consistency, we implemented small classification adjustments. For example, we consolidated the originally fine-grained agricultural subclasses (e.g., broad acre, greenhouse, fruit-growing, vineyard) into a single category, “Cultivation.”

But perhaps the most surprising aspect of this figure is how little the distribution across activities changed over the 10 year period shown. In the AI software applications shown in Figures \ref{fig:taaft_combined} and \ref{fig:taaft_trend_combined}, the number of activities for which software applications are used more than doubled in the two-year period from 2022 to 2024. But in the ten-year period shown in Figure \ref{robot_long_2015_2024_distribution}, there is relatively little change in which activities use robots. Some of this is presumably because of stability in the IFR categories used to classify the robots. But the IFR did change some categories during this period, and if there had been substantial growth of robot uses in other activities, it seems likely that IFR would have added some new categories. 

In fact, even the distribution of robot use across these activities has had relatively little change. Most of the activity types changed by only a few percentage points over the ten-year period. And the largest exception to this was the activity ``Modify physical objects," which is mainly affected by cleaning robots, and which increased from approximately 66\% in 2015 to approximately 76\% in 2024. 

So, while the IFR is currently the best source for studying robotic applicability on work-related activities, our analysis raises the question of why the uses of robots haven't yet spread more widely across activities like software applications have and indicates the need for more nuanced taxonomies of robotic systems. 


\subsection{An integrated view of how AI software applications and robotic systems are used today}
\label{sec:comparison}

\begin{figure}[h]
    \centering
    \includegraphics[width=0.9\textwidth]{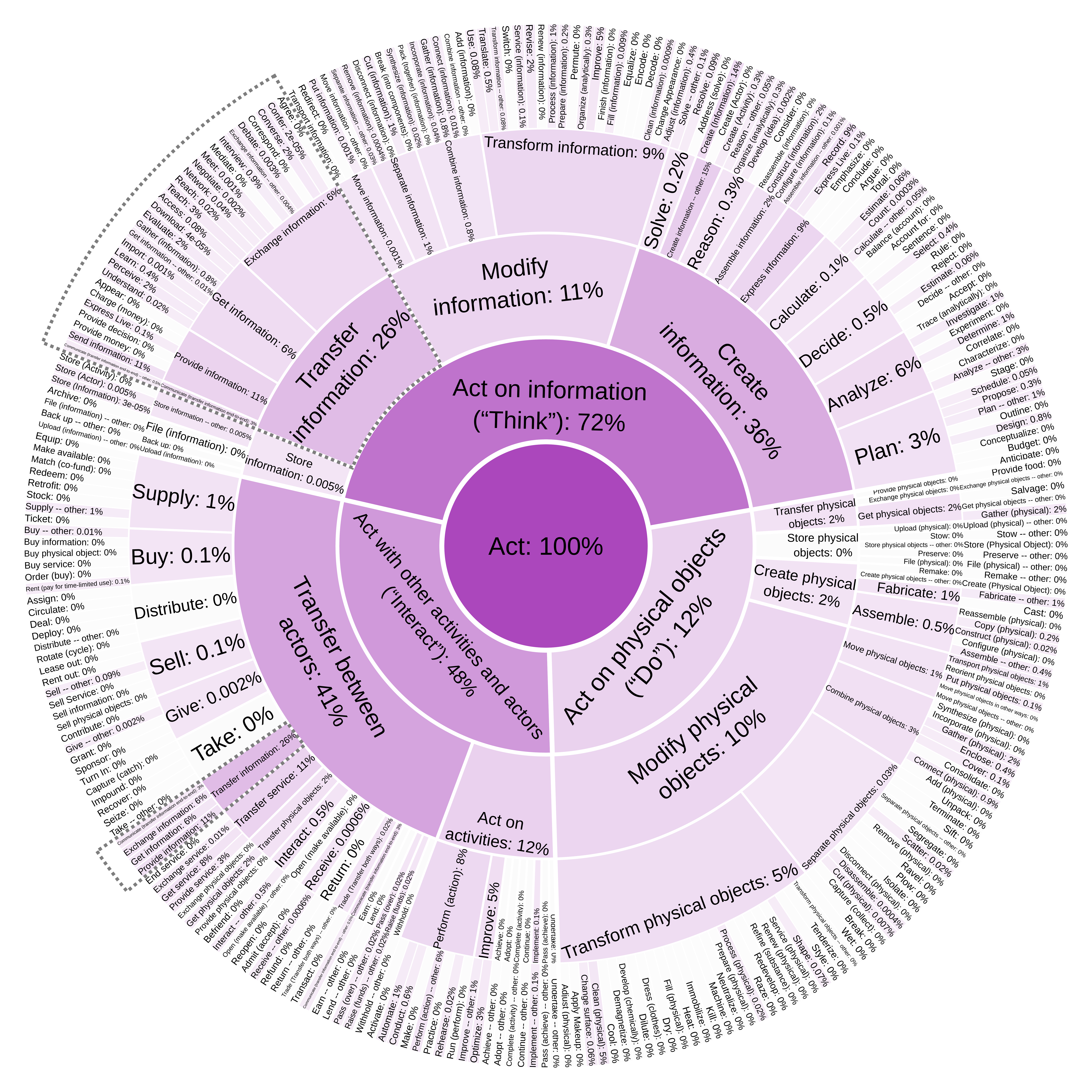}
    \caption{
        \textbf{Overall view of how AI is used today.} Both AI software applications and robotic systems are included. Numbers and color intensity show the percent of market value associated with each activity in the ontology. (Percentages in a ring sum to more than 100\% when activities are classified in multiple places, but these double counts are removed at higher levels.)
    }
    \label{fig:combined_revenue}
\end{figure}

\autoref{fig:combined_revenue} provides a concise summary of the empirical results reported in this paper. Using the methods described in Section \ref{sec:method:pipeline:combined}, 
we combined the weighted market values of both AI software applications and robotic systems and mapped them to the ontology activities where they were classified. The figure shows an estimate of how the market value of all AI systems used today is distributed across the work activities these systems help perform.


Like previous figures in this paper, this one confirms the highly uneven uses of AI across different activities. In this case, only 1.6\% of activities account for over 60\% of AI market value. And many activities account for little or none of the AI market value. As described in the Discussion section below, this suggests not only where to find promising opportunities for deploying today's AI systems; it also suggests many possible places that new generations of AI could be used in the future\textemdash either for good or for bad. 


The figure also allows us to answer questions about the overall uses of AI systems and how these uses are divided among AI software applications and robotic systems.

\paragraph{Are AI systems overall used most for thinking, doing or interacting?} 

The figure shows that a large majority (72\%) of the estimated AI market value involves our category of ``Act on information (\textit{`Think'}).'' So ``thinking'' is the most important way of using AI market value today. Two specialized types of thinking account for most of this: ``Create information'' (36\%) and ``Transfer information'' (26\%).   

Although some people may think of AI as being mostly about robots, the primary type of activity robots perform is ``Act on physical objects (`\textit{Do}')," and this activity uses only about 12\% of the overall AI market value. So robots that perform purely physical tasks today are a relatively small contributor to overall market value provided by AI. In this segment, ``Modify physical objects'' uses the largest share (10\%).

``\textit{Interact}'' activities account for 48\% of combined AI market value, with ``Transfer between actors'' accounting for 41\%. But interactions generally involve either information or physical objects, and in this case, much of the market value of ``Interact'' is for ``Transfer information'' (26\%) which is shared with ``Think.''

\paragraph{Which type of AI systems (software or robotics) contribute more market value to different activities?} As \autoref{combined_revenue_percentage} in Appendix~\ref{combined_percentages} shows, according to our estimates, AI software applications account for 75\% of the total market value for AI systems, and robotic systems account for 25\%. 
 
 More specifically, AI software is almost completely responsible for the ``Think'' activities, robotic systems are fully responsible for the ``Do'' activities, and the ``Interact'' activities are supported by a combination of AI software (34\%) and robotics (14\%).

\section{Discussion}

In order to analyze and predict where AI can be used, we  created a deep ontology of work activities that incorporates and substantially reorganizes the approximately 20K activities in the US Department of Labor's O*NET occupational database. But unlike O*NET and similar task taxonomies, this ontology is structured to systematically take advantage of the concept of inheritance as used in object-oriented programming and many applied ontologies. This means that the work activities in the ontology are structured into family trees of similar types of activity, and\textemdash importantly\textemdash that properties of the abstract activities in these family trees are usually inherited by their more specialized descendants.

The paper then shows how this ontology can help answer the question of where AI can be used. To do this, we classified two large datasets about AI uses\textemdash in software applications and robotic systems\textemdash into the activities in our ontology. And we assumed that the activities with more current AI uses were those with more \textit{AI applicability}. Crucially, we also assumed that if an abstract activity (like ``Write'') had high AI applicability, then many of its more specialized descendants (like ``Write email to schedule a meeting'') were also likely to have high AI applicability. Finally, we created sunburst diagrams to graphically display the relative AI applicability of all the activities in our ontology.


In future work, we expect to further formalize and empirically test the assumptions made in doing these things. But, as described in the next two subsections, we believe they are already useful in practice for systematically identifying opportunities and risks from using AI.

\subsection{Identifying AI opportunities}
\label{sec:discussion:opportunities}
The results in Section~\ref{sec:result} about which activities are widely supported by current AI applications can serve as a systematic map of potential AI uses. For example, Appendix~\ref{ap:taaft:atomic_activities} shows a number of atomic activities (and their associated O*NET tasks) for which AI applications are already commercially available. This is an obvious way for managers, workers, and others to identify AI uses they can consider adopting now.

More interestingly, for any generic activity in the ontology that has a large number of AI applications, it is likely that AI could be applicable to many of the specializations of that activity, even those activities that do not \textit{yet} have AI applications. For example, Figure~\ref{fig:taaft_create_infomation_other} shows the activity ``Create information -- other'' which has almost one-fifth of all the AI applications in our dataset. But many of the atomic activities on the outer ring of this figure are gray, indicating that they do not yet have any AI applications in our dataset. Similarly, Figure~\ref{robot_inst_2024} highlights specific activity gaps\textemdash i.e., tasks that cannot yet be completed using robots in the robotics dataset~\cite{IFR2025Service, IFR2025Industrial}. 

Thus, these two figures (Figures \ref{fig:taaft_create_infomation_other} and \ref{robot_inst_2024})  can serve as systematic \textit{checklists} for considering possible uses of AI. To use them in this way, we can consider at least three reasons why an activity (like ``Create information -- other'') with many AI applications would have specializations that \textit{don'}t have many applications. Each of these reasons provides a different kind of opportunity. For example:
\begin{enumerate}
    \item \textit{Technical opportunities}. For some of the activities, there may not yet be AI techniques that are capable of supporting the activity. For instance, looking at Figure~\ref{fig:taaft_create_infomation_other} might suggest that for some specific activities involving text generation, the current technology may create too many inaccurate hallucinations. In video creation activities, the current quality of AI-generated video may not be suitable. Or in robotics, examining Figure~\ref{robot_inst_2024} might help robotics experts identify gaps that are not yet met by current systems\textemdash e.g., long-horizon task-and-motion planning in unstructured settings; contact-rich manipulation with compliant control; or uncertainty-aware scene understanding and state estimation. In all these cases, there are opportunities for researchers and developers to create new AI capabilities.
    
    \item  \textit{Economic opportunities}. In other cases, there may be activities for which current AI capabilities could support the activity, but it just isn't economically useful to use AI in this way. This provides entrepreneurial opportunities, such as (a) developing better ways of identifying customers willing to pay more for an AI-enabled service, or (b) identifying less expensive ways to obtain the data needed to train a suitable AI model. 
    \item \textit{Unrecognized opportunities}. There may also be activities for which there are AI uses that are both technically and economically feasible, but no one has recognized and taken advantage of this yet. These may be some of the most promising entrepreneurial opportunities of all.
\end{enumerate}

\subsection{Identifying AI risks}
In addition to being a tool for identifying AI opportunities, this ontology can also serve as a checklist for identifying potential AI risks. For instance, if we look at only the first four levels in Figure~\ref{fig:combined_revenue}, we see a number of activities for which there are potential risks from the use of AI:

\begin{itemize}
    \item \textit{Create information}. Our societal experience with social media has already demonstrated how the Internet can amplify the \textit{spread} of misinformation of various sorts. Now, we see that AI systems can certainly amplify the \textit{creation} of such information. It is not clear exactly how to deal with these risks, but it is certainly an issue worth consideration. 
    \item \textit{Provide information}. Another kind of risk arises when AI systems provide information that is correct but that enables people to do dangerous things like create atomic or biological weapons. How should we deal with these risks?
    \item  \textit{Sell}. Another area in which AI systems could introduce potential risks is in selling. For example, more and more people are now using AI chatbots to find information about products they are considering buying. Should there be rules requiring chatbots to disclose the sponsors who affect which products the chatbots emphasize? 
    \item \textit{Decide}. More broadly than just purchase decisions, it seems likely that AI systems will increasingly be used to help humans make decisions of all kinds. Are there analogs of truth-in-advertising laws that such systems should comply with? When AI systems are advising people about personal problems like a therapist would do, are there equivalents of professional  codes of conduct? 
\end{itemize}

The issues just listed are only a tiny subset of the issues we, as a society, might want to consider about AI, and there are many other more specialized activities in the outer rings of our ontology (see Appendix~\ref{ap:taaft:full_ontology}) that might suggest other more specialized kinds of issues to consider. 

The ontology presented here doesn't provide answers for how to deal with these risks, and these answers might take many forms, including governmental policies, self-regulation, and changing community norms. But this ontology does help systematically identify AI risks, and we hope future versions of the ontology will also help design governance processes to deal with these risks. 

In general, as \textit{New York Times} columnist Tom Friedman says, AI is being ``infused''~\cite{friedman_opinion_2025} into many parts of our lives. We hope the frameworks in this paper provide a useful conceptual tool for thinking systematically about how this AI-infusion is working and how we might want to influence it.

\subsection{Adapting to changing AI capabilities}
\label{subsec:capabilities}
Even if we had perfect knowledge of all the detailed places where today's AI systems could be used, that does not mean that we would know where future generations of AI systems would be useful. And, since AI capabilities are rapidly changing, it would be very helpful to be able to predict in detail where new AI capabilities could be used.

Fortunately, the inheritance capability of our ontology can make it possible to do this much more systematically than before. As a simple example, consider the ontology activity of ``Dress (clothes)." The AI intensity of this activity is currently shown as 0\%, meaning that no AI systems in our datasets are currently being used to put clothes on a person. But imagine that someday soon, a future robot has the physical and interpersonal skills to do this successfuly. Where could such a robot be used? 

Of course, one could just try to imagine possibilities for this, with greater or lesser degrees of success. But consulting the online platform of our ontology would immediately show a range of examples. The current examples in the ontology include: (a) dress patients who are unable to dress themselves, (b) dress children, and (c) dress mannequins for store displays. And, more morbidly, it even shows an example that many people probably wouldn't have thought of: (d) dress cadavers. 

If you worked for a company that was making such robots, a list like this of potential customer types could be very useful. And if you were trying to assess risks of such robots, such a list could also be very helpful.

Of course, this is a relatively simple example, but the more important implication is that the structure of our ontology could help systematically consider scenarios of many different kinds of AI capabilities and their potential uses down to a very detailed level. And this could be helpful both for identifying new opportunities and for avoiding potential risks.

\subsection{Limitations}
\label{subsec:limitations}
We view the work described here as an early progress report for a promising long-term research agenda, not just about understanding where AI can be used, but also about analyzing and designing work activities in many other ways. As such, there are a number of limitations of the current work.

\subsubsection{Ontology construction} 
We constructed our ontology using a combination of human and AI reasoning, including heavy reliance on the structure of synonyms and hypernyms in the WordNet linguistic database. We believe that this provided a durable, linguistically precise, and human-understandable ontology structure. We also believe that this structure does a good job of capturing the deep similarities in different types of activities, including similarities on the dimension of whether AI is applicable to the activity. 

But there may well be better approaches to constructing useful ontologies for work activities, and we believe an important part of the research agenda in this domain will include exploring various other such methods. This may, for example, include (a) using different analytic techniques (such as semantic embedding and clustering) and (b) involving larger communities of human contributors.  And even with our current ontology, we believe there are multiple ways to improve the naming, organization, and related content (such as definitions and descriptions) of some of the activity categories.

\subsubsection{Application classification pipeline} 
As noted above, our application classification pipeline classified applications into appropriate categories in our ontology at least as well as humans did. But we expect that there are multiple ways to improve the accuracy and reduce the cost of this process. Possibilities include: (a) better LLM prompting, (b) more precise definitions of the activities themselves, and (c) explicit consideration of more properties of activities (such as their sub-activities). Our classification pipeline also did not distinguish between whether an application could \textit{automate} the activity or  \textit{augment} an actor already performing the activity. In the future, we plan to introduce this distinction.


\subsubsection{Estimating AI applicability}
In the Results sections above, we estimate the AI applicability for different activities from the number of units and the market values of AI software applications and physical robots that are available for these activities. But these numbers are certainly not perfect estimators of actual AI applicability. For example, many random factors affect the decisions by software developers and customers about whether to produce or buy an AI application or robot for a given activity, including personal interests and biased perceptions. 

And the method we used to combine the separate intensity estimates for software applications and robots into a single diagram (\autoref{fig:combined_revenue}) is based on estimates of the market value of the applications customers pay for in each area.

All these estimates are based on what we believe are plausible\textemdash but certainly not perfectly accurate\textemdash assumptions. And we leave for future work attempts to empirically validate the accuracy of these predictions.

\section{Future research}
In this paper, we have demonstrated one way of using a deep ontology to organize knowledge about a very comprehensive set of approximately 40,000 work activities. This early work resulted in graphically compelling displays of where AI is being used today.

We believe there is now great potential for using this approach to analyze, design, and improve work processes in many other ways. One way of thinking about the potential of this approach is by analogy to another scientific framework: the periodic table of the elements in chemistry~\cite{emsley_periodic_2011}. While the analogy is certainly not perfect, it provides an intriguing way of thinking about the potential benefits of ontologies of work activities.

In chemistry, the periodic table identifies (a) different fundamental \textit{types} of matter (called ``elements''), (b) different \textit{ways of} \textit{combining} these elements (called ``molecules''), and (c) different \textit{properties} of these elements and compounds (e.g., electronegativity and metallicity). For over 100 years, chemists, chemical engineers, and others have been using this periodic table and the other knowledge it helps organize to create many key elements of our modern industrial civilization.

By analogy to the periodic table, we observe that an ontology of activities can identify (a) different \textit{types} of actions (``activities''), (b) different \textit{ways of} \textit{combining} these activities (e.g., as ``sub-activities'' of a larger activity), and (c) different \textit{properties} of these activities (e.g., ``AI applicability''). In this paper, we have described one way of specifying and organizing such an ontology to represent and predict the activities in which AI can be used.

But we believe this is just the beginning of a broad research agenda. Perhaps the most urgent part of this agenda is developing richer representations of the \textit{parts} (or sub-activities) of various kinds of activities. And a further step is to represent the \textit{processes} for performing an activity, including not only the sub-activities of the activity but also the interdependencies (such as prerequisites and flows) among these sub-activities.

With these elements, we believe there are many opportunities for using this approach to:
\begin{enumerate}
    \item \textit{Optimize existing processes}. For example, with rich libraries of different variations for how a process can be done, automated systems can not just optimize parameters in an existing set of equations for a given process, they can also optimize across different processes for achieving the same goal, each with different performance estimation equations.
    \item \textit{Rapidly adapt processes to new situations}. For example, when unusual situations arise, this approach could be used to automatically construct new processes by systematically examining alternative methods for accomplishing the parts of the standard process that are not possible in the current situation.
    \item \textit{Design innovative new processes}. For example, in smaller-scale earlier work upon which this paper builds, we successfully used distant analogies with other processes to generate unexpected new process ideas~\cite{malone_tools_1999}.
    \item \textit{Analyze and design societal processes}. In addition to using this approach for specific policies in individual organizations, the approach could also be used to analyze and design large-scale societal processes, such as (a) identifying skill changes needed for workers in the AI economy, or (b) identifying potential design changes in widely used processes that would improve quality of life for workers. We can even imagine that this approach could be used to design new forms of societal decision-making, such as new forms of democratic decision-making enabled by new kinds of AI.  
\end{enumerate}

\section{Conclusion}

In this paper, we have presented a broad framework that integrates detailed data about the uses of both AI software applications and robotic systems. We have also shown how organizing this information in a deep ontology provides a powerful basis for systematically identifying both opportunities and risks in  current and future AI uses.

Understanding how we humans can use the increasingly powerful technology of AI may be one of the most important questions facing humanity today. And we hope this work will help business people, technologists, policy makers, and many others address these questions, not only more systematically, but also more wisely.

\section{Methods}

\label{sec:method}

\subsection{Representing knowledge using an ontology}
\label{sec:background:ontology}

The word ``ontology'' originated as a branch of philosophy involving the study of what exists. In particular, the field of ontology seeks to ascertain the most important features of entities found in the world and the relations between them~\cite{Hofweber_2023}. ``Applied ontology'' involves identifying the types of objects that exist in a given field of study along with their properties and relationships~\cite{noauthor_ontology_2025}. For example, the discipline of biology has a set of concepts to describe the types, properties, and relationships of living things. 

Starting in the 1990s, computer science researchers began to use the word ``ontology,'' to refer to formal, digital representations of these applied ontologies~\cite{noauthor_ontology_2025}, and that is the sense in which we will use the word here. More recently, the term ``knowledge graph'' has also been widely used to describe a way of digitally representing ontologies and similar kinds of knowledge about objects and their relationships~\cite{hogan_knowledge_2021}. These ontologies based on knowledge graphs have been developed and widely used in fields such as biology~\cite{gaudet_primer_2017}, medicine~\cite{noauthor_snomed_nodate}, and finance~\cite{noauthor_financial_2023}, and our ontology is one of these.

Of particular importance as a foundation for our work here is an early ontology of business activities called the MIT Process Handbook~\cite{malone_tools_1999}. The Process Handbook included over 5,000 business processes and activities, with an emphasis on simultaneously representing two kinds of hierarchies: (a) hierarchies of increasingly specialized \textit{types} of activities (e.g., writing and its specialized types such as writing novels, writing news stories, and writing scientific articles), and (b) hierarchies of increasingly fine-grained \textit{parts} of activities (e.g., baking a cake and its parts such as assembling ingredients, mixing, and heating).

\begin{figure}[htbp]
    \centering
    \includegraphics[width=0.9\textwidth]{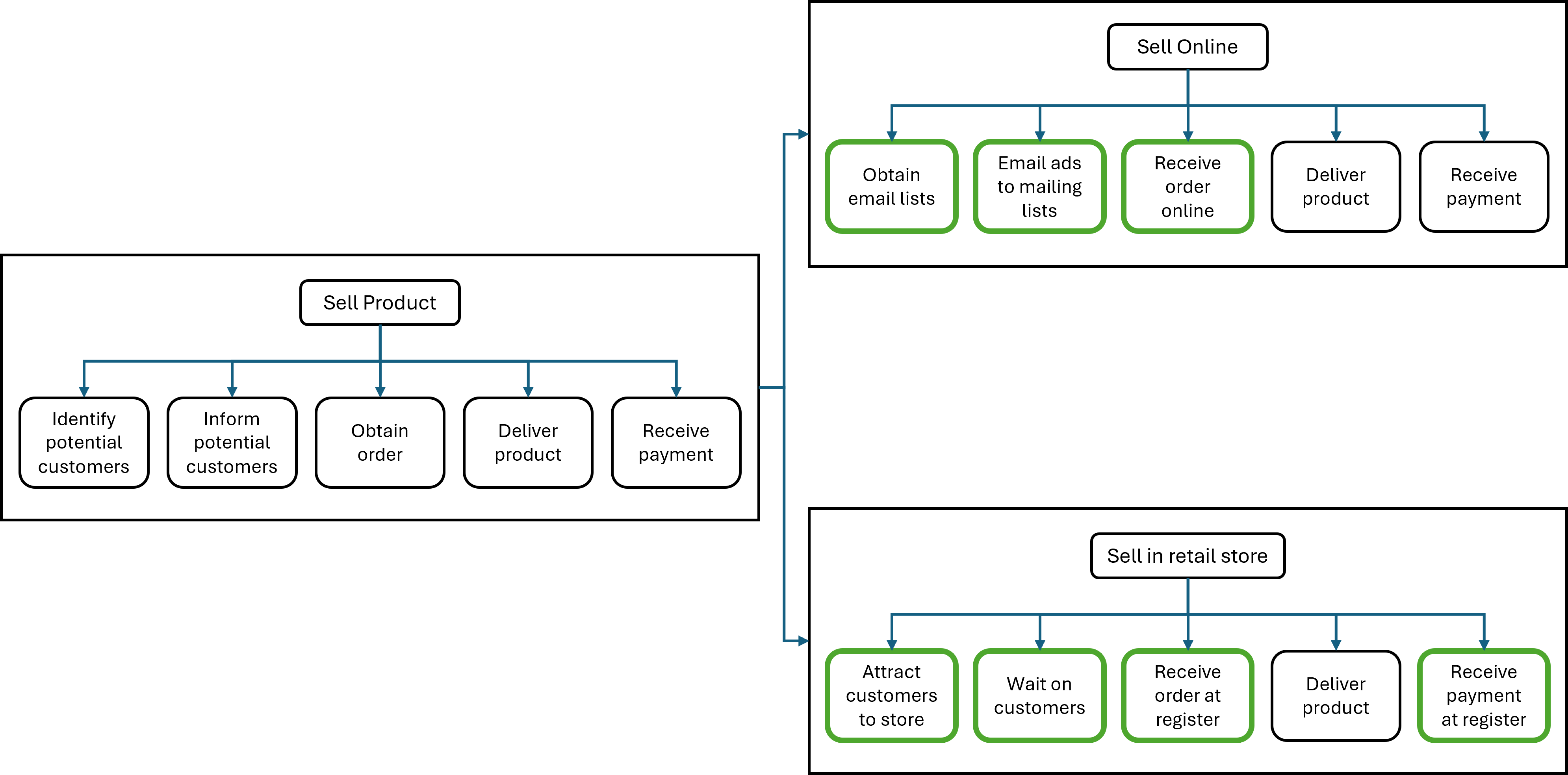}
    \caption{
        \textbf{``Sell online'' and ``Sell in retail store'' are specializations of the generic sales process ``Sell product.''}
        Sub-activities that are different from those in the generic process are shaded. Adapted from \cite{malone_tools_1999}.
    }
    \label{fig-processhandbook}
\end{figure}

A key feature of this kind of ontology is that the properties of a given type of object are \textit{inherited} by the more specialized types of that object except when they are explicitly overridden at the specialized level. For example, \autoref{fig-processhandbook} shows how the five parts (or sub-activities) of the activity ``Sell product'' are inherited by two specialized types (or ``specializations'') of that activity. In this case, several of the sub-activities are overridden in the two specializations, and the sub-activities are replaced by their own specializations (e.g., ``Identify potential customers'' is overridden by ``Obtain email lists'' in ``Sell by email'' and ``Attract customers to store'' in ``Sell in retail store''). For simplicity, we often refer to the specializations of an activity as its ``children'' and the generalizations of an activity as its ``parents.''

In future work on this ontology, we plan to add sub-activities like those shown in~\autoref{fig-processhandbook}, and we expect this to be helpful, not just in identifying where AI can be used, but also in answering broader questions such as how work processes are likely to change to take advantage of AI capabilities. In the work described in this paper, however, we focus only on the inheritance of the property of \textit{AI applicability}\textemdash  whether the activity can be performed using AI.

\subsubsection{Multiple inheritance in ontologies}
\label{sec:background:multiple}

Many ontologies (including the one we use here) include \textit{multiple inheritance}, where an activity can have multiple parents. For example, the activity ``Talk'' can be classified as a specialization (or child) of ``Transfer information,'' since it usually involves transferring information to another person. But it can also be classified as a child of ``Create information'' since it also involves creating information. If an activity has multiple parents, and the parents have different properties, then the properties of at least one of the parents must be overridden in the child activity.

And, as will happen multiple times in this paper, when characteristics like counts of AI applications are aggregated from children up to their ancestors, these \textit{aggregated counts} remove double counting at the levels where the ancestral lines come together. 


\subsection{Creating a large-scale ontology of work activities}
\label{sec:method:construction}

\subsubsection{Design principles}

To create an ontology of work activities, we began with the very comprehensive set of activities in the O*NET database (version 29.1, released November 2024). But we worked to reorganize these activities according to three desirable properties that are not always present in O*NET: \textit{deep similarity}, \textit{atomicity}, and \textit{specificity}. 

\paragraph{Deep similarity}

The first and most important of these properties is \textit{deep similarity}. The biological taxonomy of species of living things (e.g., plants, animals, mammals, birds, etc.) provides a good example of this property. Linnaeus and others who developed this taxonomy did not primarily classify living things according to superficial characteristics such as their color and size, but according to deep characteristics such as their structure (e.g., bones and organs) and function (e.g., warm-bloodedness and live birth)~\cite{mayr_principles_1969,wiley_phylogenetics_2011,linnaeus_systema_1758}.

Similarly, in our ontology, we have classified activities in terms of deep similarities in their structure and function~\cite{malone_tools_1999}. For example, we group activities by their functional outcomes (e.g., creating, modifying, storing, and transferring) and by the type of entity they act upon (information, physical objects, or other actors), rather than on more superficial characteristics such as the industry or geographical location of the work. This matters because functionally similar activities are more likely to have similar \textit{AI applicability}.  

In this proof-of-concept version of our ontology of work activities, we based these classifications on the meanings of the words (especially the verbs) describing the activities. To do this, we used the synonym and hypernym-hyponym relationships represented in the commonly used WordNet lexical database~\cite{miller_wordnet_1995,fellbaum_wordnet_1998} combined with our human editorial judgments to classify deeply similar activities together. Hypernym-hyponym relationships are the semantic relations between a more generic term (hypernym) and a more specific term (hyponym). For example, the verbs ``walk'' and ``run'' are \textit{hyponyms} of ``move,'' which is in turn considered their \textit{hypernym}. Using this approach, for instance, the activity ``Connect physical objects'' includes as subtypes (or what we call \textit{specializations}) activities with names that involve physical objects and that start with ``Connect'' or its synonyms like ``Link,'' such as ``Connect cables'' or ``Link chains.'' It also includes as specializations activities that involve physical objects and that start with hyponyms of ``Connect,'' such as ``Adhere,'' ``Clamp,'' ``Sew,'' and more. In turn, ``Connect physical objects'' would be considered what we call a \textit{generalization} of these activities.



\paragraph{Atomicity}

The second desirable property we designed our ontology to preserve is \textit{atomicity}. This is an important limitation in O*NET because the lowest level activities in O*NET, called ``tasks,'' are mostly compound descriptions of multiple activities in a job role. For example, one such activity is ``Acquire, distribute, and store supplies.'' In our ontology, we decomposed this compound task into three atomic activities, represented as verb-object pairs: ``Acquire supplies,'' ``Distribute supplies,'' and ``Store supplies,'' and each of these atomic activities is classified in a different place in the ontology. Thus, we use the word ``atomic'' here to mean a single activity that captures one specific functional operation and that is usually represented as a verb-object pair. We do not, however, mean that ``atomic'' activities are uniquely correct or irreducible descriptions of work activities, since work activities can often be decomposed in many different ways. 

\paragraph{Specificity}

The final desirable property is \textit{specificity}. By specificity, we mean that (a) a given concept is represented by the same name everywhere it occurs, and (b) different concepts have different names. This means that each node in the ontology corresponds to a single, distinct activity, and our labels make those distinctions explicit. For example, we distinguish ``Perform (artistic)'' from ``Perform (action)'' and ``Load (digital)'' from ``Load (physical)'' rather than using the same surface verb for multiple meanings, whereas O*NET does not make this distinction. (For more examples, see Appendix~\ref{ap:polysemous-activities})

\subsubsection{Construction Process}

\begin{figure}[!htbp]
    \centering
    \includegraphics[width=1\textwidth]{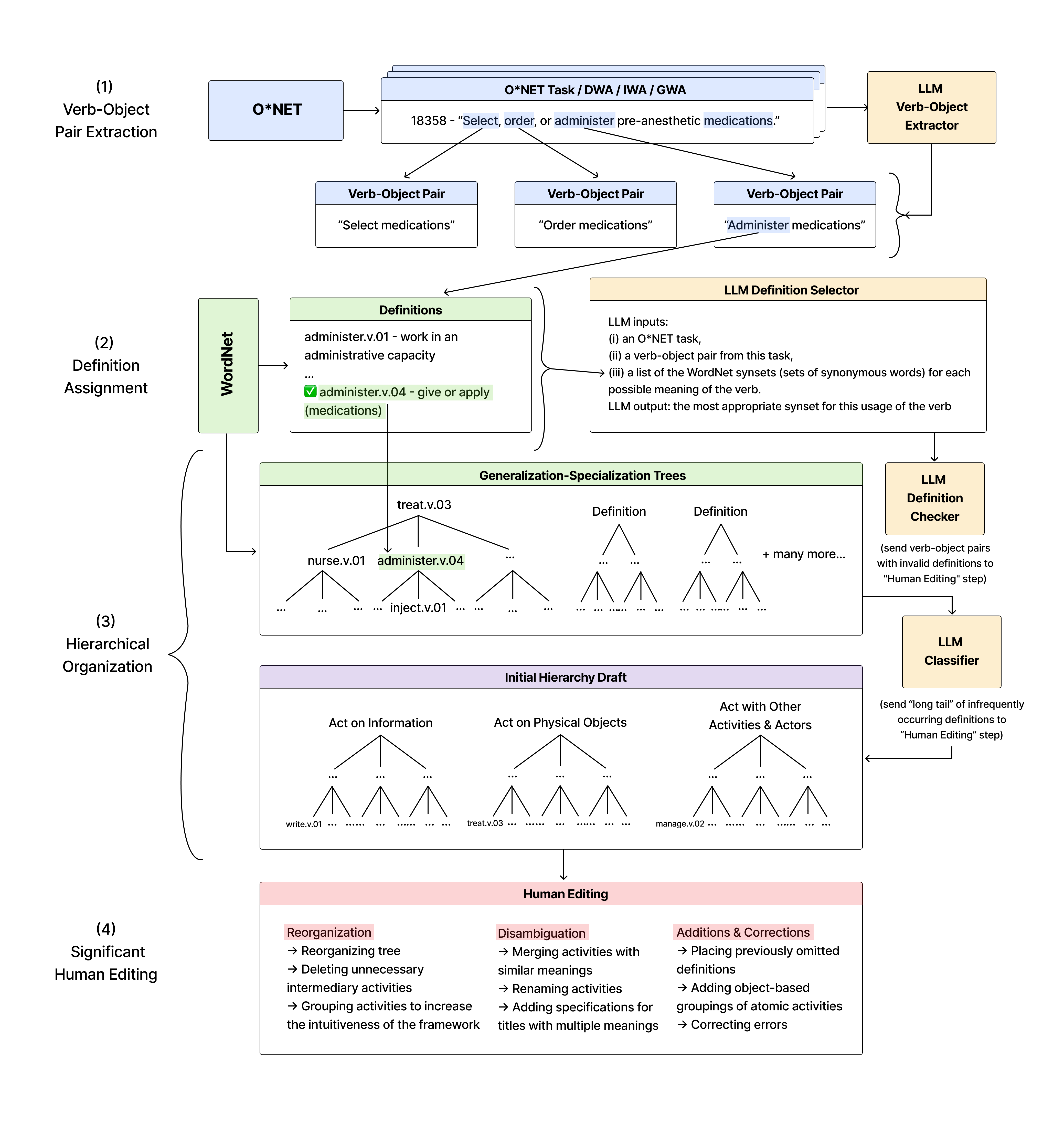}
    \caption{
        \textbf{Flow chart demonstrating the ontology construction process, including steps:} (1) extracting verb-object pairs from O*NET items (to achieve \textit{atomicity}), (2) assigning appropriate definitions from WordNet for each verb in context (to achieve \textit{specificity}), (3) organizing activities into hierarchical structure using WordNet’s ``hypernym'' (generalization) trees, and (4) significant human editing to improve the structure (to achieve \textit{deep similarity}).
    }
    \label{fig-constructionflow}
\end{figure}

To construct the ontology, we explored a number of different approaches\textemdash including using LLMs only, using humans only, and using other knowledge sources for reference. (We elaborate on these explorations in Appendix~\ref{ap:initial-explorations}). We then combined elements from the approaches in our explorations to develop an initial version of the ontology. Finally, we reviewed, edited, and expanded the initial ontology. \autoref{fig-constructionflow} provides a full overview of the construction process.



\paragraph{Step 1: Semi-automated construction of initial ontology}

After initial explorations, we formulated a hybrid approach using a combination of top-down and bottom-up methods, as well as a combination of automated reasoning and human judgment (see~\autoref{fig-constructionflow}).

\begin{figure}[htbp]
    \centering
    \includegraphics[width=1\textwidth]{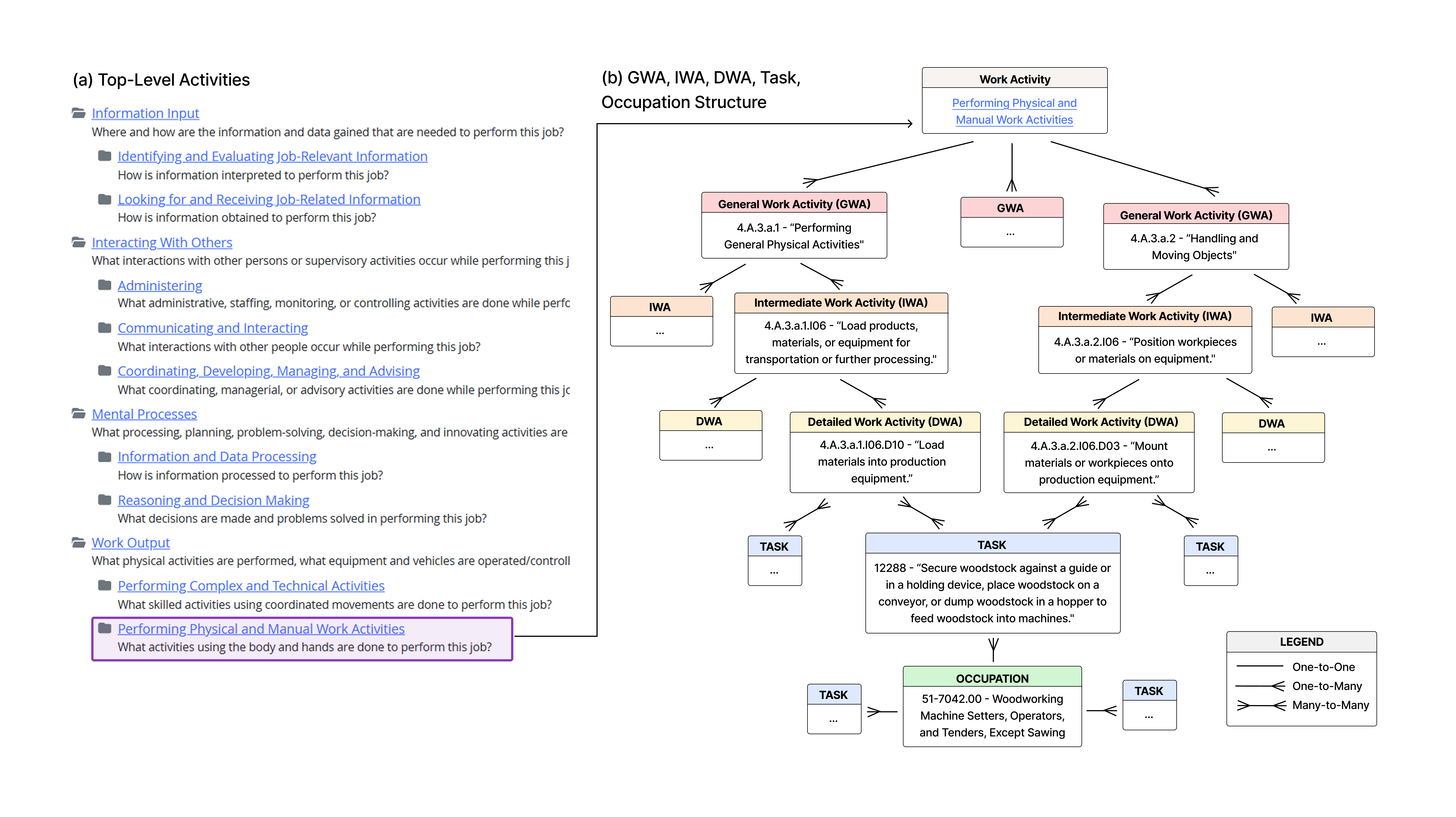}
    \caption{
        \textbf{Overview of the O*NET hierarchy structure with an example of a ``woodworking'' occupation and its associated GWAs, IWAs, DWAs, and tasks.}
        The top-level activities in (a) can be found at \url{https://www.onetonline.org/find/descriptor/browse/4.A.}
    }
    \label{fig-onetstructure}
\end{figure}

In the bottom-up approach, we started with decomposing activities included in the O*NET database, which contains approximately 18{,}000 tasks that get aggregated successively into 2{,}082 Detailed Work Activities (DWAs), 332 Intermediate Work Activities (IWAs), and 41 General Work Activities (GWAs)~\cite{ONET_Resource_Center_2016}. These GWAs are then grouped into four major categories: Information Input, Interacting with Others, Mental Processes, and Work Output. (See~\autoref{fig-onetstructure} for an example of the relationships between tasks, GWAs, IWAs, and DWAs.) 

These DWAs, IWAs, GWAs, and tasks are often compound descriptions, so we used LLM reasoning to break them into atomic activities, the smallest units of work described, which we represent as verb-object pairs. For instance, the task ``Acquire, distribute and store supplies,'' was decomposed into three verb-object pairs: ``acquire supplies,'' ``distribute supplies,'' and ``store supplies.'' This resulted in 40,825 verb-object pairs, of which less than half (15,989) were unique, indicating many O*NET tasks share atomic activities.

Then we used LLM reasoning to select the most appropriate definition for each verb in the verb-object pairs from the available definitions in WordNet, a widely used linguistic database that groups words by their meaning and maps the semantic relationships between them \cite{princeton_university_wordnet_2026}. WordNet at times contains definitions with significant similarity\textemdash for example, WordNet has both definitions ``engage in drawing'' and ``represent by making a drawing of, as with a pencil, chalk, etc. on a surface'' for the verb ``draw.'' Due to the lack of clear distinction betewen the definitions, the LLM would sometimes choose one and other times choose the other for the same context. Thus to reduce the number of potential definitions, we presented the LLM with only the definition with the most hyponyms for each ``supersense'' (a broad semantic category in WordNet, such as verb.creation, verb.motion, verb.social, etc. \cite{noauthor_lexnames5wn_nodate}). We used the number of hyponyms as a proxy for generality, preferring broader verb senses that are more likely to anchor the hierarchy. 

Sometimes, the chosen definition was incorrect, so we used another LLM that checked whether the chosen definition matched the context given. We omitted invalid definitions from the initial version of the hierarchy. We manually revised and re-incorporated such instances in the Revision phase.

We used the selected meanings to merge the verb-object pairs into groups where the verbs all have the same meaning and the objects are identical. We call these groups ``atomic activities.'' To make it easier for LLM and human editing, we also chose to first work only with meanings that had more than 10 occurrences in O*NET tasks, excluding the so-called ``long tail'' of meanings. These we reincorporated in the Revision phase. We then used the hypernym relationships (akin to generalizations) in WordNet to organize the atomic activities into 60 hierarchical trees with the more specialized verbs at the bottom of each tree and the more general verbs at the top of each tree. (See Appendix~\ref{ap:wordnet-hierarchy-trees} for a full list of the 60 general verbs.)

Next, we used a top-down approach to create a top-level hierarchy into which all these trees of atomic activities were classified. At the highest level, we adopted a tripartite distinction between acting on information (which we nickname ``Think''), acting on physical objects (which we nickname ``Do''), and interacting with other actors (which we nickname ``Interact''), commonly used in work studies~\cite{fine_introduction_1971}. Then, we used LLM reasoning to classify the hypernym trees from the previous stage into this top-level framework. 

\paragraph{Step 2: Human and LLM editing to revise ontology}

After the construction of the initial ontology, we refined it through a combination of automated checks and human review. First, two authors of this paper acted as human editors and revised the classifications, adding a second-level framework of Create, Modify, Transfer, and Store underneath each branch within the Do/Think/Interact framework. Our human editors also made other edits, such as renaming activities, deleting unnecessary intermediary activities, merging activities with similar meanings, and grouping activities to increase the intuitiveness of the framework. We opted to use WordNet’s hierarchy as a linguistic starting point for our classifications, but we made significant changes because our principle of deep similarity aims to capture functional inheritance (inheritance of properties), which sometimes diverges from WordNet’s lexical relationships. For instance, WordNet places ``Type'' (as in typing on a keyboard) under ``Write'' based on lexical relationships, but functionally, typing is a physical manipulation activity that may or may not involve writing; one can type numbers, code, or random characters. We moved such activities to reflect functional rather than lexical inheritance.

Next, we added explicit specifications to 215 activities using the same verb but different meanings (e.g., Perform (artistic) vs. Perform (action), Load (physical) vs. Load (digital)). (see Appendix~\ref{ap:polysemous-activities} for more detail). We then placed 644 previously omitted meanings from both invalid definitions and the long tail of infrequently occurring O*NET tasks. To do this, we used LLM-suggested placements reviewed by human editors. (This method was developed iteratively, with details in Appendix~\ref{ap:synset-placement}).

To address ambiguities in WordNet definitions\textemdash especially verbs spanning both physical and informational uses\textemdash we iteratively developed a framework for classifying direct objects into four categories: Physical Object, Information, Actor, and Activity (e.g., ``Improve activity''). Then we used this framework and a combination of LLM reasoning and human judgment to resolve 615 inconsistencies between verb locations in the ontology and object types. For example, if the activity ``Cut video'' is placed under the ``Act on physical objects $>$ ... Cut (physical)'' branch, while ``video'' is labeled as Information, this constitutes an inconsistency that can be resolved by placing it under the mirror branch with the matching object, ``Act on information $>$ ... Cut (information)''. The principles we developed to define these categories are detailed in Appendix~\ref{ap:direct-object-principles}. The method to resolve inconsistencies is detailed in Appendix~\ref{ap:inconsistency-resolution}. 

Finally, we used these noun classifications to introduce object-based intermediate activities between generic activities and atomic activities (e.g., adding the intermediate activity ``Observe Physical Object'' between the generic activity ``Observe'' and the atomic activity ``Observe Equipment''), improving clarity and structural consistency.

In summary, using a combination of automated LLM reasoning, WordNet, and editorial judgments by authors of this paper, we produced an ontology with atomicity, specificity, and deep similarity that we hope can serve as a foundational analysis framework for future studies. 

This process produced an ontology of work activities with 1,113 generic activities and 15,989 unique atomic activities covering all 20,950 O*NET tasks, DWAs, IWAs, and GWAs. Generic activities are usually verbs, like ``Decide,'' ``Assemble,'' ``Clean,'' ``Communicate,'' sometimes with broad object categories, like ``Create information'' or ``Create physical objects.'' Atomic activities are more specific verb-object pairs (like ``Collect fees'') extracted from O*NET tasks. The fact that there are fewer unique atomic tasks in total than O*NET tasks also means that many atomic tasks are shared across O*NET tasks, showing the compression power of our ontology.

\autoref{fig-ontologyexcerpt} shows a sample of the ontology contents. At the far left is the most generic activity type of all, ``Act,'' followed by a ``Collection'' of related activities that are all answers to the question ``Act on what?'' (e.g., ``Act on information,'' ``Act on physical objects,'' ``Act with other activities or actors''). Beneath those activities are specializations of each (e.g. ``Create information,'' ``Modify information,'' etc., are specializations of ``Act on information''). To the right is an expansion of one of these activities, ``Create information,'' which includes more specializations, collections, and finally atomic activities and their associated O*NET tasks.

\begin{table}[htbp]
    \centering
    \caption{
        \textbf{Counts and examples across different levels of the ontology.}
        Note that the Total row is not the direct sum of the rows above, but rather counts only once the activities that exist at multiple levels.
    }
    \label{tab:activity_levels}
    \small
\begin{tabular*}{\textwidth}{@{\extracolsep\fill}p{2.8cm}p{1.3cm}p{5.8cm}p{1.8cm}}
\toprule
\textbf{Type of activity} 
& \textbf{Activity level} 
& \textbf{Examples} 
& \textbf{Number of activities} \\
\midrule

\multirow{5}{*}{Generic activities} 
& 1 
& Act 
& 1 \\
\cmidrule(lr){2-4}

& 2 
& Act on information, Act on physical objects, Act with other actors and activities 
& 3 \\
\cmidrule(lr){2-4}

& 3 
& Create information, Modify information, Transfer information, Store information, Create physical objects, Modify physical objects, \ldots 
& 10 \\
\cmidrule(lr){2-4}

& 4 
& Decide, Calculate, Plan, Fabricate, Buy, Improve, \ldots 
& 69 \\
\cmidrule(lr){2-4}

& Cumulative 1 - n-3
& Select, Count, Budget, Cast, Rent, Optimize, \ldots 
& 1,113 \\
\cmidrule(lr){2-4}

& $n-2$ 
& Select (Information), Select (Actor), Budget (Activity), Cast (Physical Object), Optimize (Information), \ldots 
& 1,551 \\
\midrule

Atomic activities 
& $n-1$ 
& Select method, Select personnel, Budget work, Cast mold, Optimize cost, \ldots 
& 15,989 \\
\midrule

O*NET items 
& $n$ 
& (16122) Select methods, techniques, or criteria for data warehousing evaluative procedures; \newline
(4.A.3.a.2.I17.D01) Cast molds of patient anatomies to create medical or dental devices 
& 20,950 \\

\midrule
\textbf{Total (excluding collections)} 
& --- 
& --- 
& 39,603 \\
\bottomrule
\end{tabular*}

\end{table}

\autoref{tab:activity_levels} shows more detailed statistics about the contents of the final ontology. The first three levels are broad categories (e.g., ``Act on information,'' ``Modify physical objects'') into which all the more specialized activities can be classified. At these levels, there may be relatively few characteristics that are inherited by their descendants. Level 4, however, is the first level where we expect relatively detailed sets of properties to characterize the ``family trees'' headed by the 69 activities at this level. There are a total of 2,664 generic activities from level 1 to the last level before atomic activities. Approximately 34.1\% of generic activities fall under ``Act on information,'' 33.5\% under ``Act on physical objects,'' and 32.4\% under ``Act with other activities and actors.''

    
    

In terms of the depth of the ontology, the median path length from the root activity (``Act'') to leaf generic activities is 6. Below this, there are 3 more levels: noun-grouped activities, atomic activities, and O*NET tasks, making the median total path length 9. The minimum path length to O*NET tasks is 6 and the maximum is 14. 

We also track 22 activities that multiply inherit from more than one top-level domain (``Think'' vs. ``Do'' vs. ``Interact''), indicating many activities (e.g., ``Talk'', ``Greet'', ``Gather'') cut across informational, physical, and interpersonal domains (see Appendix \ref{ap:taaft:multiple_inheritance}). 

\subsubsection{Rationale for Choice of Methods}

While we do not claim our methods are optimized, we had several reasons for choosing them over other potential methods. First, we chose not to create the hierarchy of activities using semantic clustering, because we wanted the units of our ontology to represent discrete word \textit{senses}, not approximate similarities in a continuous embedding space. Second, we chose not to use a purely LLM-generated taxonomy and instead used a Human-in-the-Loop (HITL) approach. While recent work suggests LLMs can sometimes outperform crowd workers for text annotation~\cite{gilardi_chatgpt_2023}, our experience suggests that the LLM's tendency to hallucinate often results in taxonomies in which human experts see many limitations. Based on this, we wanted to investigate the potential benefits of using human experts to verify that the hierarchical relationships represent actual functional inheritance rather than mere keyword association; in other words, we distinguish ``is a kind of'' relationships from ``often happens with'' relationships.

We also utilized WordNet as the basis for our automated ontology creation, rather than purely LLM-based classifications. This was primarily to ensure deterministic results from the automated part of our classification process. While LLMs excel at reasoning, their outputs can be non-deterministic and hallucinate novel categories. Determinism matters for ontology construction because the ontology will serve as a stable reference for classifying evolving datasets over time. If the ontology itself were generated non-deterministically, different runs could produce different structures, making longitudinal comparisons impossible. As a commonly used and well-respected lexical database curated by human experts, WordNet provided us with a deterministic and meaning-grounded starting point.

Methodologically, our effort brings the precision of the field of semantics to the characterization of work activities. In contrast to existing task datasets, our ontology is characterized by deep similarity, atomicity, and specificity. These features abstract beyond task descriptions that vary over time, allowing our ontology to offer a durable semantic framework that can anchor evolving datasets like O*NET.

\subsubsection{Demonstration of Usefulness}

To demonstrate the usefulness of our approach, this section provides concrete examples of benefits from the three key features described above: atomicity, specificity, and deep similarity.

\begin{figure}[h]
    \centering
    \includegraphics[width=0.75\textwidth]{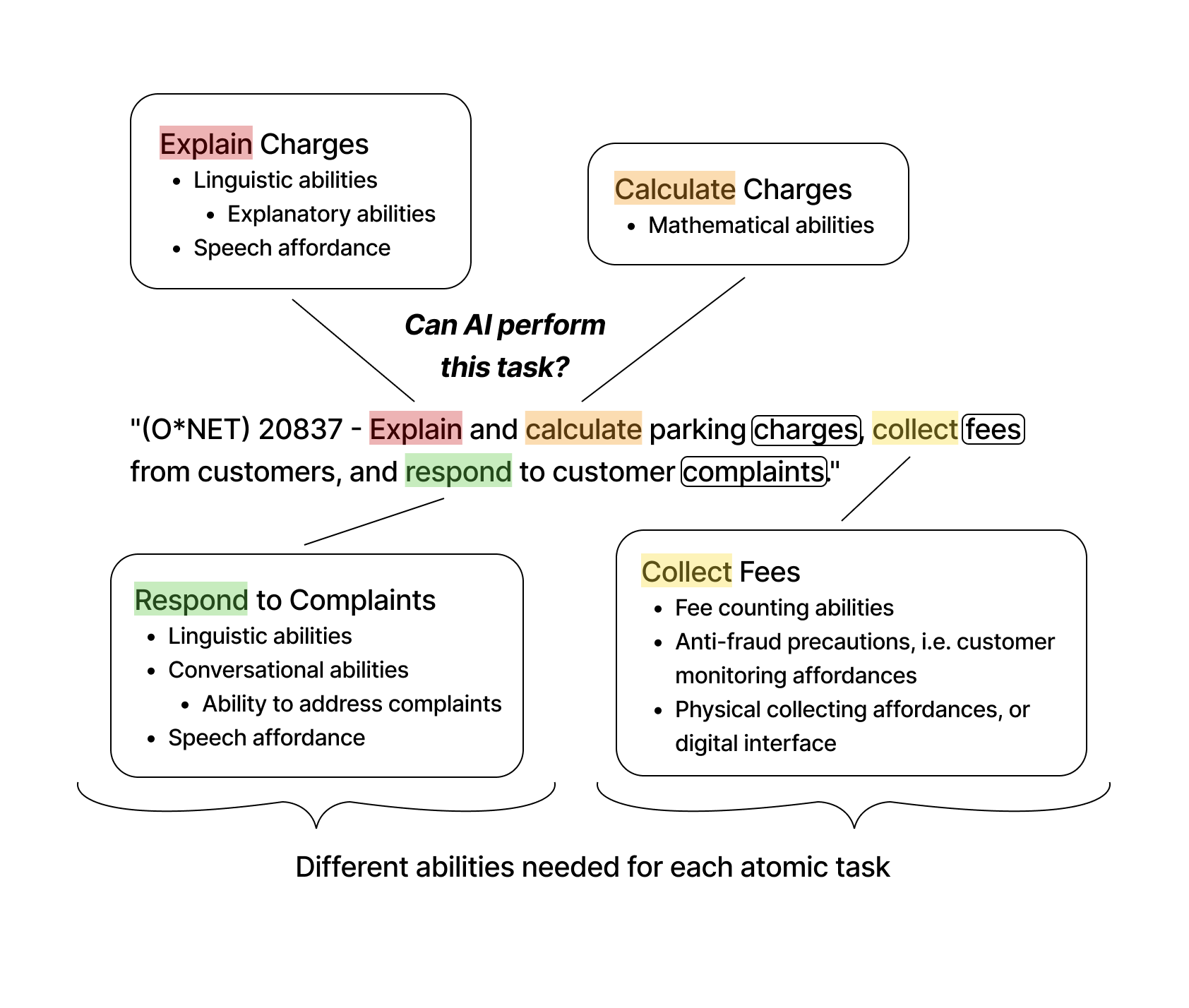}
    \caption{
        \textbf{Example of how O*NET tasks break down into atomic tasks.}
        Demonstration of how this breakdown helps with the analysis of AI applicability, because each atomic task requires different abilities.
    }
    \label{fig-atomicity}
\end{figure}

First, the \textit{atomicity} of our ontology enables analysis of tasks on their own, as opposed to in compound forms like the instances found in O*NET. \autoref{fig-atomicity} shows a demonstration of this principle in practice by using an O*NET task that includes 4 distinct atomic activities. Previous AI impact analyses using O*NET tasks~\cite{eloundou_gpts_2023, handa_which_2025, tomlinson_working_2025} often clump together the atomic activities in an O*NET task. They also often compensate for this by, for example, taking a measurement for an O*NET task and arbitrarily dividing it equally among all the O*NET DWAs the task includes~\cite{handa_which_2025}. As~\autoref{fig-atomicity} illustrates, the fine-grained focus provided by atomic activities can give much more precision to many similar analyses. 

\begin{figure}[h]
    \centering
    \includegraphics[width=0.8\textwidth]{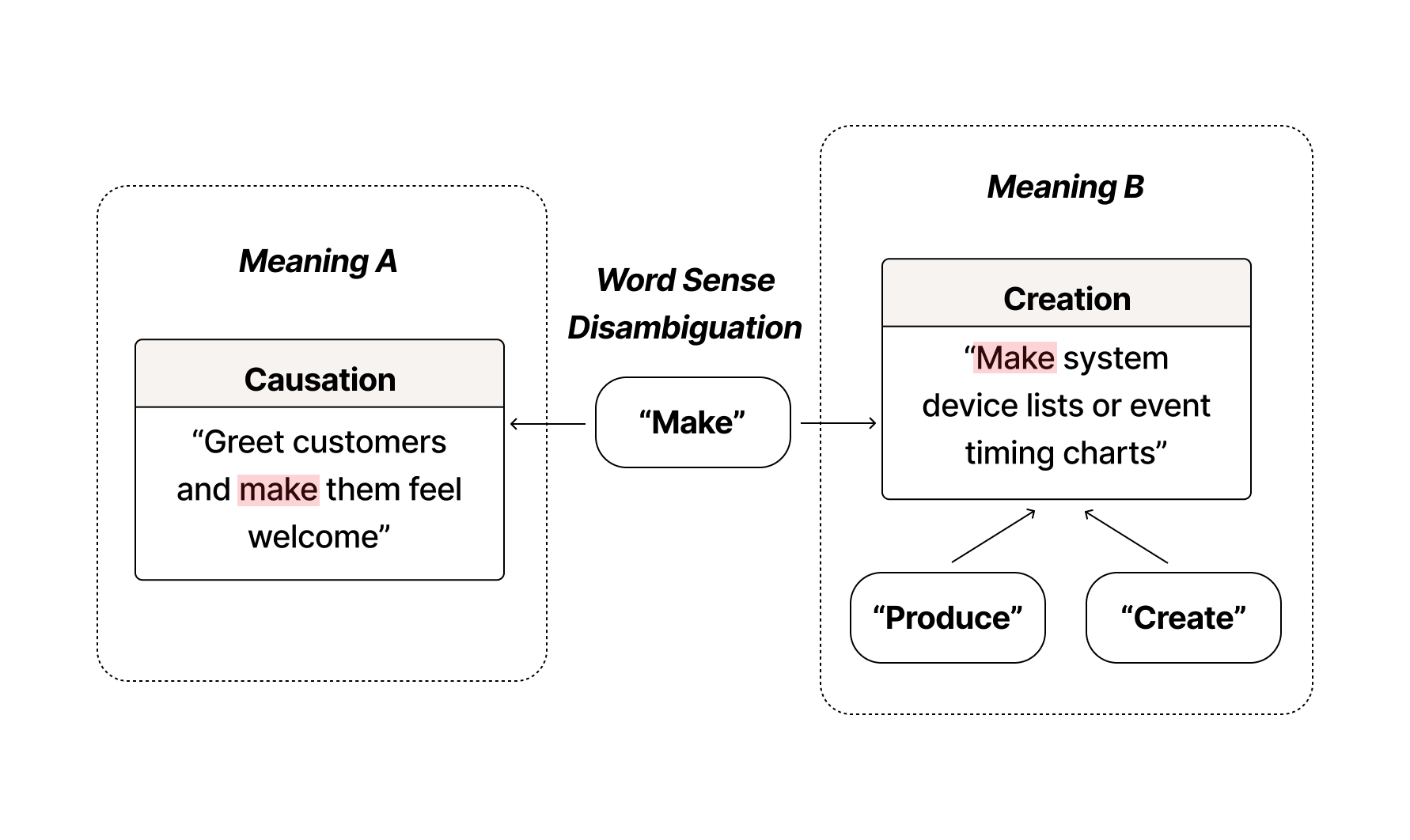}
    \caption{
        \textbf{Demonstration of the importance of \textit{specificity},} achieved using word sense disambiguation, in the case of ``Make'', with distinct senses meaning creation and causation; ``Produce'' and ``Create'' also are synonyms of the Creation meaning.
    }
    \label{fig-specificity}
\end{figure}

Second, the \textit{specificity} of our ontology allows precise expression of activities without being confused by synonyms or multiple meanings for a single word (polysemy). Ultimately, ontologies should be grounded in \textit{meanings} rather than \textit{words}. \autoref{fig-specificity} shows an example of this. Consider the verb \textit{make}. In O*NET task 16513 (``Make system device lists or event timing charts''), ``make'' is synonymous with verbs such as produce and create used in other tasks describing the same activity (e.g., producing graphics or creating diagrams). However, \textit{make} is also polysemous. In tasks such as ``Make customers feel welcome,'' ``make'' expresses causation rather than creation. Without what linguists call ``word sense disambiguation'' (WSD),  these distinct meanings of \textit{make} could either be conflated (grouping causative actions with artifact creation) or fragmented (treating synonymous expressions as separate activities). Our ontology resolves these issues by mapping synonyms to a shared atomic activity while separating the distinct senses of verbs that have multiple meanings, enabling precise classification and downstream reasoning. For example, an AI tool capable of automatically generating diagrams or system documentation can be correctly associated with occupational tasks that include \textit{produce}, \textit{create}, \textit{make} in the creative sense, but not with causative uses of \textit{make} that involve directing or compelling human behavior.
 
Third, the \textit{deep similarity} of our ontology enables aggregating and propagating insights. When analyzing tens of thousands of work activities, hierarchical frameworks enable simplification of understanding and high-level insights. Many papers analyzing work impact recognize the value of higher-level frameworks\textemdash for example, \cite{handa_which_2025} develops a custom clustering-based hierarchy to classify the O*NET tasks associated with prompts to the 
Claude LLM. O*NET itself has a hierarchical structure with tasks, DWAs, IWAs, and GWAs. However, we propose that such frameworks need deep similarity\textemdash that is, child categories should share, or ``inherit,'' properties from parent categories. Properties that can be inherited include, for example, the parts or steps of an activity and the potential applicability of AI to the activity. Unlike clustering-based hierarchies that group activities by statistical co-occurrence, our ontology groups activities by functional inheritance. 



To illustrate why deep similarity matters for AI analysis, consider the activity family of \textit{inspection}\textemdash examining equipment, materials, or operations to assess their condition or compliance. Most readers would expect these closely related tasks to be grouped together: they share a common cognitive structure (perceive, compare to standard, judge) and, crucially, similar AI applicability (computer vision systems that can inspect equipment for defects can likely also inspect materials or operations for defects). In our ontology, all inspection activities are descendants of the generic activity ``Inspect,'' enabling this kind of cross-activity reasoning.

\begin{table}[h]
    \centering
    \caption{
        \textbf{O*NET scatters functionally similar inspection activities across three separate GWA branches.}
        All six DWAs share the core cognitive operation of inspection (perceive, compare, judge), yet O*NET's hierarchy places them under different top-level categories: two under ``Inspecting Equipment,'' three under ``Monitor Processes,'' and one under ``Handling and Moving Objects.'' In our ontology, all six activities inherit from a common ``Inspect'' ancestor, enabling AI applicability to propagate across the entire inspection family.
    }
    \footnotesize
    \begin{tabularx}{\textwidth}{@{}p{2.2cm} X p{2.8cm} X@{}}
    \toprule
    \textbf{GWA Branch} & \textbf{GWA Title} & \textbf{DWA ID} & \textbf{DWA Title (Inspection Activity)} \\
    \midrule
    \multicolumn{4}{@{}l}{\textit{Branch 1: Information Input $\rightarrow$ Inspecting Equipment, Structures, or Material}} \\
    4.A.1.b.2 & Inspecting Equipment, Structures, or Material & 4.A.1.b.2.I07.D18 & Inspect equipment to ensure proper functioning \\
    4.A.1.b.2 & Inspecting Equipment, Structures, or Material & 4.A.1.b.2.I10.D03 & Inspect shipments to ensure correct order fulfillment \\
    \midrule
    \multicolumn{4}{@{}l}{\textit{Branch 2: Information Input $\rightarrow$ Monitor Processes, Materials, or Surroundings}} \\
    4.A.1.a.2 & Monitor Processes, Materials, or Surroundings & 4.A.1.a.2.I02.D09 & Inspect operational processes \\
    4.A.1.a.2 & Monitor Processes, Materials, or Surroundings & 4.A.1.a.2.I11.D02 & Inspect products or operations to ensure standards are met \\
    4.A.1.a.2 & Monitor Processes, Materials, or Surroundings & 4.A.1.a.2.I09.D06 & Inspect condition of natural environments \\
    \midrule
    \multicolumn{4}{@{}l}{\textit{Branch 3: Work Output $\rightarrow$ Handling and Moving Objects}} \\
    4.A.3.a.2 & Handling and Moving Objects & 4.A.3.a.2.I19.D03 & Disassemble equipment to inspect for deficiencies \\
    \bottomrule
    \end{tabularx}
    \label{tab:Example_of_ONET_scattering_Inspect}
\end{table}

In O*NET's Work Activities hierarchy, however, functionally similar inspection tasks are scattered across three different top-level Generalized Work Activities (GWA) branches. As shown in \autoref{tab:Example_of_ONET_scattering_Inspect}, ``Inspect equipment to ensure proper functioning'' falls under GWA 4.A.1.b.2 (\textit{Inspecting Equipment, Structures, or Material}), while ``Inspect operational processes'' falls under GWA 4.A.1.a.2 (\textit{Monitor Processes, Materials, or Surroundings})\textemdash a different GWA branch entirely. Even more surprisingly, ``Disassemble equipment to inspect for deficiencies'' is classified under GWA 4.A.3.a.2 (\textit{Handling and Moving Objects}), grouping it with material handling rather than with other inspection activities.

This scattering has practical consequences for AI impact analysis. A researcher using O*NET's hierarchy to assess whether AI can automate inspection tasks would need to separately evaluate three GWA branches and somehow recognize that they share an underlying capability. Our ontology resolves this by grouping all inspection activities under a single ``Inspect'' ancestor, so that evidence of AI applicability to any inspection task can propagate to related tasks through inheritance. When we find that computer vision systems can ``Inspect equipment,'' we have principled grounds to hypothesize they may also support ``Inspect shipments'' or ``Inspect operational processes''\textemdash a form of systematic capability transfer that O*NET's fragmented structure cannot support.



\subsection{Software platform for creating an ontology}

\label{sec:method:platform}

\begin{figure}[htbp]
    \centering
    \includegraphics[width=\textwidth]{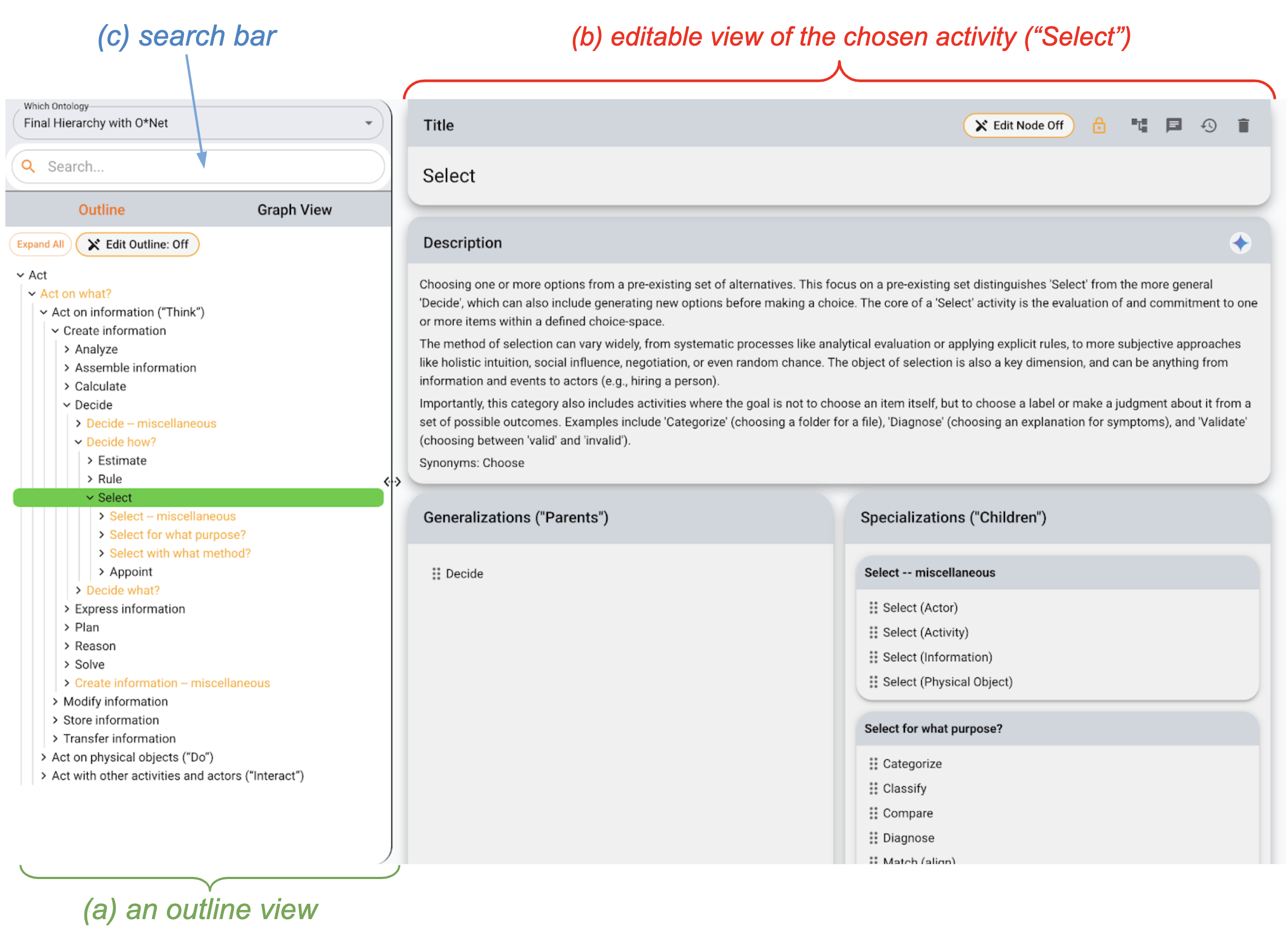}
    \caption{
        \textbf{User interface for ontology platform,} including (a) an outline view of the specialization hierarchy, starting with the most generic activity type called ``Act,'' (b) an editable view of the chosen activity (``Select''), including properties such as Title, Description, Generalizations, and Specializations, and (c) a search bar to find activities by keywords or semantically similar phrases.
    }
    \label{fig:ontology_platform_UI}
\end{figure}

To create and maintain this ontology at the scale of tens of thousands of activities and relationships, we developed a dedicated platform for collaborative ontology editing. \autoref{fig:ontology_platform_UI} shows an example of the interface, which supports three core functionalities: (i) structured browsing of the specialization hierarchy of activities, (ii) inspecting and editing an activity's title, definition, synonyms, and links (generalizations, specializations, and parts), and (iii) searching using a combination of keyword search and semantic (meaning-based) retrieval (Appendix~\ref{ap:platform}). 

The platform provides guardrails to support coherent, auditable editing at scale. For example, it blocks edits that would disconnect (or ``orphan'') an activity, flags near-duplicate activities for review, and maintains node-level change histories with attribution to the people who made changes. For downstream analyses and automated classification, it exports versioned JSON snapshots so evaluations and scripts reference a fixed ontology instance.

In addition to being essential for creating the ontology, this platform also served as a primary instrument for the human evaluation of automated classifications described in Section \ref{sec:method:pipeline:automated}. A key feature of the platform for this purpose was the hybrid search engine described above that combines keyword search with semantic retrieval. This helps humans who are classifying activities (``human annotators'') locate candidate activities based on either semantic meaning or exact wording. 


\subsection{Data sources about AI uses}
For our analysis, we examine two representative datasets about AI usage. The first dataset is a catalog of AI software applications from a website called ``There's an AI for That\textsuperscript{®}'' (TAAFT)~\cite{nedelcu_theres_2026}. These applications act on information, not on physical objects. The second dataset, from the International Federation of Robotics (IFR) \textit{World Robotics} reports~\cite{IFR2025Service, IFR2025Industrial}, describes robotic systems that act on physical objects. Both types of systems can also act on or with other actors and activities. Together, these two datasets provide a comprehensive overview of contemporary AI usage.

\subsubsection{AI software applications}
\label{ai_software_applications}
``There's an AI for That\textsuperscript{®}'' (TAAFT)~\cite{nedelcu_theres_2026} is a website that provides a comprehensive catalog of AI software applications. The company provided us with a dataset that included descriptions of 13,275 individual AI applications launched between November 2013 and July 2025. This dataset excluded applications listed on the site that (a) were inactive or no longer maintained, (b)  were platform-specific mini-tools developed by TAAFT itself, or (c) involved only custom tools built using ChatGPT (``custom GPTs''), which were considered largely obsolete. 

The included applications span a wide range of tasks and use cases, including, but not limited to, text generation, data analysis, automation, image-related tasks, and domain-specific workflows. Although this is certainly not an exhaustive census of all AI software applications, we believe it is broadly representative of the AI software applications in use today. The dataset includes AI applications deployed on different platforms, including websites, iOS, Android, and Chrome extensions. For each AI application, the dataset includes descriptions, taglines, primary tasks, tags, pricing information (including price, pricing model, and billing frequency), launch dates, and indicators of community engagement such as the number of saves (see Appendix~\ref{ap:taaft:sample} for example data).

\subsubsection{Robotic systems}
\label{robotic_systems}

The International Federation of Robotics (IFR) publishes annual \textit{World Robotics} reports \cite{IFR2025Industrial, IFR2025Service}. These reports comprise the world's most comprehensive dataset on robotic systems. 

IFR defines a robot as a ``programmed actuated mechanism with a degree of autonomy, capable of performing locomotion, manipulation, or positioning'' (adopted by the International Standardization Organization (ISO) \cite{ISO8373_2021}. This definition excludes simple factory machines that repeatedly perform the same movements without autonomous decision-making. IFR classifies robots into two major categories, industrial and service robots. Service robots include professional, medical and consumer robots (for detailed definitions, see Appendix~\ref{ifr taxonomy}). Note that the IFR reports specifically exclude two types of robots: (i) military robots (robots for the defense sector), because they are against the organization's values for peaceful use of robotics; and (ii) autonomous passenger transportation vehicles as part of the automotive industry.

In each major category, IFR segments robotic systems into classes and subclasses.  For example, industrial robots are divided into 6 classes (e.g., ``Welding and soldering'', ``Assembly and disassembly,'' etc.), and these classes are further divided into 35 subclasses (e.g., ``Welding and soldering'' includes the subclasses ``Arc welding,'' ``Spot welding,'' and four others.) Service robots are similarly categorized into classes and subclasses. In all, IFR recognizes 20 classes of robotic systems and 66 subclasses.  \autoref{tab:ifr_industrial}, \autoref{tab:ifr_service_professional}, \autoref{tab:ifr_medical_prof}, and \autoref{tab:ifr_service consumer} in Appendix~\ref{ifr taxonomy} list all the  classes and subclasses defined by IFR.  

We used IFR data on the global number of robot installations in each subclass in 2024, the most recent year for which data is available. In that year, there were a total of 20.8 million total installations of robotic systems, of which 20.1 million were consumer robots, 542K industrial robots, 199K professional service robots, and 17K medical robots.

\subsection{Methods for classifying activities in an ontology}
\label{sec:method:pipeline}

\subsubsection{Automated classification}
\label{sec:method:pipeline:automated}

Since it was not feasible to manually classify the more than 13,000 AI software applications into a deep ontology, we developed an automated classification pipeline that was validated by human annotators. 

\paragraph{Philosophy of classification}

We classified each application according to its primary activity: the essential verb-object activity the application performs or helps perform. To do this, we determine the \textit{most specific ontology activity whose scope fully covers the application's primary activity}. Because many applications are multi-purpose and descriptions can be ambiguous, classification is treated as a subjective judgment based on the rule in the previous sentence.

We consider an automated classification pipeline satisfactory if its human-AI inter-annotator agreement (IAA) is comparable to, or exceeds, the human-human IAA achieved under equivalent instructions. This criterion evaluates consistency with human judgment; it does not establish a single objective ground-truth label for each application.

\paragraph{Automated classification pipeline}

To automatically map the AI software dataset into our ontology, we investigated a number of alternative methods and evaluated three final pipeline variants (see Appendix~\ref{ap:classification} for details and prompt text): 

\begin{itemize}
    \item \textbf{Single Prompt Partial Ontology (SPPO).} We first use semantic embeddings to retrieve a subset of ontology activities ($k\in\{20,50,100\}$) that are semantically similar to the application description and tagline. A single GPT-5.1 prompt then (i) identifies a verb--object pair describing the application's primary activity and (ii) selects the best-matching ontology activity from the retrieved shortlist.
    \item \textbf{Multi-Prompt Partial Ontology (MPPO).} We first prompt GPT-5.1 to extract a clean verb--object description of the application's primary activity. We then retrieve $k\in\{20,50,100\}$ candidate ontology activities using semantic similarity to this verb--object phrase. Finally, a second prompt selects the best-matching ontology activity from the retrieved shortlist.
    \item \textbf{Single Prompt Full Ontology (SPFO).} We cache the full ontology in the model context and use a single GPT-5.1 prompt to identify the application's primary verb--object activity and select the most specific matching ontology node from the full hierarchy, without retrieval.
\end{itemize}

Appendix~\ref{ap:classification} provides additional details about the evolution of these variants, including the prompts and the main tradeoffs we observed across accuracy, specificity, hallucinations, and cost.

Our final pipeline uses the SPFO approach with GPT-5.1 (high-reasoning-effort). By caching the full ontology in the system prompt, this model can traverse the hierarchy rather than rely on a retrieved shortlist. As described in the next subsection, this pipeline had the highest agreement with human annotators, though differences across variants were not statistically significant.

\paragraph{Human evaluation of classifications}
\label{sec:human_evaluation}
To evaluate the classification quality of the three candidate pipeline variants and inform the final model selection, we conducted a small-scale human evaluation. We randomly sampled 30 applications from the full TAAFT dataset. Seven authors of this paper manually classified the primary activity for each of these 30 applications and then compared their classifications with those produced by the three GPT-5.1 model architectures described above.


\textit{Human classification.}
To make their classifications, each human annotator saw the application name, tagline, and description for each application. The human annotators were also given access to the ontology platform, which provided the full ontology (including activity names, descriptions, and synonyms) as well as a search engine for finding relevant ontology activities (see Appendix~\ref{ap:platform}). The human annotation instructions were intentionally aligned with the prompt for the AI models: both required identifying the application’s primary verb-object pair and selecting the most appropriate (i.e., closest/most similar) ontology activity, operationalized as choosing the most specific node whose scope fully covers that pair (see Appendix~\ref{ap:human_model_prompts} for the full instructions provided to human annotators). 

\textit{Human evaluation of model classifications.}
After completing their own classifications, human annotators were shown the classifications generated by the three AI pipeline variants, each accompanied by the model-generated rationale. For each application, participants were asked to consider their own classification and compare it with each AI model’s output. They were also asked to review the model rationales for additional context. Then the human annotators evaluated each model’s classification using one of three options: ``better than my own classification,'' ``equally good as my own classification,'' or ``worse than my own classification.'' To ensure fair comparison, model identities were anonymized as ``LLM1,'' ``LLM2,'' and ``LLM3'' during evaluation. 

\textit{Analysis Methods and Results.}
We report both human-human inter-annotator agreement (IAA) and human-AI IAA following prior work~\cite{tomlinson_working_2025,mccain_how_2025,chatterji_how_2025,patwardhan_gdpval_2025}. Because our task involves classification within a hierarchically structured ontology rather than a flat categorical label space, general-purpose IAA measures are not directly applicable. Instead, we adopt similarity-based IAA formulations with broader applicability for hierarchical structures~\cite{braylan_measuring_2022}. 
Specifically, we use the Wu-Palmer (WuP) network similarity metric~\cite{wu_verb_1994} and weighted Cohen’s $\kappa$, as described in more detail in Appendix~\ref{ap:iaa_detail}. 

As shown in~\autoref{fig:4_1_human_ai_agreement}, all three pipeline variants achieve human-AI IAA levels comparable to human-human IAA. Across both metrics, human-AI IAA is similar to and slightly higher than human-human IAA, indicating that LLMs are capable of closely approximating the subjective judgments made by human classifiers~\cite{he_can_2025}. Within the human-AI conditions, the Single Prompt Full Ontology approach shows the highest alignment with human classifications, although the differences across model architectures are not significant. However, the AI-AI IAA is significantly higher than any of these human-referenced agreements, indicating that the models are more self-consistent than they are with human annotators, and than human annotators are amongst themselves.

\begin{figure}[htbp]
    \centering
    \includegraphics[width=0.8\textwidth]{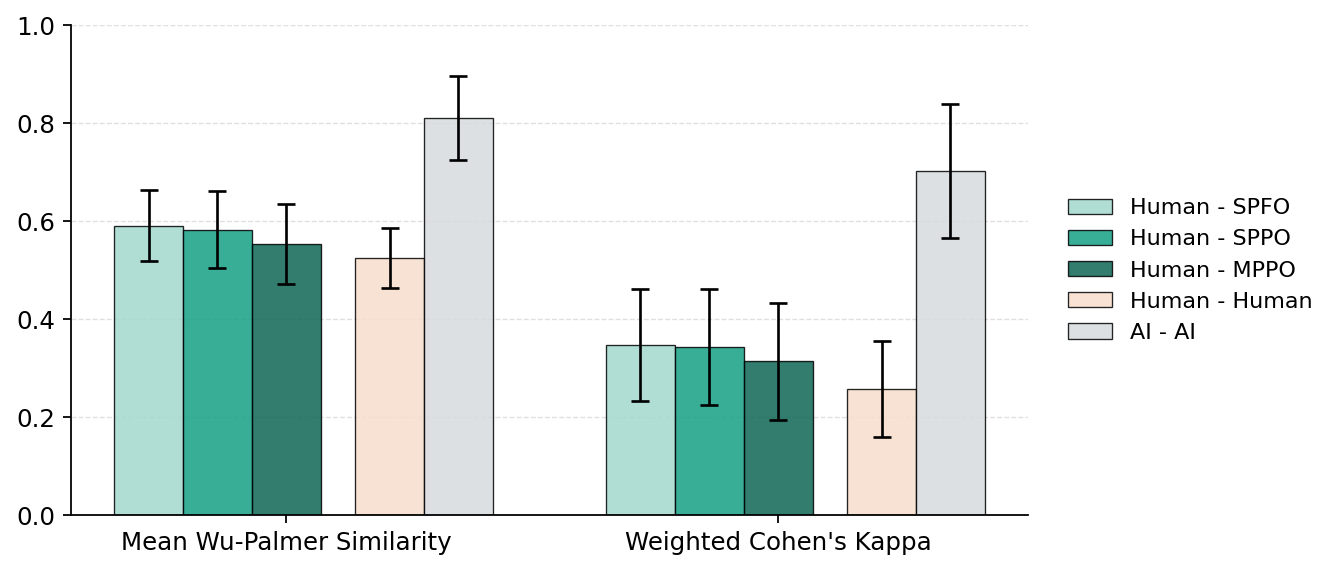}
    \caption{
        \textbf{Human-AI inter-annotator agreement for three GPT-5.1 (high-reasoning-effort) classification pipeline variants:} SPPO = Single Prompt Partial Ontology, MPPO = Multi-Prompt Partial Ontology, SPFO = Single Prompt Full Ontology. Left: Mean Wu-Palmer Similarity. Right: Weighted Cohen’s $\kappa$. Plots show 95\% confidence intervals computed via nonparametric bootstrap with 1,000 resamples.
    }
    \label{fig:4_1_human_ai_agreement}
\end{figure}

For our purposes, the most important implication of these results is that the AI classification pipelines agree as much with human raters as other humans do, and this indicates that the automatic classifications are acceptable substitutes for human classifications.

\begin{figure}[h]
    \centering
    \includegraphics[width=0.77\textwidth]{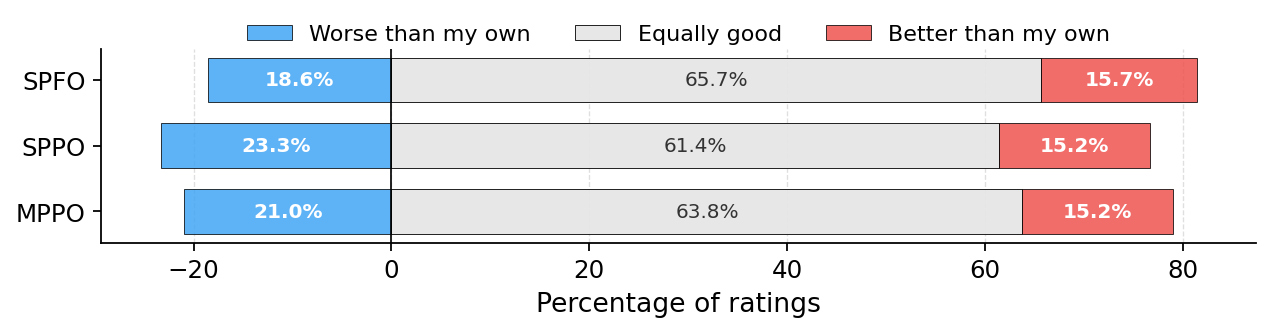}
    \caption{
        \textbf{A diverging stacked bar chart showing human evaluation of model classifications relative to their own.} The skew toward the right (red) indicates that in over 15\% of cases, human annotators rated the AI's classification and its reasoning as superior to their own initial judgment. The skew toward the left (blue) indicates that in over 18\% of cases, human annotators rated the AI's classification and its reasoning as inferior to their own initial judgment.
        Overall, human annotators judged the AI classifications to be as good as their own in over 60\% of cases, with roughly comparable proportions rated as better than (over 15\%) or worse than (over 18\%). With the final model we used (SPFO), human annotators rated the AI classifications as equal to or better than their own in 81.4\% of cases.
    }
    \label{fig:4_1_human_evaluation}
\end{figure}

We also report the human evaluators’ comparisons of the three pipeline's classifications against their own in a stacked bar chart (Figure~\ref{fig:4_1_human_evaluation}). No statistically significant differences were found across the three pipeline variants based on either the Friedman test ($p$ = 0.090) or a repeated-measures ANOVA ($p$ = 0.383). However, the Single Prompt Full Ontology approach appears to achieve the highest overall rating, receiving more ``better'' and ``equally good'' ratings and fewer ``worse'' ratings than the other models.

Most importantly for our purposes, the human annotators said that $\sim$81\% of the AI classifications are as good as or better than their own. And this provides another indication that our automated classification pipeline is at least as good as human classifications, while also being much faster and cheaper.  

These findings are also broadly consistent with prior human validation results reported by Anthropic~\cite{handa_which_2025}. Their classification task was, in some sense, easier than ours, since their human judges were judging binary acceptability over (a) a 3-level task hierarchy (vs. a median depth of 9 levels in our ontology) and (b) $\sim$100 categories (vs. ~15,989 activities in our ontology). But they observed an 86\% human-judged match rate, which is roughly comparable to our finding that $\sim$81\% of AI classifications were as good or better than human annotations. This comparability suggests that LLM-based classification can scale to much deeper and more fine-grained taxonomies.


\subsubsection{Manual classification}
\label{sec:method:pipeline:manual}

In the study of robotic systems, the research team did not rely on the automated classification pipeline used for the TAAFT applications. Instead, the team manually assigned the subclasses of robotic systems to activity nodes in the ontology. This was possible because the IFR dataset includes only 66 subclasses. 

Two authors with expertise in robotics independently mapped each IFR subclass to a single activity in the ontology. In many cases, a subclass label corresponded directly to a specific activity, without judgment being required. For example, the ``Transportation and logistics'' subclass was readily classified under ``Transport physical object.'' In other instances, greater judgment was required.  If disagreements arose, they were resolved through discussion and review by the full author team. 

\autoref{tab:ifr_to_ontology} in the Appendix~\ref{ifr taxonomy} provides the final mapping of IFR subclasses to ontology activities.


\subsection{Combining usage measures of AI software applications and robotic systems}
\label{sec:method:pipeline:combined}

In the results in Sections~\ref{sec:taaft} and \ref{sec:robotics} below, we measured AI usage intensity in terms of the number of \textit{units} of AI software applications and robotic systems, respectively. But in Section~\ref{sec:comparison}, we wanted to see a holistic view that combines both kinds of systems. 

To do this, we integrated the data from the TAAFT and IFR datasets. But because AI software applications and robotic systems differ so substantially, merely comparing the number of units deployed of each may not yield any meaningful insights. Therefore, to assess the impact of both of these technologies simultaneously, we needed a common metric for comparison. For this purpose, we converted the unit volume for each type of system into \textit{market values} expressed in US dollars. This served as a measure of the intensity of AI usage, which we take as a proxy for AI applicability. 

For AI software applications in the TAAFT dataset, we first estimated relative market \textit{shares} using the number of saves, prices, and billing frequency. Then, to obtain market \textit{value} estimates, we multiplied the respective market shares for each application by the global market value of non-robotic AI applications in 2024. To obtain this number, we subtracted the global market value of robotic systems (USD \$46.1 billion~\cite[p.~7]{statista2024robotics}) from the overall AI market value (USD \$186.4 billion~\cite[p.~20]{statista2025ai}). This results in an estimate of USD \$140.3 billion for the market value of all AI software applications. Finally, we estimated the market value of AI applications in each ontology activity by aggregating the market values of all the individual applications classified into that activity. A detailed description of this method for calculating the market value of AI applications is presented in Appendix~\ref{estimate_AI_market_value}.

For activities linked to robotic systems, we estimated market value by starting with the global robotics market size of USD 46.1 billion in 2024, as reported by Statista~\cite[p.~7]{statista2024robotics}. We then noted the share of the overall market accounted for by each key robotics industry segment (e.g., industrial, medical, consumer, etc.), again using data from Statista~\cite[p.~10]{statista2024robotics}. We next estimated an average unit price for each of the 66 IFR robot subclasses through a combination of research in industry and manufacturer sources and calibration based on linking price ranges with the segment market shares reported by Statista. Once assumptions about average system price were in place, 2024 revenues for each subclass were calculated by multiplying unit volume and average system price. Revenues were then assigned to the activities into which each subclass had been classified. Full methodological details, mappings, and calculation procedures are provided in Appendix~\ref{estimate_robot_market_value}.

\section{Acknowledgments}

This research has been supported:  (a) by the National Research Foundation (NRF), Prime Minister’s Office, Singapore, under its Campus for Research Excellence and Technological Enterprise (CREATE) program, through the Mens, Manus and Machina (M3S) interdisciplinary research group (IRG) of the Singapore–MIT Alliance for Research and Technology (SMART) center, (b) by Google.org through the MIT Pathways for AI Training and Hiring (PATH) program, (c) by the Toyota Research Institute, and (d) by the MIT Quest for Intelligence. We are especially grateful to Andrei Nedelcu and Tudor Iliescu, from There’s An AI For That, for providing the data about AI software applications that we analyzed. We also especially thank Steven Rick for early work that led to this project; Sam Ouhra and Soe Min Thant for their software development support; Xinyue Chen, Zhexuan Ma, and Jiacheng Zhang for their assistance in data analysis; and Emily Hu and Iyad Rahwan for helpful suggestions on the writing of the paper itself. In addition, we are very grateful to the following for their helpful advice at various stages of the project: Abdullah Almaatouq, Charles Fine, Gianni Giacomelli, Isabella Loaiza, Armando Solar-Lezama, Nancy Taubenslag, Dicle Uzunyayla, and Michelle Vaccaro.

\clearpage
\begin{appendices}

\section{Related work}
\label{sec:related_work}

Assessing the potential future impact of AI requires detailed knowledge of work activities. The most important sources of information about work activities are occupational databases assembled by national governments. The largest and most widely used is the U.S. Dept of Labor’s Occupational Information Network (O*NET)~\cite{ONET_Resource_Center_2016} and its predecessor, the Dictionary of Occupational Titles  (DOT)~\cite{US_Dept_of_Labor_1991}. 
 
In addition to collecting data on worker characteristics and work context for each occupation, O*NET also gathers detailed information on work activities through surveys. Survey respondents receive a questionnaire that lists tasks previously associated with their occupation and are asked how important each task is in their current job. Respondents can also add new tasks that have not been included in the standardized list. 
 
As noted above, O*NET covers over 900 occupations and lists more than 20{,}000 work tasks, DWAs, IWAs, and GWAs.
 
Other governments have developed similar datasets, for example, Singapore’s Skills Future~\cite{Skills_Future_2026} and the EU’s European Skills, Competences, Qualifications and Occupations  (ESCO)~\cite{European_Commission_2026}. 
 
The primary purpose of occupational databases is to provide information to people who are seeking work or planning careers. These databases are also valuable to government entities responsible for education and workforce training. 
 
In the early 2000s, these sources began to be used by researchers to gain insights into the impact of new technologies on the workplace. One influential early study \cite{autor_skill_2003} grouped occupations based on whether they primarily involved undertaking routine vs. non-routine tasks. The rationale was that ever more capable computers and robots would increasingly be able to replace humans in performing repetitive tasks that involved following explicit instructions. This study found that occupations with many routine tasks had experienced job losses and wage reductions in the 1980s and 1990s. 
 
This initial work spurred a plethora of subsequent studies that examined the impact of new technologies on tasks and jobs. This stream of work came to be known as employing the ``task approach'' or ``task-based approach,'' because of its focus on the properties of tasks performed within an occupation, as opposed to other features of the occupation like skills, work contexts, education levels, etc.~\cite{autor_task_2013}.

Most studies undertaken before 2010 relied on grouping occupations listed in the DOT, based on commonalities in their underlying tasks. The impact of new technologies on each occupational grouping was then assessed. After O*NET went online in 2008, researchers began to take a finer-grained approach by examining the potential impact of emerging technologies not only on groups of occupations but also on individual occupations and even individual tasks. 
 
One influential study \cite{Frey_and_Osborne_2013} examined the O*NET tasks associated with  occupations and projected the potential for information technology to drive job losses in each. Another study \cite{brynjolfsson_what_2018} developed a 21-step rubric to identify tasks machine learning (ML) could perform and then used it to assess ML’s potential impact on individual O*NET tasks and, in turn, occupations where those tasks were important.
 
In the realm of physical work, recent studies have suggested that industrial robots may affect workplace patterns in different ways than service robots~\cite{acemoglu_robots_2020, lee_robots_2025}. These findings highlight the need for fine-grained, task-level analyses to delineate the potentially variegated impacts of different types of robotic systems. 
 
After the launch of ChatGPT in 2022, a number of researchers undertook studies to examine the impact of generative AI systems. A group of controlled experiments showed generative AI could increase output, speed, and/or quality in tasks as disparate as writing~\cite{Noy_Zhang_2023}, customer support~\cite{brynjolfsson_generative_2025}, software development~\cite{cui_effects_2025} and duties typically undertaken by management consultants~\cite{dellacqua_navigating_2023}. Notably, these studies all examined the impact of generative AI on discrete tasks in particular organizational settings, but none provided generalized findings across a broad range of tasks. 
 
In the past year, several leading LLM developers\textemdash Anthropic~\cite{handa_which_2025, appel_anthropic_2025}, Microsoft~\cite{tomlinson_working_2025}, and OpenAI~\cite{chatterji_how_2025}\textemdash have used a task-based approach to examine the impact of their systems on work practices. Anonymized data from prompt logs was mapped against tasks in the O*NET database, and the results were then aggregated at various levels to identify occupations and categories of tasks most likely to be affected by the usage of LLMs in the future.

Not surprisingly, LLMs were primarily used for information-related tasks, such as writing, software development, obtaining information, and conveying information to others (for example, in response to customer inquiries). One additional interesting finding was that over time, users increasingly asked LLMs to undertake larger and more complex tasks, while also providing the tool with less guidance~\cite{appel_anthropic_2025}. 
\autoref{tab:dataset_comparison} summarizes the data sources used and the approach employed in these studies. 
  
These studies have identified tasks for which users have most commonly asked general-purpose LLMs to provide help. But they did not examine the potential impact of AI across the full range of work tasks, nor did they identify tasks where other AI tools with different functionality may have an impact. 

In the approach described in this paper, we classify work activities using a principled, systematically structured, and comprehensive ontology.  By grouping similar activities into ``family trees'' of specific, related categories, we avoid some of the confusion caused by O*NET and previous classification systems. And we believe that this, together with rich data sources like those we analyze here, can help us gain a more precise understanding of the workplace changes occurring today and likely to occur in the future.

\begin{table}[h]
    \centering
    \caption{
        \textbf{Summary of approaches used by Anthropic, Microsoft, and OpenAI in studies of users' conversations with their LLMs.}
    }
    \label{tab:dataset_comparison}
    \begin{tabular*}{\textwidth}{@{\extracolsep\fill}p{3.2cm}p{2.8cm}p{2.8cm}p{2.8cm}}
    \toprule
    \textbf{Study sponsor} 
    & \textbf{Anthropic} 
    & \textbf{Microsoft} 
    & \textbf{OpenAI} \\
    \midrule
    
    \textbf{LLM employed} 
    & Claude 
    & Copilot 
    & ChatGPT \\
    \midrule
    
    \textbf{Approximate number of conversations with LLMs in dataset} 
    & 1 million 
    & 200{,}000 
    & 1 million \\
    \midrule
    
    \textbf{Method to map user conversations} 
    & Claude used to position each conversation within three-level hierarchical groupings of O*NET tasks developed by the Anthropic research team
    & GPT-4o used to map each conversation to best-fitting O*NET Intermediate Work Activity (IWA) 
    & GPT-5 mini used to map each conversation to the best-fitting O*NET Intermediate Work Activity (IWA) \\
    \midrule
    
    \textbf{Level of aggregation at which task classification is reported} 
    & O*NET tasks and associated occupations/skills 
    & IWAs, which are linked to relevant Generalized Work Activities (GWAs); IWA data then used to assess potential future impact of LLMs on occupations 
    & IWA data aggregated and reported at GWA level; IWA data also mapped to 24 categories created by OpenAI to classify features and capabilities of current and future GPTs \\
    \botrule
    \end{tabular*}
    \footnotetext{Source: Anthropic (\cite{handa_which_2025, appel_anthropic_2025}), Microsoft (\cite{tomlinson_working_2025}), and OpenAI (\cite{chatterji_how_2025})}
\end{table}


\clearpage
\section{Ontology Construction Details}
\label{sec:appendix-construction}
\subsection{Details on Initial Explorations}
\label{ap:initial-explorations}

Before we established the methodology described in the main paper, we explored a number of different approaches, including using automated methods only, using humans only, and using other knowledge sources for reference. 

Using humans only, we attempted to organize O*NET tasks that contain frequently occurring verbs, such as ``Direct,'' ``Administer,'' ``Manage,'' ``Clean,'' etc., into a hierarchical structure. From this process, we gained several insights: 1) tasks needed to be broken down into atomic units (focused on verb-object pairs), as compound tasks are difficult to organize hierarchically, and 2), we needed to differentiate between different meanings of the same word (e.g. ``administer program'' vs ``administer drug''), and different words with the same meaning (e.g. ``administer drug'' vs ``dispense drug''). These insights led to the design principles of \textit{atomicity} and \textit{specificity}.

Because human-only approaches would not be feasible with a dataset as large as O*NET, we explored the use of LLMs as a scalable approach to process large amounts of data. Working with  LLMs only, we explored both using multiple steps to construct the hierarchy (first having an LLM organize the most frequently occurring verbs in O*NET, then using separate LLMs to organize subtrees beneath those verbs), as well as a single step (giving all verbs in O*NET to an LLM along with a comprehensive prompt specifying the desired properties of the hierarchy). We found that it was difficult to achieve the elegance and logical consistency we desired using LLMs only, even after many iterations of prompt engineering. Moreover, we found that multi-step approaches made it easier for LLMs to maintain logical consistency, and for the output to be more interpretable by our team. We also briefly explored using hierarchical clustering of semantic embeddings. However, neither LLMs nor embedding clustering could provide the specificity feature we needed.

Thus, we explored bringing in existing lexical and ontological databases to ground the hierarchy construction. While we looked into FrameNet \cite{baker_berkeley_1998}, SUMO \cite{niles_towards_2001}, and other ontologies, we ultimately decided to use WordNet for its extensive definitions, hypernym relationships, and broad usage across existing research. By automating (using LLMs) the definition-selection process for verbs in their task context, we could leverage WordNet’s existing hypernym structure as a starting point for our own hierarchy. Ultimately, we ended up using mostly human editing of the WordNet-based hierarchy to produce our initial draft of the ontology. Once we had established a structure that we thought met our principles, we used more LLM assistance in expanding this structure to incorporate more activities.

\subsection{Hierarchy Trees from WordNet}
\label{ap:wordnet-hierarchy-trees}

The following is a full list of the 60 top-level ``synsets'' (sets of synonyms that comprise a single meaning) from the WordNet hierarchy trees that appear in O*NET tasks, DWAs, IWAs, and GWAs. Wordnet includes multiple definitions for many verbs, and the definition we used is indicated by the suffix v.01, v.02, etc. The prefix [virtual] indicates when the synset does not directly appear in an O*NET item, but rather is the top-level parent of synset(s) that do directly appear.

\setlength{\columnsep}{-0.4cm}
\begin{multicols}{3}
  \begin{itemize}[label={}]
    \item $[$virtual$]$ account.v.02
    \item $[$virtual$]$ act.v.01
    \item $[$virtual$]$ agree.v.01
    \item $[$virtual$]$ change.v.02
    \item $[$virtual$]$ create.v.02
    \item $[$virtual$]$ decide.v.01
    \item $[$virtual$]$ examine.v.02
    \item $[$virtual$]$ express.v.02
    \item $[$virtual$]$ forget.v.01
    \item $[$virtual$]$ get\_rid\_of.v.01
    \item $[$virtual$]$ guide.v.05
    \item $[$virtual$]$ have.v.01
    \item $[$virtual$]$ include.v.01
    \item $[$virtual$]$ induce.v.02
    \item $[$virtual$]$ kill.v.01
    \item $[$virtual$]$ mean.v.03
    \item $[$virtual$]$ perceive.v.01
    \item $[$virtual$]$ prevent.v.02
    \item $[$virtual$]$ show.v.04
    \item $[$virtual$]$ spread.v.01
    \item $[$virtual$]$ take.v.08
    \item $[$virtual$]$ think.v.03
    \item $[$virtual$]$ touch.v.01
    \item $[$virtual$]$ trade.v.01
    \item $[$virtual$]$ transfer.v.05
    \item $[$virtual$]$ understand.v.01
    \item $[$virtual$]$ unmake.v.01
    \item $[$virtual$]$ use.v.01
    \item $[$virtual$]$ watch.v.01
    \item address.v.02
    \item analyze.v.01
    \item change.v.01
    \item close.v.01
    \item complete.v.05
    \item confirm.v.01
    \item connect.v.01
    \item control.v.01
    \item correspond.v.03
    \item cover.v.01
    \item cultivate.v.01
    \item determine.v.01
    \item enter.v.01
    \item enter.v.02
    \item exercise.v.04
    \item gather.v.01
    \item get.v.01
    \item hire.v.01
    \item make.v.01
    \item meet.v.01
    \item move.v.02
    \item perform.v.01
    \item remove.v.01
    \item search.v.01
    \item start.v.08
    \item stop.v.01
    \item support.v.01
    \item take.v.04
    \item travel.v.01
    \item treat.v.03
    \item work.v.01
  \end{itemize}
\end{multicols}

\subsection{Placement of previously omitted meanings}
\label{ap:synset-placement}

The construction of the initial hierarchy draft omitted 644 synsets from O*NET tasks, for two types of reasons: first, synsets that were flagged by the LLM definition checker to be invalidly assigned were omitted to allow manual definition assignment in the Revision phase; second, synsets that infrequently occurred in O*NET were omitted to reduce the number of hierarchical trees to make it more manageable for LLMs and humans to organize into an overarching framework. To place these omitted definitions, an LLM was used to make an initial check and placement recommendation for each synset. Each prompt includes the full initial hierarchy and representative O*NET tasks using that synset, as well as the verb’s definition, synonyms, and hypernym path from WordNet. 

The model’s returned output included a choice to either create a new node or append the verb as a synonym to an existing one, the full recommended path to place the synset, and a short explanation if the synset was deemed incorrect for the associated tasks. Each path was deterministically validated as a real path in the hierarchy, and the LLM was re-prompted up to three times to replace invalid recommendations. 

A subset of outputs was manually reviewed, leading to prompt adjustments emphasizing the importance of deeper node selection, clarifying the identification of incorrect synsets, and creating a parallel run without the verb path to test whether better recommendations would be made. This updated prompt was used over the missing O*NET tasks and synsets a second time, and the resulting recommendations were reviewed and edited by humans to finalize the placement.

\subsection{Principles for direct-object classification system}
\label{ap:direct-object-principles}

To classify the direct objects of verb-object pairs into high-level categories (Physical Object, Information, Actor, Activities), we developed definitions, examples, and differentiation principles for the categories that were fed into an LLM classifier, detailed below:

\textit{Physical Object}
\begin{itemize}
    \item Tangible, material things whose physical properties (shape, weight, size, location, mechanical use) are important for the activity.
\item Includes tools, artifacts, natural objects, substances, and places.
\item Examples: lamp, hat, tree, house, city, glass of water.
\end{itemize}

\textit{Information}
\begin{itemize}
    \item Abstract entities whose meaning, content, or representation is important for the activity.
\item Includes knowledge, ideas, data, qualities, quantities, and symbolic representations.
\item Examples: the contents of a book, a YouTube video, a teacher's answer, the concept of freedom, beauty, number.
\end{itemize}

\textit{Actor}
\begin{itemize}
    \item Entities that perform or can be treated as performing actions.
\item Includes people, animals, organizations, groups, and roles.
\item Examples: teacher, dog, government, team, leader, citizen.
\end{itemize}

\textit{Activity}
\begin{itemize}
    \item Happenings, occurrences, or activities.
\item Includes both bounded events (specific occurrences) and activities (processes, practices).
\item Examples: wedding, earthquake, meeting, competition, concert, management, research, inspection, negotiation, teaching, training, conversation, swimming, celebration, investigation, design, maintenance, construction.
\end{itemize}

The distinction between categories depends on what aspects of the objects are important for the task context. For example, for our purposes, we treat a book as an information object if it is being used for what it ``means'' or ``represents.'' But if the same book is used only as a doorstop, we would treat it as a physical object. In general, we treat something as a physical object if its most important properties for the activity are physical, and we treat it as information if its most important properties are its meaning.

\subsection{Verb-object inconsistency resolution system}
\label{ap:inconsistency-resolution}

Our way of choosing definitions for activities, relying on WordNet definitions, often leaves room for ambiguity in verbs with meanings that apply to both physical objects and information. For example, the definition chosen for ``Cut'' from WordNet is ``separate with or as if with an instrument,'' which can apply to both physical objects and information. However, in our ontology activities acting on physical objects and activities acting on information are in separate branches. Therefore, we needed more specific versions of ``Cut'' in both branches: ``Cut physical objects'' and ``Cut information.'' 

We used our noun classification framework to identify 615 conflicts between verb path and object classification (for example, if ``Cut Music'' was classified under the path ``Act on physical objects $> … >$ Cut physical objects'', it should be moved to a mirror branch under ``Act on information $> … >$
Cut information''). Some instances of conflict, however, do not reflect a verb misclassification. For example, ``Observe equipment'' should still be under ``Act on information,'' even though equipment is a physical object, because the process of observation involves only processing information. To deal with these complexities, we used a combination of LLM reasoning and human judgment to decide what to do with each instance of conflict identified. In each case, an LLM model made a recommendation with reasoning about whether to move the task to the mirror category or not, and then a human reviewed. We first tested the LLM on a subset of conflicts, then updated the prompt to improve it before running the process on the full batch; then, two of our human editors reviewed each recommendation and revised or accepted it.

\subsection{Distinguished activities with the same name but different meanings}
\label{ap:polysemous-activities}

We distinguished activities for which O*NET used the same verb for different meanings. Two categories of such distinctions are made. First, activities using the same title but linked to different WordNet meanings were distinguished through manual human editing. Second, activities using the same title, linked to the same WordNet meaning, but distinguished based on direct object in our classification (e.g. ``Adapt (information) (Adapt.v.01)'' and ``Adapt (physical) (Adapt.v.01)'') were identified through automated checks for verb-object inconsistencies (see Appendix \ref{ap:inconsistency-resolution}) and validated by human editors. The following is a full list of differentiations of 215 activities with the same name but different meanings:

\begin{itemize}
    \item Access (Access.v.01), Access (physical) (Access.v.02)
\item Adapt (information) (Adapt.v.01), Adapt (physical) (Adapt.v.01)
\item Add (information) (Add.v.01, Include.v.01), Add (physical) (Add.v.01, Include.v.01)
\item Address (Address.v.03), Address (express) (Address.v.02, Deliver.v.01), Address (solve) (Address.v.05)
\item Adjust (information) (Adjust.v.01), Adjust (physical) (Adjust.v.01)
\item Administer (manage) (Administer.v.01), Administer (treat) (Administer.v.04)
\item Admit (accept) (Admit.v.03), Admit (let in) (Admit.v.02)
\item Align (information) (Align.v.01), Align (physical) (Align.v.01)
\item Alter (information) (Alter.v.03), Alter (physical) (Alter.v.03)
\item Apply (for) (Apply.v.03), Apply (physical) (Put\_on.v.07)
\item Arrange (make arrangements for) (Arrange.v.02), Arrange (put in proper or systematic order) (Arrange.v.01)
\item Attend (care) (Attend.v.02), Attend (go to) (Attend.v.01)
\item Balance (account) (Balance.v.02), Balance (equilibrate) (Balance.v.01)
\item Blend (information) (Blend.v.01, Blend.v.03), Blend (physical) (Blend.v.01, Blend.v.03)
\item Blow (air) (Blow.v.14), Blow (shape by blowing) (Blow.v.06)
\item Break (Break.v.02, Sever.v.01), Break (into components) (Break.v.02)
\item Call (Call.v.05), Call (out) (Call.v.25)
\item Capture (catch) (Capture.v.06), Capture (collect) (Trap.v.03)
\item Charge (electrically) (Charge.v.24), Charge (money) (Charge.v.03)
\item Clean (information) (Clean.v.01, Clean.v.02), Clean (physical) (Clean.v.01, Clean.v.02)
\item Clear (information) (Clear.v.05, Clear.v.22), Clear (physical) (Clear.v.05, Clear.v.22)
\item Collect (information) (Roll\_up.v.02), Collect (physical) (Roll\_up.v.02)
\item Complete (activity) (Complete.v.01), Complete (information) (Complete.v.05)
\item Configure (information) (Configure.v.01), Configure (physical) (Configure.v.01)
\item Connect (information) (Affix.v.01, Attach.v.01, Connect.v.01, Join.v.02), Connect (physical) (Affix.v.01, Attach.v.01, Connect.v.01, Join.v.02)
\item Construct (information) (Construct.v.01, Raise.v.09), Construct (physical) (Construct.v.01, Raise.v.09)
\item Copy (information) (Copy.v.04, Duplicate.v.01, Reproduce.v.01, Twin.v.01), Copy (physical) (Copy.v.04, Duplicate.v.01, Reproduce.v.01, Twin.v.01)
\item Customize (information) (Customize.v.02), Customize (physical) (Customize.v.02)
\item Cut (information) (Cut.v.01), Cut (physical) (Cut.v.01)
\item Defend (prevent) (Defend.v.02), Defend (verbally) (Defend.v.01)
\item Deposit (money) (Deposit.v.02), Deposit (physical) (Situate.v.02)
\item Develop (chemically) (Develop.v.15), Develop (idea) (Explicate.v.02)
\item Disconnect (information) (Disconnect.v.02), Disconnect (physical) (Disconnect.v.02)
\item Dock (Dock.v.05), Dock (cut) (Dock.v.04)
\item Draw (drawing) (Draw.v.04, Draw.v.06), Draw (extract) (Draw.v.05)
\item Dress (butchery) (Dress.v.06), Dress (clothes) (Dress.v.02), Dress (wound) (Dress.v.14)
\item Drive (force) (Force.v.06), Drive (operate) (Drive.v.01, Drive.v.03)
\item Engage (people) (Involve.v.06), Engage (physical) (Engage.v.10)
\item Enter (Enter.v.01), Enter (information) (Enter.v.01)
\item Extend (information) (Expand.v.02, Extend.v.17, Widen.v.04), Extend (physical) (Expand.v.02, Extend.v.17, Widen.v.04)
\item File (formal) (File.v.01), File (information) (File.v.05), File (physical) (File.v.05), File (surface) (File.v.02)
\item Fill (information) (Fill.v.01, Inflate.v.02), Fill (physical) (Fill.v.01, Inflate.v.02)
\item Finish (information) (Complete.v.02), Finish (physical) (Complete.v.02)
\item Frame (compositional) (Frame.v.04), Frame (physical) (Frame.v.02)
\item Gather (information) (Gather.v.01), Gather (physical) (Gather.v.01)
\item Grade (Grade.v.03), Grade (shape) (Grade.v.02)
\item Harmonize (information) (Harmonize.v.04, Harmonize.v.05), Harmonize (physical) (Harmonize.v.04, Harmonize.v.05)
\item Incorporate (information) (Incorporate.v.02), Incorporate (physical) (Incorporate.v.02)
\item Insert (information) (Feed.v.04, Insert.v.01), Insert (physical) (Feed.v.04, Insert.v.01)
\item Install (information) (Install.v.01), Install (physical) (Install.v.01)
\item Lay (Lay.v.03), Lay (physical) (Lay.v.03, Lay\_out.v.05, Range.v.05)
\item Load (digital) (Load.v.03), Load (physical) (Load.v.01)
\item Maintain (information) (Conserve.v.02, Keep.v.01), Maintain (physical) (Conserve.v.02, Keep.v.01)
\item Map (between) (Map.v.06), Map (out) (Map.v.01)
\item Mark (information) (Code.v.01, Mark.v.02, Tag.v.01), Mark (physical) (Mark.v.02)
\item Match (align) (Match.v.01), Match (co-fund) (Match.v.02)
\item Model (Model.v.03), Model (represent) (information) (Model.v.01, Model.v.05), Model (represent) (physical) (Model.v.01, Model.v.05)
\item Open (Open.v.01), Open (make available) (Open.v.06)
\item Order (authoritative) (Order.v.01), Order (buy) (Order.v.02)
\item Organize (analytically) (Organize.v.04, Structure.v.01), Organize (people or events) (Hold.v.03, Organize.v.05)
\item Orient (person) (Orient.v.02), Orient (physical) (Orient.v.03)
\item Pack (compress) (Pack.v.02), Pack (into) (Pack.v.13), Pack (together) (information) (Pack.v.01), Pack (together) (physical) (Pack.v.01)
\item Paint (information) (Paint.v.02), Paint (physical) (Paint.v.02)
\item Pass (achieve) (Pass.v.09), Pass (over) (Pass.v.05, Present.v.04)
\item Patch (connect) (Patch.v.02), Patch (repair) (Patch.v.03)
\item Perform (action) (Do.v.03, Perform.v.01, Prosecute.v.03, Take.v.01), Perform (artistic) (Perform.v.03)
\item Post (announcement) (Post.v.01), Post (financial) (Post.v.07)
\item Prepare (information) (Fix.v.12), Prepare (physical) (Fix.v.12)
\item Press (flatten) (Press.v.04), Press (touch) (Press.v.01)
\item Process (information) (Process.v.01, Process.v.03), Process (physical) (Process.v.01)
\item Raise (funds) (Raise.v.04), Raise (physical) (Raise.v.02)
\item Read (gauges, instruments, etc.) (Read.v.08), Read (out loud) (Read.v.03), Read (something written or printed) (Read.v.01)
\item Reassemble (information) (Reassemble.v.01), Reassemble (physical) (Reassemble.v.01)
\item Refine (information) (Refine.v.06), Refine (substance) (Refine.v.04)
\item Regulate (authoritative) (Regulate.v.02), Regulate (modulate) (information) (Regulate.v.01), Regulate (modulate) (physical) (Regulate.v.01)
\item Remove (information) (Remove.v.01), Remove (physical) (Remove.v.01)
\item Renew (information) (Regenerate.v.01), Renew (physical) (Regenerate.v.01)
\item Repair (information) (Correct.v.01, Correct.v.08, Repair.v.01), Repair (physical) (Correct.v.01, Correct.v.08, Repair.v.01)
\item Replace (information) (Replace.v.01), Replace (physical) (Replace.v.01)
\item Report (Report.v.01, Report.v.02), Report (to authorities) (Report.v.04)
\item Roll (Roll.v.01), Roll (shape) (Roll.v.08, Roll\_Out.v.01)
\item Run (movement) (Run.v.01), Run (perform) (Run.v.21), Run (placement) (Run.v.03)
\item Serve (deliver) (Serve.v.11), Serve (role) (Serve.v.02)
\item Service (information) (Service.v.02), Service (physical) (Service.v.02)
\item Shave (groom) (Shave.v.01), Shave (surface) (Shave.v.02)
\item Sign (signal) (Sign.v.05), Sign (signature) (Sign.v.04)
\item Snip (information) (Snip.v.02), Snip (physical) (Snip.v.02)
\item Splice (information) (Splice.v.04), Splice (physical) (Splice.v.04)
\item Strike (hit) (Strike.v.01), Strike (ignite) (Strike.v.11)
\item Substitute (replace) (Substitute.v.01), Substitute (role) (Substitute.v.03)
\item Synchronize (information) (Synchronize.v.01, Synchronize.v.03, Synchronize.v.04), Synchronize (physical) (Synchronize.v.01, Synchronize.v.03, Synchronize.v.04)
\item Synthesize (information) (Synthesize.v.01), Synthesize (physical) (Synthesize.v.01)
\item Tailor (information) (Tailor.v.01), Tailor (physical) (Tailor.v.01)
\item Tap (draw from) (Tap.v.02), Tap (touch) (Tap.v.03)
\item Touch Up (information) (Touch\_up.v.01), Touch Up (physical) (Touch\_up.v.01)
\item Trace (analytically) (Trace.v.01), Trace (drawing) (information) (Trace.v.07), Trace (drawing) (physical) (Trace.v.07)
\item Transcribe (music) (Transcribe.v.03), Transcribe (write) (Transcribe.v.01)
\item Travel (Transport self), Travel (Travel.v.01)
\item Troubleshoot (information) (Trouble-shoot.v.01), Troubleshoot (physical) (Trouble-shoot.v.01)
\item Turn (on/off) (Turn.v.22), Turn (rotate) (Revolve.v.01, Turn.v.01, Turn.v.17)
\item Upload (information) (Upload.v.01), Upload (physical) (Upload.v.01)
\item Wrap (Envelop.v.01, Wrap.v.01), Wrap (gift) (Gift-Wrap.v.01)
\end{itemize}

\subsection{Full list of activities with multiple inheritance}

As described in Section \ref{sec:background:multiple}, the ontology features \textit{multiple inheritance}, where activities have multiple parents. The full list of activities with multiple inheritance is presented below\footnote{Suffixes such as ``.v.01'' correspond to WordNet sense identifiers and are included to differentiate distinct meanings of the same word.}.

\label{ap:taaft:multiple_inheritance}
\small
\begin{longtable}{
    >{\raggedright\arraybackslash}p{0.31\linewidth}
    >{\raggedright\arraybackslash}p{0.61\linewidth}
}
    \caption
    [Activities and their parent generalizations]
    {\textbf{Activities and their parent generalizations.
    }
    \label{tab:activity_generalizations}.} \\
    
    \toprule
    \textbf{Activity} & \textbf{Generalizations} \\
    \midrule
    \endfirsthead
    
    \caption[]{\textbf{Activities and their parent generalizations.} (continued)} \\
    \toprule
    \textbf{Activity} & \textbf{Generalizations} \\
    \midrule
    \endhead
    
    \midrule
    \multicolumn{2}{r}{\textit{Continued on next page}} \\
    \midrule
    \endfoot
    
    \bottomrule
    \endlastfoot
    
    
    \textbf{Agree (Agree.v.01)} 
    & Collaborate (Collaborate.v.01) \\
    & Exchange information (Change.v.06) \\
    \midrule
    
    \textbf{Brush (Brush.v.04)} 
    & Reorient physical objects \\
    & Touch (Touch.v.01) \\
    \midrule
    
    \textbf{Casket (Casket.v.01)} 
    & Cover (Cover.v.01) \\
    & Enclose (Enclose.v.03) \\
    \midrule
    
    \textbf{Communicate (transfer information end-to-end)} 
    & Transfer information \\
    & End-to-end information transfer \\
    \midrule
    
    \textbf{Conduct (Conduct.v.01, Conduct.v.02)} 
    & Manage (Manage.v.02, Treat.v.01) \\
    & Perform action (Do.v.03, Perform.v.01, Prosecute.v.03, Take.v.01) \\
    \midrule
    
    \textbf{Draw (extract) (Draw.v.05)} 
    & Extract (Draw.v.07, Extract.v.01) \\
    & Pull (Pull.v.01, Pull.v.17) \\
    \midrule
    
    \textbf{Estimate (Estimate.v.01)} 
    & Calculate (Calculate.v.01, Calculate.v.02) \\
    & Decide \\
    \midrule
    
    \textbf{Express Live} 
    & Express information \\
    & Provide information \\
    \midrule
    
    \textbf{Fly (Fly.v.03)} 
    & Transport self as passenger \\
    & Operate \\
    \midrule
    
    \textbf{Gather (information) (Gather.v.01)} 
    & Combine information (Combine.v.04) \\
    & Get information (Get.v.01, Obtain.v.01, Procure.v.01) \\
    \midrule
    
    \textbf{Gather (physical) (Gather.v.01)} 
    & Combine physical objects (Create from raw material.v.01) \\
    & Get physical objects (Get.v.01, Obtain.v.01, Procure.v.01) \\
    \midrule
    
    \textbf{Implement (Carry through.v.01, Follow through.v.02)} 
    & Act on activities \\
    & Manage (Manage.v.02, Treat.v.01) \\
    \midrule
    
    \textbf{Improve (Better.v.02, Enhance.v.02)} 
    & Act on activities \\
    & Manage (Manage.v.02, Treat.v.01) \\
    & Transform information (Change.v.01, Rework.v.01, Transform.v.02, Work.v.05) \\
    \midrule
    
    \textbf{Organize (analytically) (Organize.v.04, Structure.v.01)} 
    & Reason (Reason.v.03, Think.v.03) \\
    & Transform information (Change.v.01, Rework.v.01, Transform.v.02, Work.v.05) \\
    \midrule
    
    \textbf{Propose (Offer.v.04, Project.v.08, Propose.v.01)} 
    & Plan (Mastermind.v.01, Plan.v.02, Plan.v.03) \\
    & Send information (Send.v.02, Transmit.v.04) \\
    \midrule
    
    \textbf{Pull (Pull.v.01, Pull.v.17)} 
    & Move physical objects \\
    & Remove (physical) (Remove.v.01) \\
    \midrule
    
    \textbf{Respond (React.v.01)} 
    & Provide service \\
    & Send information (Send.v.02, Transmit.v.04) \\
    \midrule
    
    \textbf{Sign (signal) (Sign.v.05)} 
    & Express Live \\
    & Inform (Advise.v.02, Inform.v.01) \\
    \midrule
    
    \textbf{Talk (Talk.v.02)} 
    & Express Live \\
    & Send information (Send.v.02, Transmit.v.04) \\
    \midrule
    
    \textbf{Unwind (Unwind.v.01)} 
    & Shape (Change\_shape.v.01, Shape.v.02, Shape.v.03) \\
    & Turn or rotate (Revolve.v.01, Turn.v.01, Turn.v.17) \\
    \midrule
    
    \textbf{Upload (information) (Upload.v.01)} 
    & Send information (Send.v.02, Transmit.v.04) \\
    & Store information (Conserve.v.02, Store.v.01, Store.v.02) \\
    \midrule
    
    \textbf{Wire (Wire.v.01)} 
    & Connect physical objects (Affix.v.01, Attach.v.01, Connect.v.01, Join.v.02) \\
    & Equip (Equip.v.01) \\

\end{longtable}
\normalsize

\newpage
\section{Ontology Platform Functionality}
\label{ap:platform}
To develop this ontology of work activities, we built a curation platform that supports synchronous multi-user editing, search, and structured browsing of a large activity knowledge base. The platform is designed to support the complex, hierarchical relationships inherent in process taxonomies while providing the coordination mechanisms necessary for coherent, auditable collaboration.

Our design builds on the MIT Process Handbook~\cite{malone_organizing_2003}, which demonstrated the utility of organizing knowledge about processes through generalization-specialization and part-whole decomposition. While early ontology engineering relied heavily on manual expert curation~\cite{gruber_translation_1993}, prior work has shown that LLMs can accelerate ontology development when used as assistants rather than sole authors~\cite{garcia_fernandez_ontology_2024}. Our platform adopts a human-in-the-loop architecture~\cite{schroeder_just_2025}, with hybrid search mechanisms that combine semantic retrieval with keyword matching~\cite{karpukhin_dense_2020}, so contributors can rapidly locate candidate nodes while maintaining domain specificity and controlling hallucination risks.

To preserve structural coherence, the platform requires each node to be linked to at least one generalization, blocks edits that would orphan nodes, flags near-duplicate nodes (based on lexical + embedding similarity) to discourage redundant node creation, and maintains node-level change histories with attribution. The platform also exports versioned ontology snapshots (in JSON format) so that analyses and automated classification can reference a fixed ontology instance.

The platform is implemented with Firestore (data storage), Firebase Authentication (user access), Google Cloud Run (deployment), Next.js (web framework), Material UI (components), Jest (testing), Y.js (real-time collaboration), and D3.js (visualization).

\subsection{Data Structure and Inheritance}

The platform represents the ontology as a directed graph where nodes represent simple activities or broader processes. Following the approach in the Process Handbook, these nodes are organized through two primary kinds of hierarchical relationships: \textit{types} and \textit{parts}.

\begin{itemize}
    \item \textbf{Types (specializations and generalizations)}: These relationships capture the taxonomic structure. A specialized activity (specialization) inherits properties (such as AI applicability and parts) from its generalization(s) but can also override them to fit a specific context.
    \begin{itemize}
        \item \textit{Example}: Consider the activity ``Select.'' It is categorized as a specialization of ``Decide.'' We classify its further specializations (e.g., ``Select job candidate'') based on the object being selected, the reason for selection, or the selection process, narrowing down level-by-level until reaching atomic tasks.
    \end{itemize}
    \item \textbf{ Parts (parts and is-part-of)}: In addition to taxonomic structure, the platform also captures the internal structure of activities. Each activity can be decomposed into a set of sub-activities, referred to as ``parts,'' and each activity can also be part of a larger activity.
    \begin{itemize}
        \item \textit{Example}: The activity ``Select'' is decomposed into parts such as ``Gather information,'' ``Identify alternatives,'' ``Make selection,'' and ``Communicate selection.'' And some of these parts (e.g., ``Gather information'') might also be part of other larger activities.
        \item \textit{Inheritance of Parts}: The platform supports sophisticated inheritance logic for these parts. For instance, ``Gather information'' and ``Identify alternatives'' are directly inherited from the parent activity ``Decide,'' while ``Make selection'' is a specialization of the corresponding part in ``Decide.''
    \end{itemize}
\end{itemize}

\subsection{Navigation and Visualization}

To assist users in navigating this high-dimensional structure, the platform provides three connected interfaces:

\begin{itemize}
    \item \textbf{Graph View}: A Directed Acyclic Graph (DAG), visualizing the broader network of specializations and generalizations.
    \item \textbf{Outline View}: A nested list view allowing users to rapidly expand or collapse branches of the ontology. Contributors can restructure relationships (e.g., changing a node’s generalization) via drag-and-drop actions in this view.
    \item \textbf{Node View}: A detailed view displaying the node’s metadata, description, actors, and inheritance paths. To prevent accidental modifications during navigation, users can toggle a ``Read-Only'' mode.
\end{itemize}

\subsection{Hybrid Semantic Search}

To facilitate discovery within the ontology, we implemented a hybrid search engine that includes both keyword search and semantic search.  For the semantic search, we embedded every node\textemdash including its title, description, synonyms, and the titles of its parents and children\textemdash using OpenAI's ``text-embedding-3-large'' model. These embeddings are stored in a Chroma vector database. When a user queries the system, we retrieve nodes with the highest cosine similarity to the embedding of the query, and we display query results in a list that alternates between keyword search results and semantic search results. 

\subsection{Synchronous Collaboration and Coordination}

Given the scale of the ontology, coordination among contributors is critical. The platform includes several socio-technical mechanisms to maintain coherence:

\begin{itemize}
    \item \textbf{Real-Time Synchronization}: We implemented a synchronous editing architecture. When a user modifies a node, the changes are instantly propagated to all other active clients\textemdash including users viewing linked nodes or those inheriting from the modified node\textemdash ensuring a consistent state across the distributed team.
    \item \textbf{Granular History and Attribution}: Every node maintains an interactive edit history, allowing users to visualize its evolution. Similarly, a global history log tracks individual user contributions, providing transparency regarding how different contributors shape the ontology.
    \item \textbf{Contextual Communication}: To resolve semantic ambiguity in real-time, the platform features an integrated chatroom for high-level alignment. Additionally, users can leave threaded comments directly on specific nodes and tag (@mention) other contributors, anchoring discussions to the relevant data points.
\end{itemize}

\subsection{Access and Extensibility}

Write access is restricted to authorized accounts to maintain content quality. The platform is architected for openness; it provides a documented API for programmatic interaction and allows users to export the full ontology in JSON format. The architecture supports broader access for human and AI contributors, subject to access controls.

\clearpage
\section{Evolution and evaluation of automated classification architectures}
\label{ap:classification}

This appendix provides implementation details for the automated classification pipeline used to map AI software applications to ontology activities (Section~\ref{sec:method:pipeline:automated}). We focus on the three pipeline variants evaluated in our human validation study: Single Prompt Partial Ontology (SPPO), Multi-Prompt Partial Ontology (MPPO), and Single Prompt Full Ontology (SPFO). We summarize the task and evaluation criteria, present the model pipeline descriptions, and include the LLM prompts (with shared text de-duplicated).

\subsection{Task and evaluation setup}

The core classification task is to map each AI application to the \emph{most specific ontology activity whose scope fully covers the application's primary activity} (the verb-object action the application performs or helps perform).

All comparisons use a fixed ontology snapshot so that models are evaluated against the same set of node titles and definitions. We define a node as a \emph{leaf} if it has no specializations in the snapshot, and as \emph{near-leaf} if all of its direct children are leaves. We define a \emph{hallucination} as any output whose predicted node title does not exactly match a node title in the snapshot. Unless otherwise noted, evaluations use a shared pilot set of applications and the same base model (GPT-5.1 with high-reasoning-effort) so that differences reflect pipeline design rather than underlying model choice.

In addition to the agreement metrics reported in the main text (Wu-Palmer similarity and weighted Cohen's $\kappa$; Appendix~\ref{ap:iaa_detail}), we tracked three practical criteria that affect usefulness at scale:
\begin{itemize}
    \item \textbf{Leaf specificity}: the fraction of predictions that are leaves or near-leaves.
    \item \textbf{Hallucinations}: the rate of outputs that are not valid ontology nodes under the fixed snapshot.
    \item \textbf{Cost and latency}: prompt + completion tokens and wall-clock runtime per application under a consistent batching and retry policy.
\end{itemize}

We also varied two controllable parameters:
\begin{itemize}
    \item \textbf{Retrieval depth (top-$k$)}: the number of candidate nodes returned by vector search (e.g., 10, 25, 50, 100).
    \item \textbf{Context window}: how much node metadata (title, definition, synonyms, parent/child links) is provided to the model when selecting a node.
\end{itemize}

\subsection{Model architecture descriptions}
\autoref{tab:appendix_architecture_summary} summarizes the three evaluated pipeline variants. For the human evaluations, both the SPPO and MPPO models retrieved a shortlist of 100 candidate nodes (k=100).

\begin{table}[h]
    \centering
    \caption{
        \textbf{Summary of automated classification pipeline variants evaluated in the human validation study.}
        All variants used the same base model (GPT-5.1, high-reasoning-effort).
    }
    \footnotesize
    \begin{tabularx}{\textwidth}{@{}l p{2.4cm} p{2.0cm} X@{}}
    \toprule
    \textbf{Variant} & \textbf{Ontology context} & \textbf{Model calls} & \textbf{Summary} \\
    \midrule
    SPPO (Single Prompt Partial Ontology) & Retrieved shortlist (top-$k$) & 1 & Retrieve candidate nodes using application text; one prompt extracts the verb-object activity and selects the best node from the shortlist. \\
    MPPO (Multi-Prompt Partial Ontology) & Retrieved shortlist (top-$k$) & 2 & First prompt extracts a clean verb-object activity; retrieval uses this activity; second prompt selects the best node from the shortlist. \\
    SPFO (Selected) & Full ontology (cached) & 1 & Cache the full ontology and use one prompt to extract the verb-object activity and select the best node from the full hierarchy (no retrieval). \\
    \bottomrule
    \end{tabularx}
    \label{tab:appendix_architecture_summary}
\end{table}

\subsubsection{SPPO (Single Prompt Partial Ontology)}
In SPPO, we retrieve a shortlist of candidate ontology nodes using semantic similarity between the application description (title, tagline, and description) and pre-computed node embeddings. We then prompt the model once to (i) identify a verb-object phrase describing the application's primary activity and (ii) select the most appropriate ontology node \emph{from the retrieved shortlist}. This variant is efficient and works well when the correct node appears in the shortlist, but it can fail when retrieval returns the wrong sense for polysemous verbs or when the correct node is missing from the candidates.

\subsubsection{MPPO (Multi-Prompt Partial Ontology)}
MPPO decomposes the task into two LLM calls. The first call extracts a clean verb-object description of the application's primary activity. We then retrieve candidate nodes using semantic similarity to this extracted activity (rather than to the full application description), and the second call selects the most appropriate ontology node from the retrieved shortlist. This decomposition reduced hallucinations and stabilized outputs, but it tended to select higher-level nodes when the extracted activity was generic, because the node-selection step only sees a limited slice of the ontology.

\subsubsection{SPFO (Single Prompt Full Ontology) (selected)}
In SPFO, we remove retrieval entirely and instead cache the full ontology in the model context. The model is prompted once to identify the application's primary verb-object activity and select the most specific matching ontology node from the full hierarchy. In our evaluation, this variant achieved the best overall tradeoff between agreement with human classifications, leaf specificity, hallucinations, and cost (with prompt caching making full-context inference feasible at scale), so it is the variant used in the final pipeline.

\subsection{LLM prompts}
\label{ap:llm_prompts}
SPPO and SPFO use the same base classification prompt; the main difference is the content of the \texttt{Ontology Nodes} field (retrieved shortlist vs.\ full cached ontology). MPPO adds an activity-extraction prompt as a first step, and its node-selection step uses the same base prompt with a small modification to the input (it includes the previously extracted activity) and output (it returns only the selected node and rationale).

\subsubsection{Base classification prompt (SPPO and SPFO)}

\notebox{
\#\# Role:

You are an analyst who classifies a software application according to: (a) the main activity it performs or helps perform, represented as a "verb + object" phrase, and (b) which of the nodes in the ontology (provided as a JSON structure in the input) is the best classification for the main activity. Work only with the supplied nodes and their fields; do not infer or invent nodes or properties. 

\#\# Ontology Definition:

In our ontology, every node represents a work-related activity. The ontology is represented as a long JSON structure, in which the title of each node is a key and its value is a sub-ontology under that node. 

If a node has many direct children, we group those children in collections. These collection names are also represented as keys in the JSON structure, but they are always in square brackets.

\#\# Output:

Return a single JSON object only (no prose), with exactly these keys and value types:

\{

  "main\_activity": "The single 'base-form verb + object' describing the main activity",
  
  "reasoning\_main\_activity": "Explain your reasoning for main\_activity. If info is sparse or ambiguous, make the best-supported choice and note low confidence in "reasoning" fields.",
  
  "most\_appropriate\_node": "title of the ontology node",

  "most\_appropriate\_node\_rationale": "your reasoning for choosing this ontology node"

\}

\#\# Constraints:

- Output must be valid JSON: double quotes around all strings, no trailing commas, no extra keys or text.

- Always include one item in "main\_activity".

- Use base verb forms plus their direct object in "main\_activity" (e.g., "write code", "conduct research", "summarize text").

- To classify the main activity of the application: 
choose the most specific node whose scope fully covers the main activity. 

- These examples illustrate how to apply this principle:

  1- For an app for automobile driving that plans a route to the user’s destination: 
  
     The most specific node whose scope fully covers this app would be something like ``plan route.'' This node covers all the detailed navigational decisions the app makes, but it is more specific than a node like ``drive car.'' 

}

\notebox{
  2- For a ``self-driving'' car app that steers the car, accelerates, brakes, etc., but still requires a human to be behind the wheel and ready to take over in unusual situations:
  
      The most specific node whose scope fully covers this app would be something like ``drive car.'' There is no other simple description that fully covers all the actions this app takes that is more specific than ``drive car.''

  3- For an image generation app that takes a verbal description of a landscape and generates an image of a landscape based on this description:
  
      A node like ``generate landscape image'' would fully cover the scope of this activity even though in practice, a user might try many different verbal descriptions before finding one that captured what the user wanted.

  4- For a cleaning robot that can clean floors, dust furniture, and empty trash cans:
  
      The most specific node that covers all these kinds of cleaning might be something like ``clean rooms.'' But if there isn’t any node like that in the ontology, the best classification might be a node called ``clean.'' This node is very general, but if there is no other node that covers all three activities (clean floors, dust furniture, and empty trash cans) and is more specific than ``clean,'' then ``clean'' might be the best classification.

Note that, in some cases, the above examples illustrate an app performing an entire activity. In other cases, they illustrate an app helping a human perform an activity. But in all cases, we try to choose the most specific activity that fully covers the scope of what the app does.

\#\# Procedure:

Internally follow this process:

1. From the provided application tagline and description, identify the main activity the application performs or helps perform, represented as a "verb + object" phrase. 

2. Internally, analyze the "verb + object" phrase and compare it to every node in the ontology. For each node, consider all its information.

3. Select the node that best satisfies the classification principle and examples above.

5. Produce the output JSON exactly as specified.

\#\# Input:

\#\#\# Ontology Nodes:

(THE NESTED ONTOLOGY GOES HERE.)

\#\#\# Application Title:

(THE APP TITLE GOES HERE.)

\#\#\# Application Tagline:

(THE APP TAGLINE GOES HERE.)

\#\#\# Application Description:

(THE APP DESCRIPTION GOES HERE.)
}

\subsubsection{MPPO step 1 prompt: activity extraction}

\notebox{
\#\# Role:

You are an analyst that describes a software application according to: (a) what main substantive activity it performs or helps perform, represented as a "verb + object" phrase, and (b) whether it performs the whole activity itself or helps a human perform the activity.

\#\# Input:

- Application Title: (THE APP TITLE GOES HERE.)

- Application Tagline: (THE APP TAGLINE GOES HERE.)

- Application Description: (THE APP DESCRIPTION GOES HERE.)

\#\# Output:

Return a single JSON object only (no prose), exactly with these keys and value types:

\{

  "main\_activity": "The single 'base-form verb + object' describing the main activity",
  
  "reasoning\_main\_activity": "Explain your reasoning for main\_activity. If info is sparse or ambiguous, make the best-supported choice and note low confidence in "reasoning" fields.",

\}

\#\# Constraints:

- Output must be valid JSON: double quotes around all strings, no trailing commas, no extra keys or text.

- Always include one item in "main\_activity".

- Use base verb forms plus their direct object in "main\_activity" (e.g., "write code", "conduct research", "summarize text").

\#\# Procedure:

Internally follow this process:

1. From the provided application tagline and description, identify the main activity it performs or helps perform, represented as a "verb + object" phrase. 

2. Specify whether it performs the whole activity itself or helps a human perform the activity. 
}

\subsubsection{MPPO step 2 prompt: node selection (differences from base prompt)}
MPPO's node-selection step uses the base classification prompt above, with two modifications:
\begin{itemize}
    \item The \textbf{output} is restricted to the selected node and rationale: \texttt{most\_appropriate\_node} and \texttt{most\_appropriate\_node\_rationale}.
    \item The \textbf{input} includes the substantive activity extracted in step 1, and the \texttt{Ontology Nodes} field contains a retrieved shortlist (top-$k$) rather than the full cached ontology.
\end{itemize}

\notebox{
\#\# Output (override):

Return a single JSON object only (no prose), with exactly these keys and value types:

\{

  "most\_appropriate\_node": "title of the ontology node",

  "most\_appropriate\_node\_rationale": "your reasoning for choosing this ontology node"

\}

\#\# Input (override):

In addition to the application fields, include the extracted activity:

- Substantive Activity: (SUBSTANTIVE ACTIVITY GOES HERE) because (SUBSTANTIVE ACTIVITY REASONING GOES HERE)

And set \#\#\# Ontology Nodes to the retrieved shortlist (top-$k$) rather than the full cached ontology.

All other instructions and constraints remain identical to the base classification prompt.
}

\subsection{Summary of lessons}

Across these variants, retrieval was the dominant bottleneck: if the correct node was missing or drowned in noisy candidates, downstream selection could not recover. Adding richer node context helped when the model could reliably absorb it, and decomposition improved stability at the expense of specificity. The full-context SPFO variant provided the best overall tradeoff between agreement with human judgments, specificity, hallucination rate, and cost, which is why it is used in the final pipeline.

\clearpage
\section{Additional details on human classification process}
\subsection{Instructions to human annotators}
\label{ap:human_model_prompts}

We provided human annotators with detailed, structured instructions aligned with the prompts used for the large language models (LLMs) performing the automated annotations. This parallel design ensured conceptual consistency between human and automated annotation procedures and facilitated comparability across annotations from LLMs and humans.

\begin{itemize}
    \item \textbf{Role}
You are an analyst who classifies a software application according to: (a) the main activity it performs or helps perform, represented as a ``verb + object'' phrase, and (b) which of the nodes in the ontology is the best classification for the main activity. Work only with the nodes included in the platform and their fields; do not infer or invent nodes or properties. 
    \item \textbf{Ontology Definition}
In our ontology, every node represents a work-related activity. The ontology is organized as a hierarchical tree structure that represents tasks across a wide range of occupations. It uses verbs and nouns to describe these activities. It includes broad nodes (e.g., ``Create information''), more specific nodes (e.g., ``Analyze''), and leaf nodes, the most specific level of the activities (e.g. ``Search literature'')
    \item \textbf{Input} You receive a set of applications with the following information: 
    \begin{itemize}
        \item Name of the application
        \item Application tagline
          \item Application Description
    \end{itemize}  
You have access to the platform of the ontology where you find the following information: 
\begin{itemize}
    \item The ontology nodes
     \item A search engine to navigate the ontology
      \item A field with the description of each node and its generalizations and specializations
\end{itemize} 
    \item \textbf{Output} 
    \begin{itemize}
        \item Main activity: The single base-form of [verb + object] describing the main activity
        \item  Reasoning of the main activity: Explain your reasoning for the main activity. If info is sparse or ambiguous, make the best-supported choice and note low confidence in ``reasoning'' fields.
        \item Most appropriate node: Choose the most appropriate node considering the constraints and the procedure described below
        \item  Most appropriate node rationale: Your reasoning for choosing this ontology node
    \end{itemize}

    \item \textbf{Constraints}
    \begin{itemize}
    \item Use base verb forms plus their direct object to describe the main activity (e.g., ``Write code,'' ``Conduct research,'' ``Summarize text'').
    \item To classify the main activity of the application, apply the following principle:
\item Choose the most specific node whose scope fully covers the main activity.

\end{itemize}

    \item \textbf{Examples (applying the classification principle)}
    \begin{enumerate}
    \item \textbf{Route-planning application for automobile driving.}  
    The most specific node whose scope fully covers this app would be something like ``Plan route.'' This node covers the detailed navigational decisions the application makes and is more specific than a node such as ``Drive car.''

    \item \textbf{``Self-driving'' car application that steers, accelerates, and brakes, but still requires a human to supervise and take over.}  
    The most specific node whose scope fully covers this application would be something like ``Drive car.'' There is no simpler description that fully covers all the actions the application takes that is more specific than ``Drive car.''

    \item \textbf{Image generation application that takes a verbal description and generates a landscape image.}  
    A node such as ``Generate landscape image'' would fully cover the scope of this activity, even if users iterate over multiple prompts before achieving the desired output.

    \item \textbf{Cleaning robot that cleans floors, dusts furniture, and empties trash cans.}  
    The most specific node covering all these actions might be ``Clean rooms.'' If no such node exists in the ontology, the best available classification may be the more general node ``Clean.'' If no other node covers all three activities (clean floors, dust furniture, and empty trash cans) while being more specific than ``Clean,'' then ``Clean'' is the appropriate choice.
\end{enumerate}
    \item \textbf{Note} In some cases, the above examples illustrate an application performing an entire activity. In other cases, they illustrate an application helping a human perform an activity. But in all cases, we try to choose the most specific activity that fully covers the scope of what the application does.
    \item \textbf{Procedure} Follow this process:
\begin{enumerate}
    \item \textbf{Identify the main activity.} From the application tagline and description, determine the main activity the application performs or helps perform, expressed as a ``verb + object'' phrase.
    \item \textbf{Search the ontology.} Using the identified ``verb + object'' phrase, search or browse the ontology to locate nodes that may be appropriate classifications.
    \item \textbf{Select the most appropriate node.} Choose the node that best satisfies the classification principle and aligns with the examples above.
\end{enumerate}
\end{itemize}

\subsection{Inter-annotator agreements details}
\label{ap:iaa_detail}
Traditional inter-annotator agreement measures (like Cohen's $\kappa$) treat all disagreements equally; classifying an image generation application as ``Write'' versus ``Create image'' counts the same as classifying it as ``Write'' versus ``Clean floor''. But in a hierarchical ontology, some disagreements are more severe than others. Wu-Palmer similarity captures this by measuring how much of the path from the root two classifications share: siblings score higher than cousins, which score higher than activities in entirely different branches. This measure is defined by $S = \frac{2N}{N_1+N_2}$, where $S$ is the similarity between the two nodes, $N$ is the distance from the root node of the hierarchy to the closest common ancestor of the two nodes, and $N_1$ and $N_2$ are the distances from the root node to the two nodes being compared. In other words, it is the proportion of links from the root node that the two nodes share. 

We report (1) mean Wu-Palmer similarity over all pairs of ratings, and (2) weighted Cohen’s $\kappa$, computed by incorporating Wu-Palmer similarity as the weighting function. For human-human IAA, we calculated the mean pairwise Wu-Palmer similarity and weighted Cohen’s $\kappa$ among all human annotators. For human-AI IAA, we calculated these metrics between each model and all human annotators. For AI-AI IAA, we calculated the same pairwise measures across the three models.

We also invited three student interns from our research team to conduct the human classification following the same procedure described in Section~\ref{sec:human_evaluation}. These classifiers can be regarded as ``junior classifiers'' who were entry-level on this annotation task and less experienced than the seven authors, who are the core expert classifiers.~\autoref{tab:iaa_results} reports the human-AI, human-human, and AI-AI inter-annotator agreement results for all three groups: junior annotators (interns), expert annotators, and all annotators combined. Notably, there is a clear increase in both human-human and human-AI agreement when comparing junior annotators with expert annotators. This suggests that the ontology captures specialized organizational knowledge that requires domain training to apply reliably.

\begin{table}[h]
    \centering
    \caption{
        \textbf{Inter-annotator agreement (IAA) across human annotation groups,} using Mean Wu-Palmer Similarity and Weighted Cohen’s $\kappa$ (mean with 95\% confidence intervals).
    }
    \label{tab:iaa_results}
    \begin{tabular*}{\textwidth}{@{\extracolsep\fill}p{2.2cm}p{1cm}p{1cm}p{1cm}p{1cm}p{1cm}p{1cm}}
    \toprule
    & \multicolumn{3}{c}{Mean Wu-Palmer Similarity}
    & \multicolumn{3}{c}{Weighted Cohen's $\kappa$} \\
    \cmidrule(lr){2-4} \cmidrule(lr){5-7}
    & Interns & Experts & All
    & Interns & Experts & All \\
    \midrule
    Human-SPFO
    & 0.591 [0.518, 0.666]
    & 0.631 [0.561, 0.703]
    & 0.619 [0.559, 0.680]
    & 0.347 [0.233, 0.456]
    & 0.411 [0.304, 0.509]
    & 0.392 [0.296, 0.476] \\
    
    Human-SPPO
    & 0.583 [0.504, 0.660]
    & 0.585 [0.511, 0.655]
    & 0.584 [0.519, 0.645]
    & 0.343 [0.225, 0.451]
    & 0.349 [0.242, 0.444]
    & 0.347 [0.254, 0.427] \\
    
    Human-MPPO
    & 0.554 [0.473, 0.635]
    & 0.619 [0.543, 0.697]
    & 0.600 [0.528, 0.670]
    & 0.314 [0.195, 0.421]
    & 0.412 [0.304, 0.514]
    & 0.383 [0.282, 0.470] \\
    
    Human-Human
    & 0.525 [0.465, 0.586]
    & 0.584 [0.525, 0.649]
    & 0.557 [0.514, 0.609]
    & 0.258 [0.161, 0.349]
    & 0.348 [0.260, 0.433]
    & 0.307 [0.234, 0.372] \\
    
    AI-AI
    & \multicolumn{3}{c}{0.812 [0.726, 0.895]}
    & \multicolumn{3}{c}{0.703 [0.566, 0.822]} \\
    \bottomrule
    \end{tabular*}
\end{table}

\subsection{Relation between human-AI classification similarity and human evaluation of model classifications}

\begin{figure}[h]
    \centering
    \includegraphics[width=0.55\textwidth]{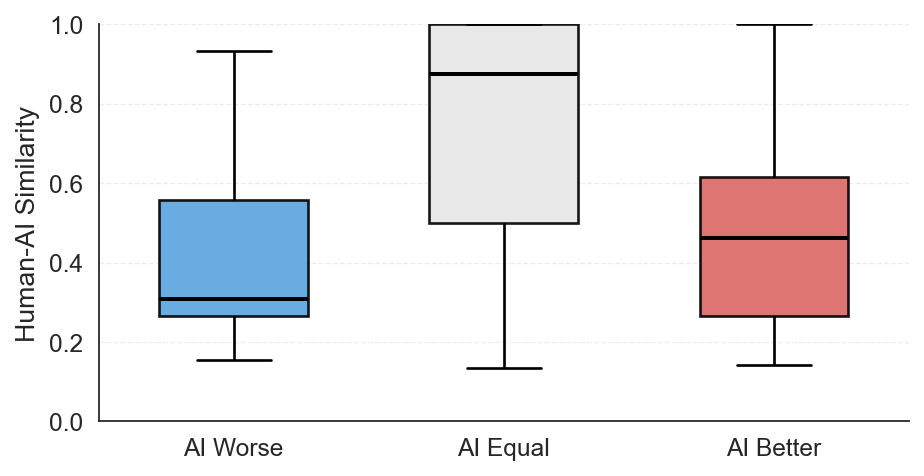}
    \caption{
        \textbf{Distribution of human-AI annotation similarity (mean Wu-Palmer similarity) across rating categories.}
        When human evaluators rated the AI’s classification as equal to their own, the corresponding human-AI annotation pairs showed higher similarity. In contrast, when the AI’s classification was rated as better or worse than the human’s, annotation similarity was lower.
    }
    \label{fig:ap_relation}
\end{figure}

We plot the distribution of human-AI annotation similarity for each manual rating category. The pattern shown in~\autoref{fig:ap_relation} reconfirms that manual evaluation labels align with the hierarchy-aware similarity measure: perceived differences\textemdash whether favorable or unfavorable\textemdash are associated with greater divergence in the ontology.

\clearpage
\section{Additional AI software application results}

\subsection{Sample TAAFT data}
\label{ap:taaft:sample}

\begin{table}[ht]
\centering
\caption{
    \textbf{Sample AI application from the TAAFT dataset.}
}
\label{tab:taaft:sample}
\begin{tabular}{p{3.0cm} p{7.5cm}} 
    \toprule
    \textbf{Name} & \href{https://theresanaiforthat.com/ai/facts/}{\&facts} \\ 
    \midrule
    \textbf{Link} & \url{https://www.andfacts.com/} \\
    \midrule
    \textbf{Type} & AI \\ 
    \midrule
    \textbf{Launch Date} & 10-Jun-23 \\ 
    \midrule
    \textbf{Task} & Market analysis \\ 
    \midrule
    \textbf{Tagline} & Analyze market signals for real-time consumer insights. \\ 
    \midrule
    \textbf{Description} & 
    \&facts is a market intelligence platform for consumer brands that analyzes real-time market and product signals to identify trends, opportunities, and risks. It supports product development, marketing optimization, competitor analysis, and international expansion by providing AI-driven insights into consumer preferences and behavior without the need for extensive manual research or consultants. \\ 
    \midrule
    \textbf{Pricing Model} & Paid-only \\ 
    \midrule
    \textbf{Starting Price} & \$199 \\ 
    \midrule
    \textbf{Billing Frequency} & One-time \\ 
    \midrule
    \textbf{Number of Saves} & 3 \\
    \bottomrule
\end{tabular}
\end{table}

\clearpage

\subsection{Sample TAAFT classification results}
\label{ap:taaft:classification}

\begin{table}[ht]
    \centering
    \caption{
        \textbf{Examples of AI applications and their classification results.}
    }
    \label{tab:taaft:classification}
    \begin{tabular}{>{\hspace{0pt}}m{0.1\linewidth}>{\hspace{0pt}}m{0.1\linewidth}>{\hspace{0pt}}m{0.1\linewidth}>{\hspace{0pt}}m{0.6\linewidth}>{\hspace{0pt}}m{0.1\linewidth}} 
    \toprule
    \textbf{Name} & \textbf{Launch Date} & \textbf{Tagline} & \textbf{Description} & \textbf{Classification} \\ 
    \midrule
    \&AI & 1-Jan-24 & AI-powered patent due diligence in seconds & \&AI streamlines patent due diligence by rapidly generating detailed claim charts, performing prior art searches, and analyzing claim robustness. The platform supports exporting results to Word or Excel, semantic search across proprietary and USPTO documents, and interactive querying through a patent-aware chatbot. Its goal is to reduce the time and cost associated with patent analysis while improving coverage and reliability. & Research (Information) \\ 
    \midrule
    \&facts & 10-Jun-23 & Analyze market signals for real-time consumer insights. & \&facts is a market intelligence platform for consumer brands that analyzes real-time market and product signals to identify trends, opportunities, and risks. It supports product development, marketing optimization, competitor analysis, and international expansion by providing AI-driven insights into consumer preferences and behavior without the need for extensive manual research or consultants. & Analyze Market \\ 
    \midrule
    008 Agent & 22-Nov-23 & AI-powered open-source softphone revolutionizing VoIP. & 008 Agent is an open-source, AI-powered softphone designed for intelligent VoIP communication. It captures and processes call events, transcribes conversations, performs sentiment analysis, and generates summaries. The system integrates with CRM tools, supports programmable conversational agents, and provides transparent, customizable workflows through a community-driven development model. & Communicate (transfer information end-to-end) \\ 
    \midrule
    04-x & 1-Aug-23 & Privacy-first chat with powerful AI models & 04-x is a privacy-focused chat platform for interacting with large language models. It supports multiple commercial and open-source models, encrypts all user data end-to-end, and allows tasks such as conversational interaction, summarization, and text generation. The system emphasizes data ownership, security, and extensibility through continuous model integration. & Converse \\ 
    \midrule
    0gpt & 9-Sep-24 & Unmask AI-generated text with precision. & 0gpt is an AI content detection tool that distinguishes between human-written and AI-generated text using linguistic and statistical analysis. It evaluates sentence structure, vocabulary diversity, and contextual coherence, highlights suspected AI-generated segments, supports multiple languages, and ensures privacy by not storing analyzed content. The tool is free and designed for broad accessibility. & Classify Content \\
    \bottomrule
    \end{tabular}
\end{table}
\clearpage

\subsection{Visualization of the complete ontology (aggregated counts)}
\label{ap:taaft:full_ontology}

\begin{figure*}[ht]
    \centering
    \includegraphics[width=\textwidth]{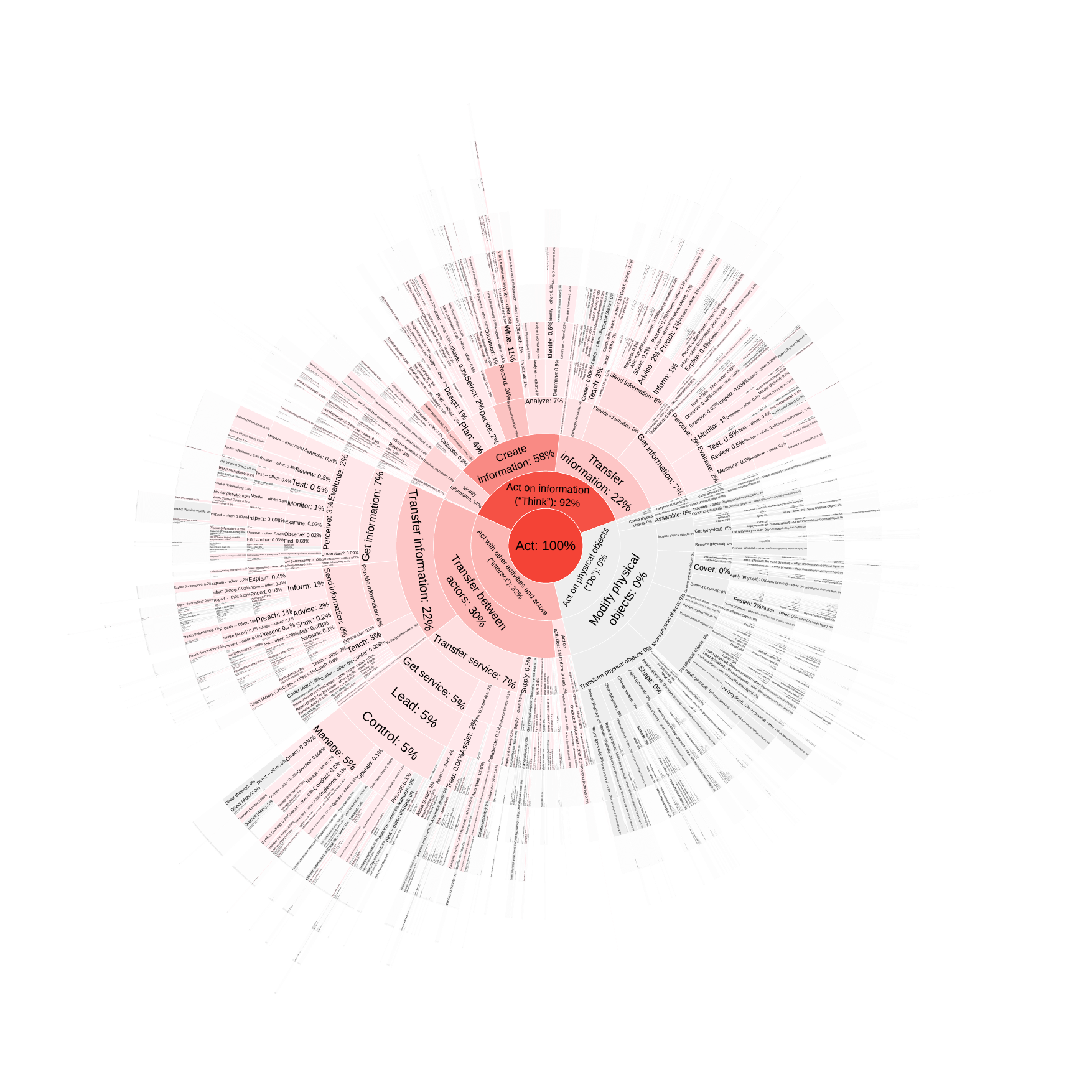}
    \caption{
        \textbf{Sunburst diagram of the complete ontology with \textit{aggregated counts} of AI software applications.} Color intensities and percentage values represent the proportion of AI applications classified directly in each activity \textit{and in all of its specializations}, relative to the total number of AI applications in the dataset.
        Readers are encouraged to zoom in to examine fine-grained specializations. 
    }
\end{figure*}
\clearpage

\subsection{Visualization of the complete ontology (direct counts)}
\label{ap:taaft:full_ontology_direct}

\begin{figure*}[ht]
    \centering
    \includegraphics[width=\textwidth]{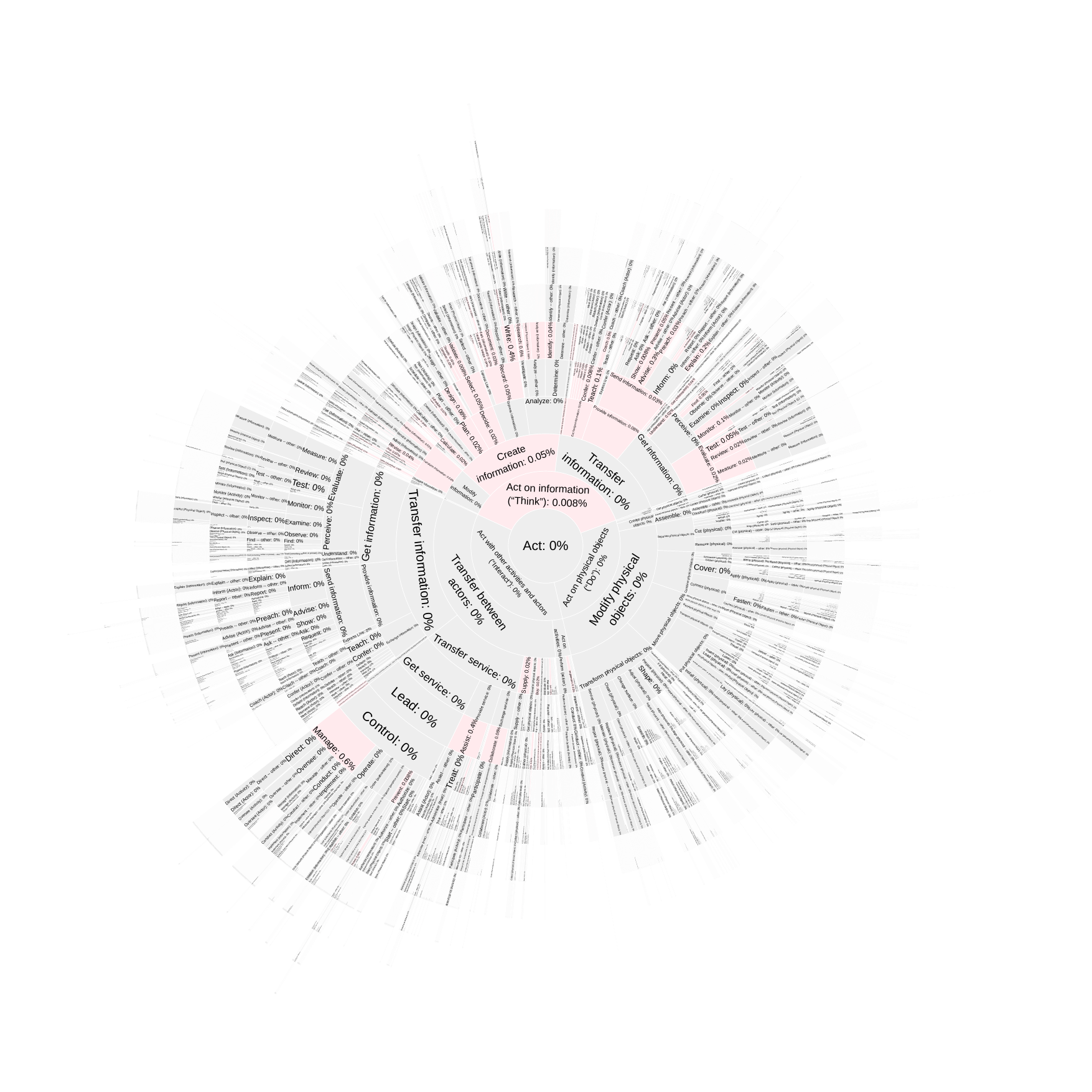}
    \caption{
        \textbf{Sunburst diagram of the complete ontology with \textit{direct counts} of AI software applications.} Color intensities and percentage values represent the proportion of AI applications directly classified \textit{directly}into each activity, relative to the total number of AI applications in the dataset.
        Readers are encouraged to zoom in to examine fine-grained specializations. 
    }
\end{figure*}
\clearpage


\subsection{Mapping atomic activities to associated O*NET tasks}
\label{ap:taaft:atomic_activities}

{\small
\begin{longtable}{
    >{\raggedright\arraybackslash}p{4.4cm}
    >{\raggedright\arraybackslash}p{8.0cm}
}
    \caption
    [Selected atomic activities and associated O*NET tasks in the ``Create Information -- other'' branch]
    {\textbf{Selected atomic activities and associated O*NET tasks in the ``Create information -- other'' branch.}
    \label{tab:taaft_onet_tasks}} \\
    
    \toprule
    \textbf{Atomic Activity} & \textbf{O*NET Tasks} \\
    \midrule
    \endfirsthead
    
    \caption[]{\textbf{Selected atomic activities and associated O*NET tasks in the ``Create Information -- other'' branch.} (continued)} \\
    \toprule
    \textbf{Atomic Activity} & \textbf{O*NET Tasks} \\
    \midrule
    \endhead
    
    \midrule
    \multicolumn{2}{r}{\textit{Continued on next page}} \\
    \midrule
    \endfoot
    
    \bottomrule
    \endlastfoot
    
    
    \textbf{Create content (3.53\%)} 
    & (O*NET) 16193 — Collaborate with Web, multimedia, or art design staffs to create multimedia Web sites or other internet content that conforms to brand and company visual format. \\
    & (O*NET) 5219 — Plan and develop instructional methods and content for educational, vocational, or student activity programs. \\
    \midrule
    
    \textbf{Develop application (1.79\%)} 
    & (O*NET) 16184 — Develop transactional Web applications using Web programming software and programming languages such as HTML and XML. \\
    & (O*NET) 16786 — Develop new software applications or customize existing applications to meet specific scientific project needs. \\
    & (O*NET) 17706 — Develop or maintain applications that process biologically based data into searchable databases for analysis, calculation, or presentation. \\
    & (O*NET) 18093 — Develop, implement, or evaluate health information technology applications, tools, or processes to assist with data management. \\
    & (O*NET) 20193 — Develop software applications or programming for statistical modeling and graphic analysis. \\
    & (O*NET) 21734 — Develop specialized computer software routines, internet-based GIS databases, or business applications to customize geographic information. \\
    & (O*NET) 3572 — Design, develop, and implement computer applications for mining operations such as mine design, modeling, mapping, or condition monitoring. \\
    & (O*NET) 5351 — Plan or develop applications or modifications for electronic properties used in components, products, or systems to improve technical performance. \\
    \midrule
    
    \textbf{Create recording (1.19\%)} &
    (O*NET) 22277 — Create photographic recordings of information using equipment. \\
    & (O*NET) 22572 — Produce recordings of music. \\
    \midrule
    
    \textbf{Generate lead (0.99\%)} &
    (O*NET) 17654 — Generate solar energy customer leads to develop new accounts. \\
    \midrule
    
    \textbf{Create site (0.81\%)} &
    (O*NET) 16193 — Collaborate with Web, multimedia, or art design staffs to create multimedia Web sites or other internet content that conforms to brand and company visual format. \\
    & (O*NET) 20058 — Develop and maintain course Web sites. \\
    & (O*NET) 20288 — Develop or maintain internal or external company Web sites. \\
    & (O*NET) 21049 — Develop Web sites. \\
    & (O*NET) 21109 — Develop and maintain Web sites for online courses. \\
    
\end{longtable}
}

\clearpage

\subsection{Distribution of AI applications (2016--2025)}
\label{ap:taaft:sunbursts}

\begin{figure}[ht]
    \centering
    \includegraphics[width=1.0\linewidth]{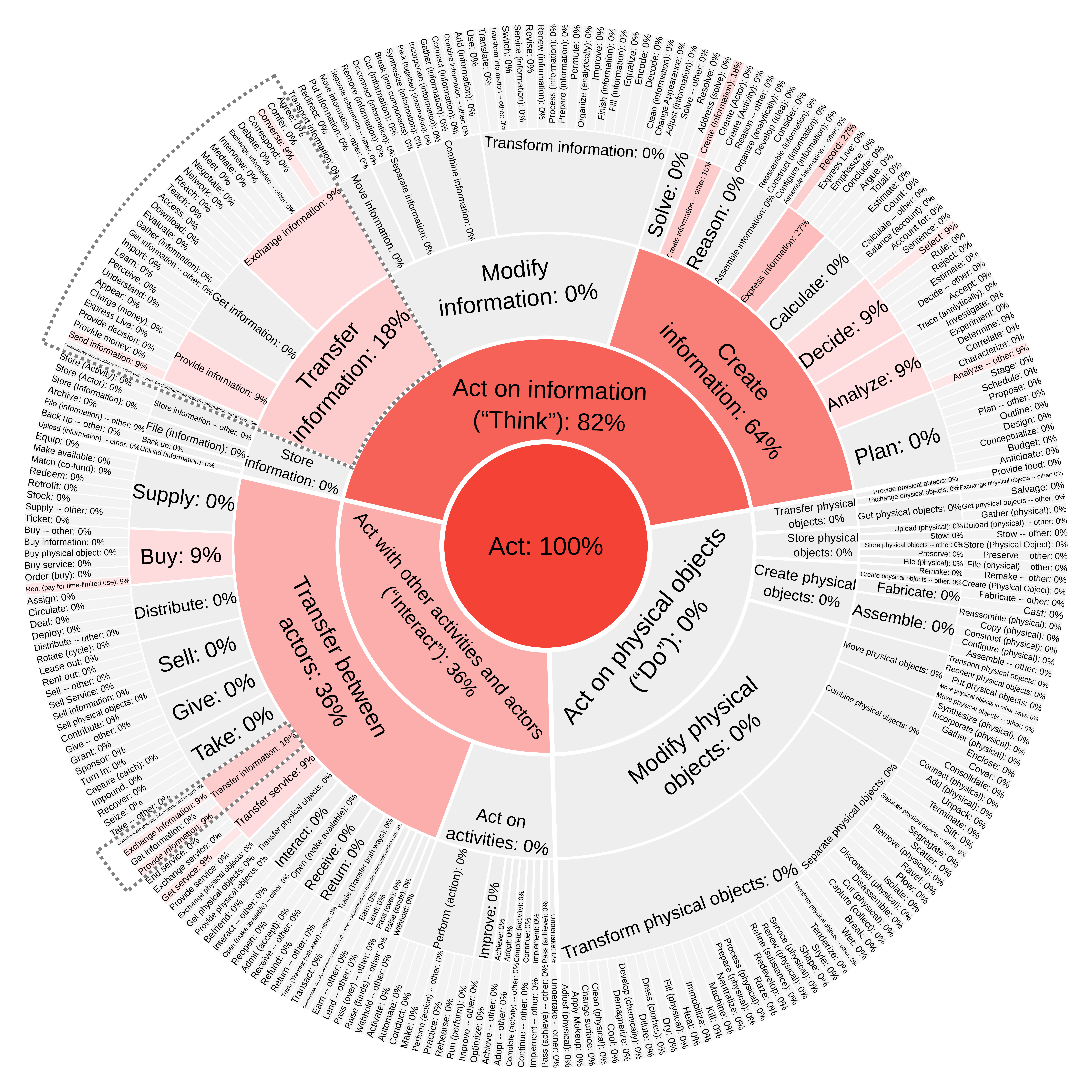}
    \caption{Distribution of AI applications in 2016.}
    \label{fig:taaft_yearly_2016}
\end{figure}

\clearpage
\begin{figure}[ht]
    \centering
    \includegraphics[width=1.0\linewidth]{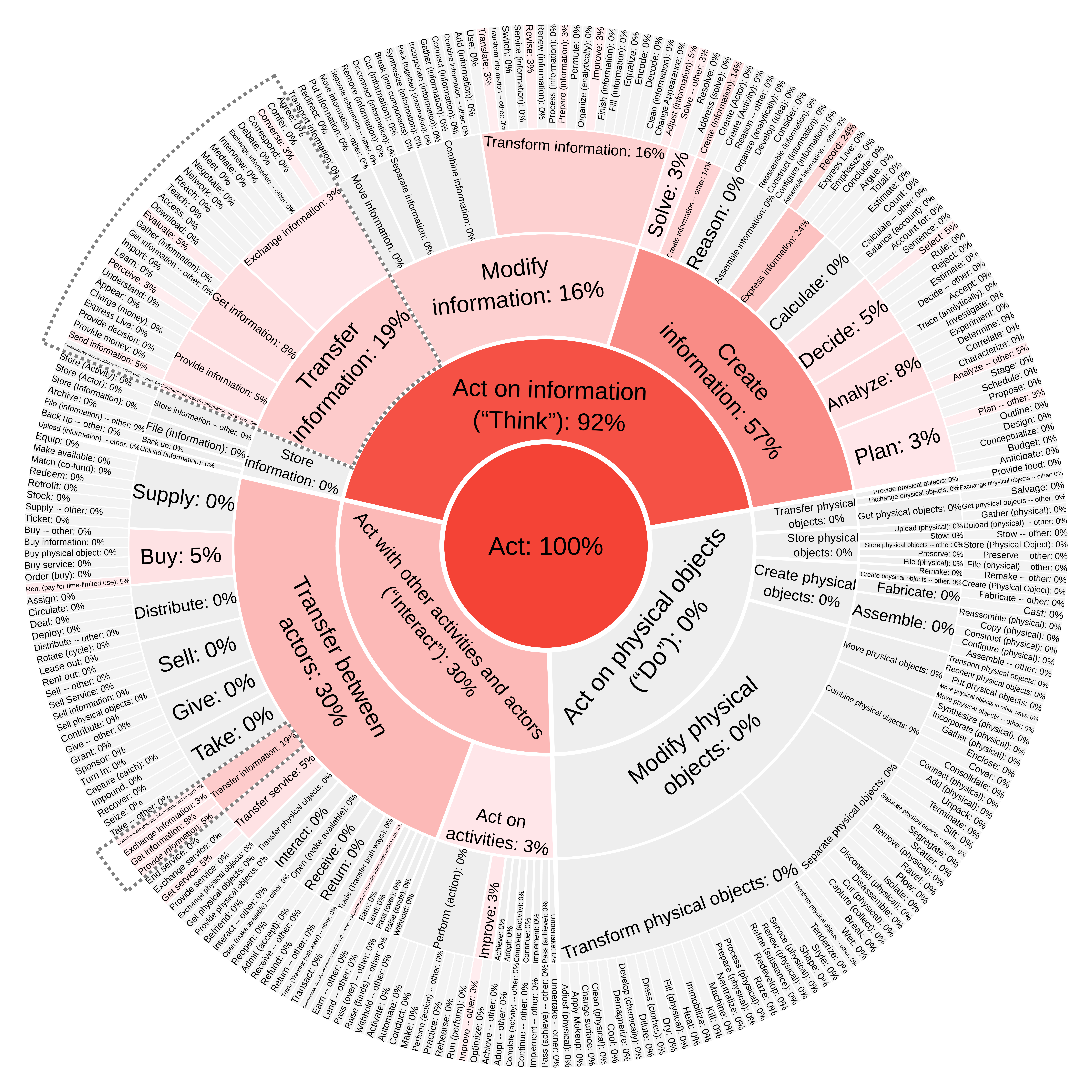}
    \caption{Distribution of AI applications in 2017.}
    \label{fig:taaft_yearly_2017}
\end{figure}

\clearpage
\begin{figure}[ht]
    \centering
    \includegraphics[width=1.0\linewidth]{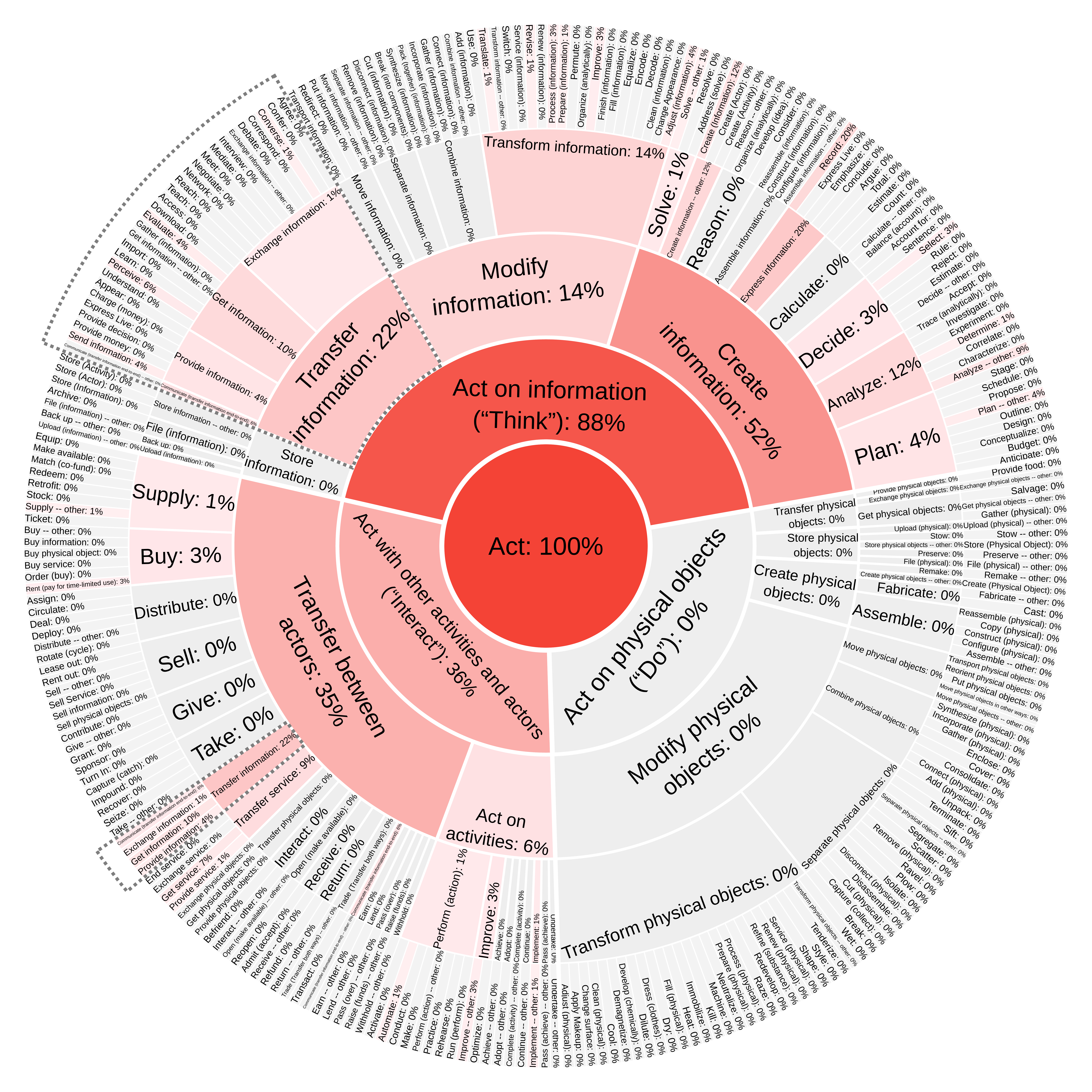}
    \caption{Distribution of AI applications in 2018.}
    \label{fig:taaft_yearly_2018}
\end{figure}

\clearpage
\begin{figure}[ht]
    \centering
    \includegraphics[width=1.0\linewidth]{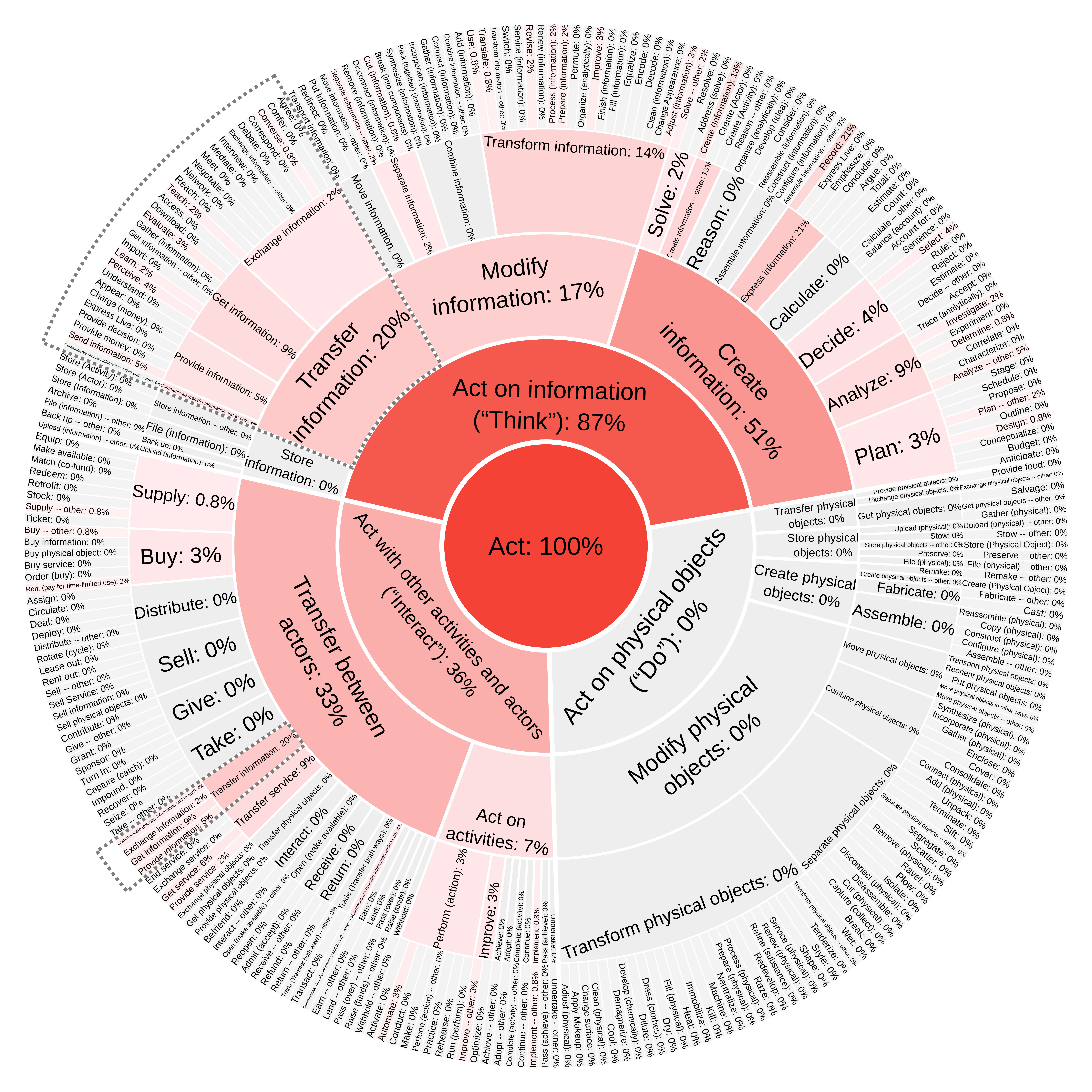}
    \caption{Distribution of AI applications in 2019.}
    \label{fig:taaft_yearly_2019}
\end{figure}

\clearpage
\begin{figure}[ht]
    \centering
    \includegraphics[width=1.0\linewidth]{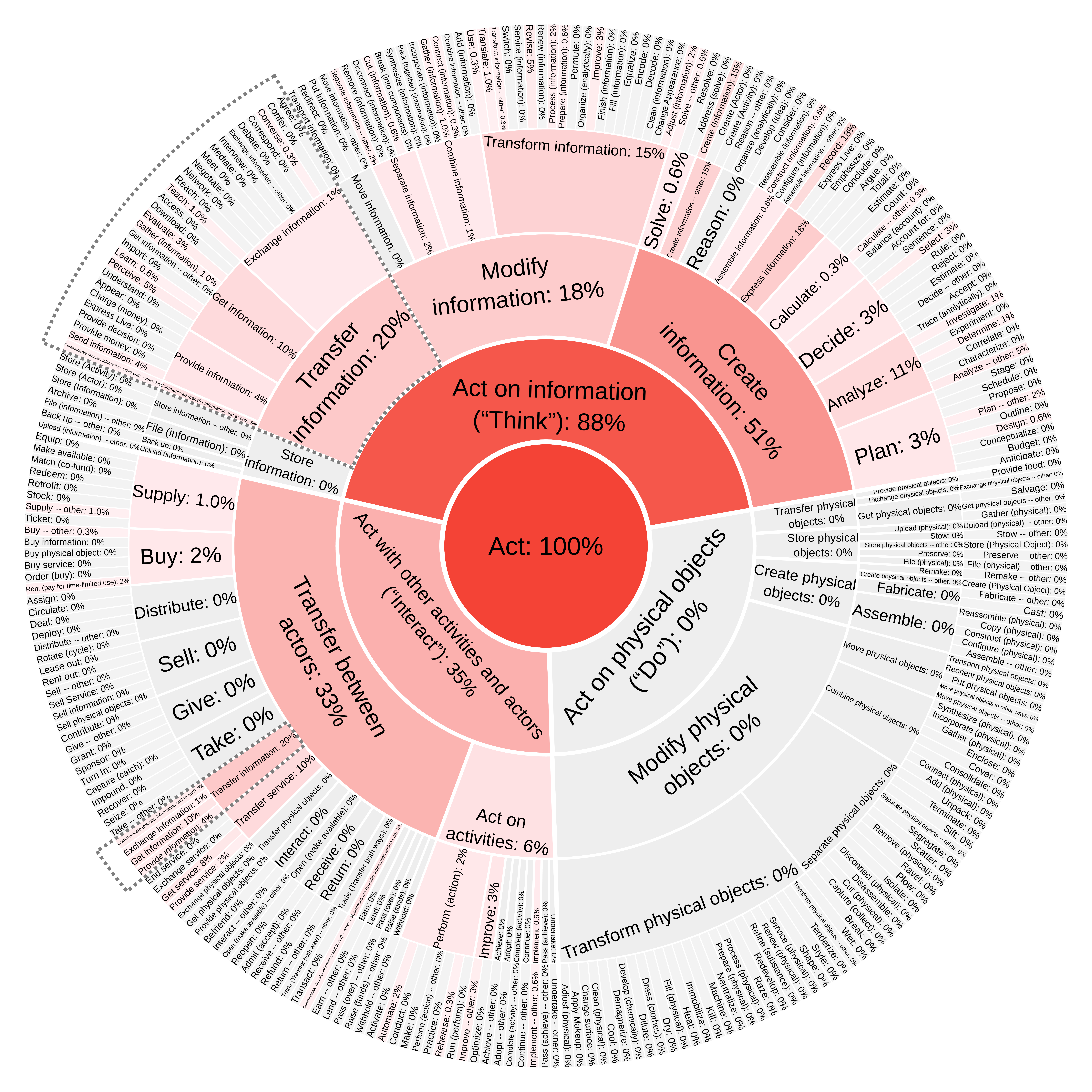}
    \caption{Distribution of AI applications in 2020.}
    \label{fig:taaft_yearly_2020}
\end{figure}

\clearpage
\begin{figure}[ht]
    \centering
    \includegraphics[width=1.0\linewidth]{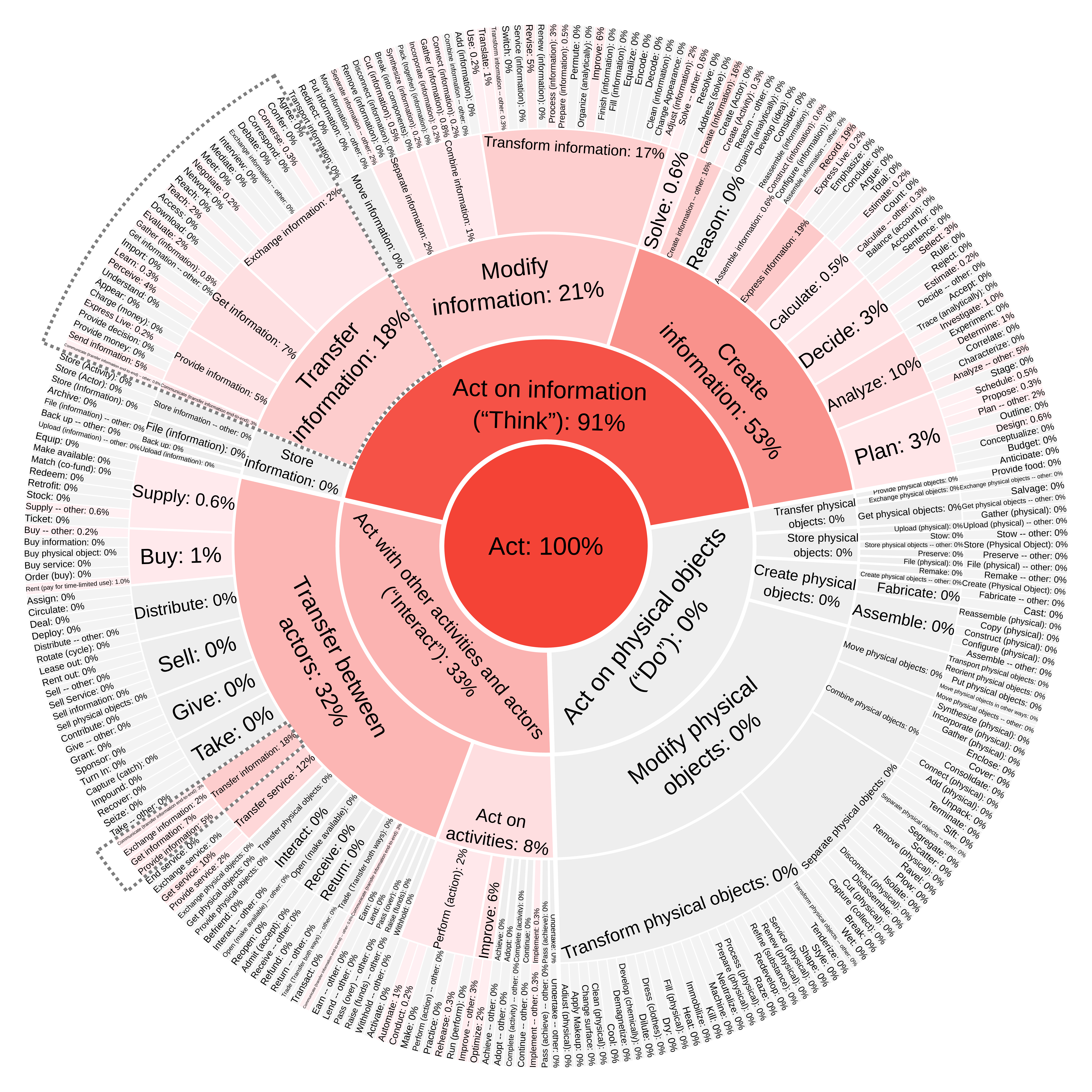}
    \caption{Distribution of AI applications in 2021.}
    \label{fig:taaft_yearly_2021}
\end{figure}

\clearpage
\begin{figure}[ht]
    \centering
    \includegraphics[width=1.0\linewidth]{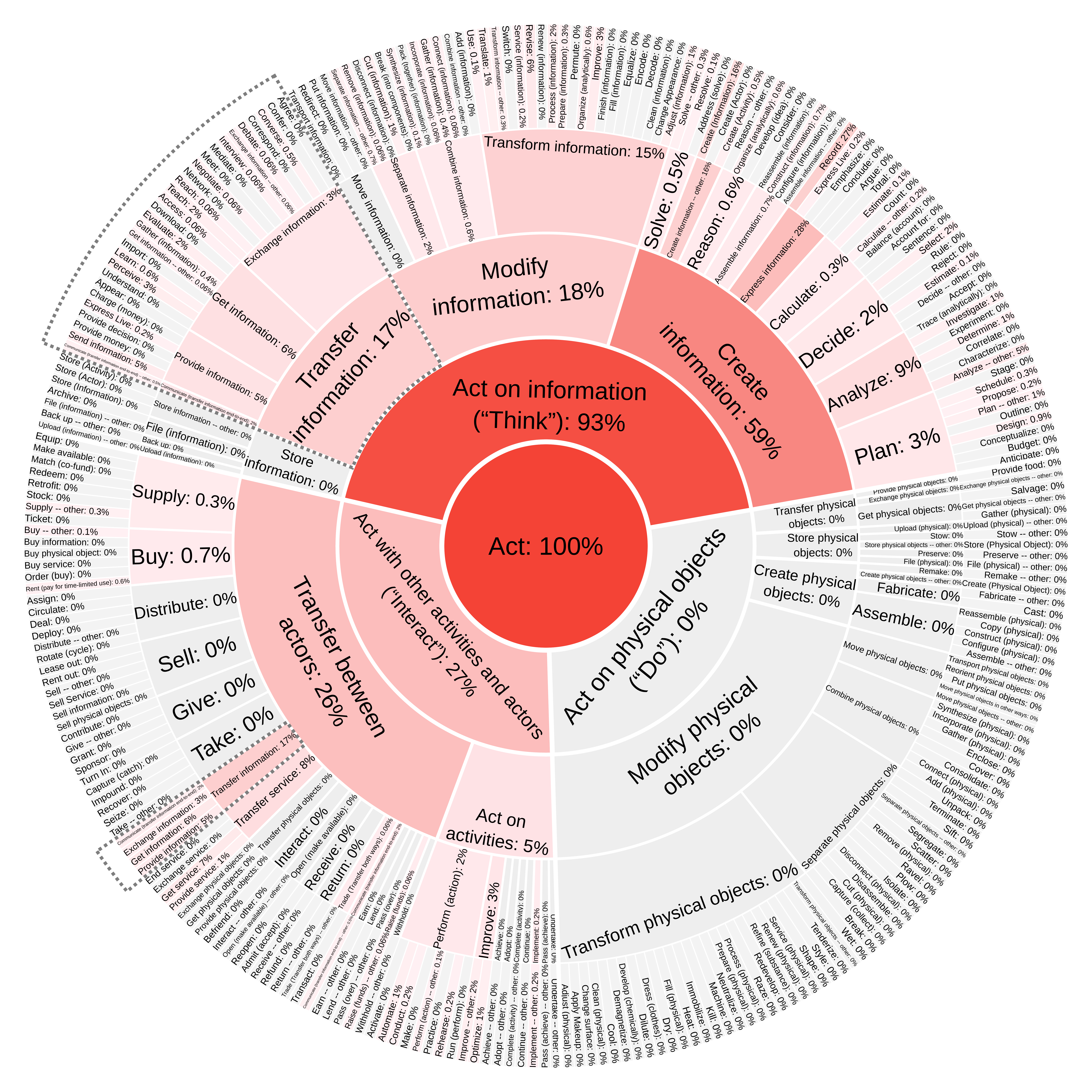}
    \caption{Distribution of AI applications in 2022.}
    \label{fig:taaft_yearly_2022}
\end{figure}

\clearpage
\begin{figure}[ht]
    \centering
    \includegraphics[width=1.0\linewidth]{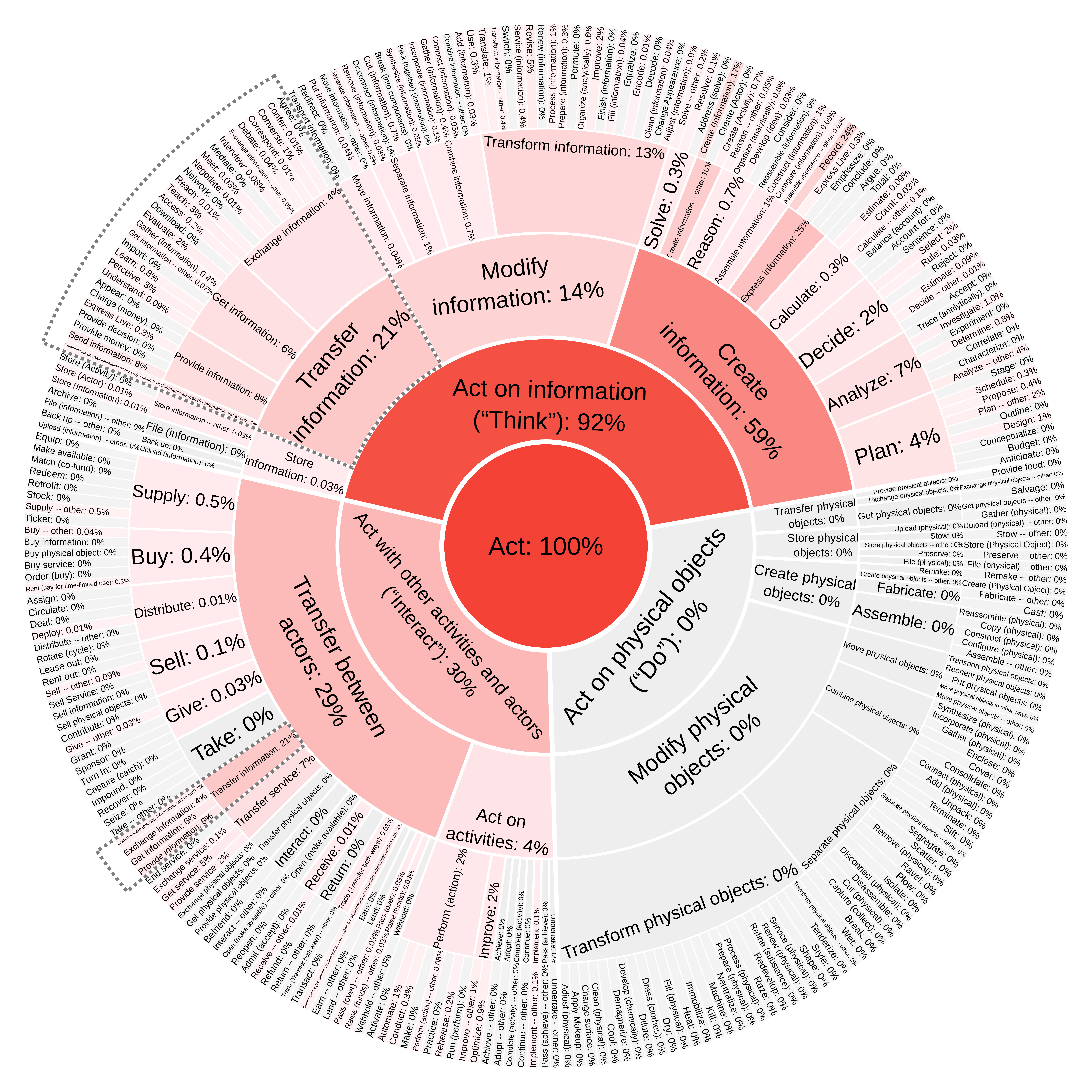}
    \caption{Distribution of AI applications in 2023.}
    \label{fig:taaft_yearly_2023}
\end{figure}

\clearpage
\begin{figure}[ht]
    \centering
    \includegraphics[width=1.0\linewidth]{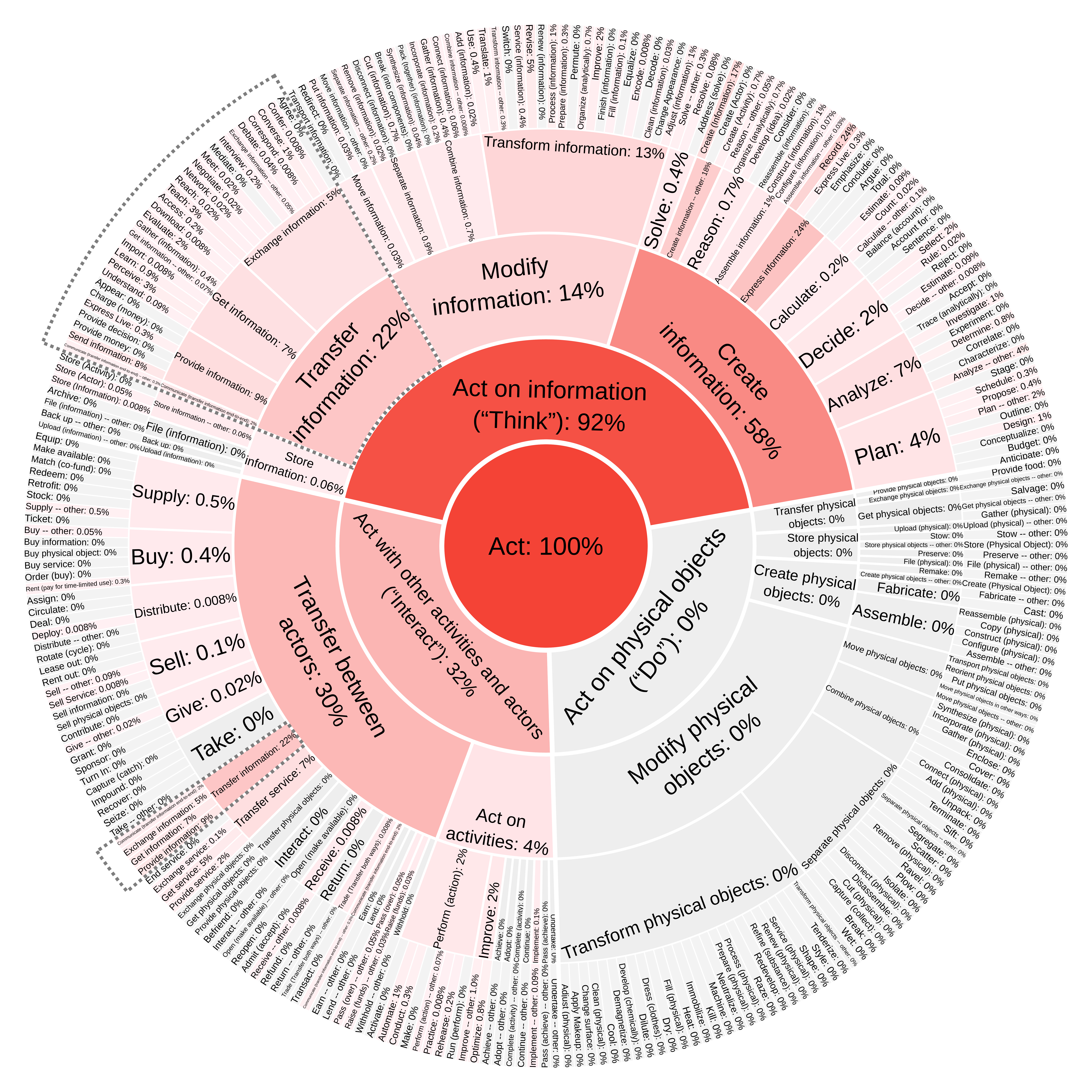}
    \caption{Distribution of AI applications in 2024.}
    \label{fig:taaft_yearly_2024}
\end{figure}

\clearpage
\begin{figure}[ht]
    \centering
    \includegraphics[width=1.0\linewidth]{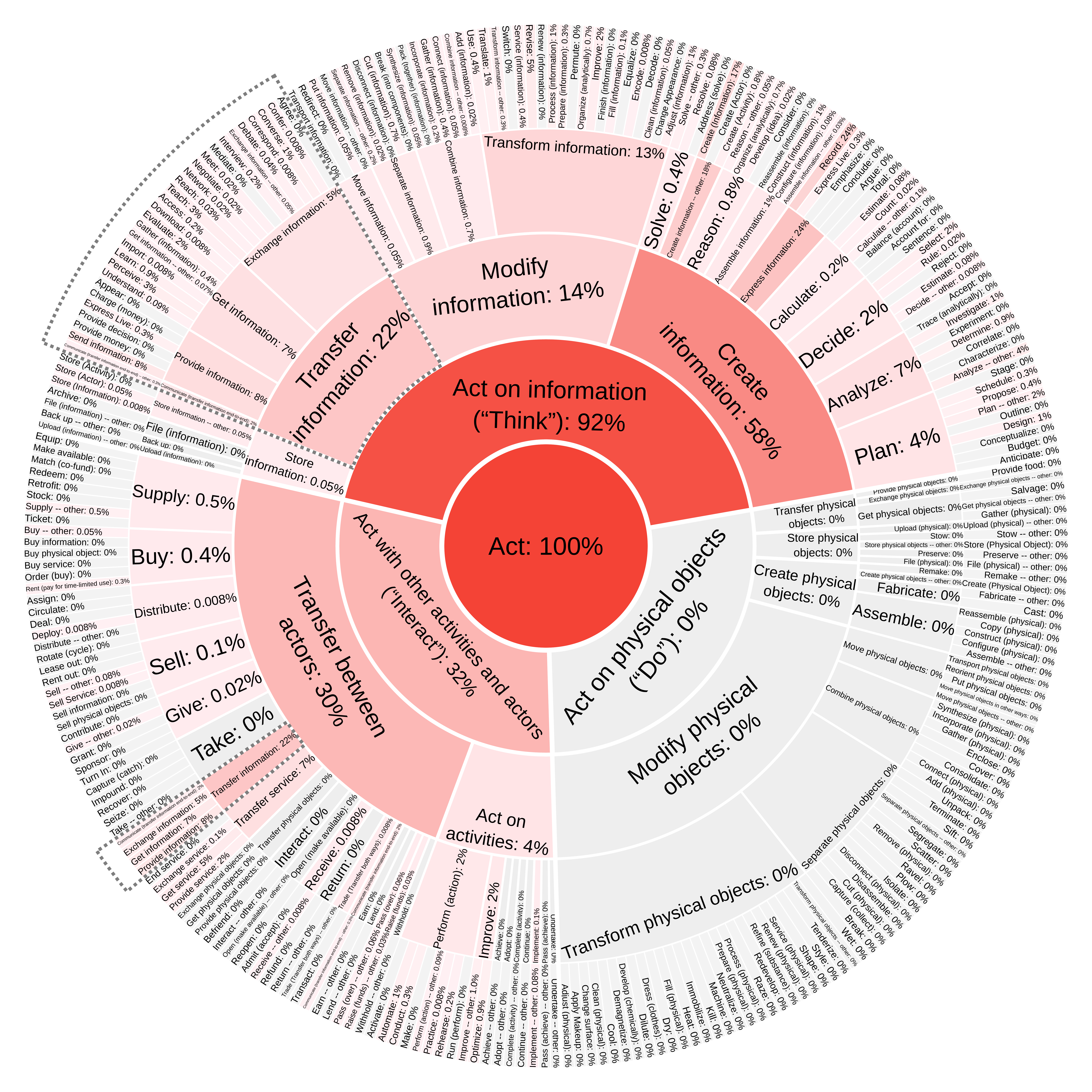}
    \caption{Distribution of AI applications in 2025.}
    \label{fig:taaft_yearly_2025}
\end{figure}

\clearpage


\clearpage
\section{Robotics dataset}
\label{robotics}
\subsection{IFR taxonomy of robotic systems}
\label{ifr taxonomy}

The International Federation of Robotics (IFR) classifies robotic systems into two main categories: industrial and service robots, defined as follows:

\begin{itemize}
    \item  \textit{Industrial robots} defined as ``automatically controlled, reprogrammable, multipurpose manipulators, programmable in three or more axes, which can be either fixed in place or mounted on a mobile platform for use in industrial automation applications'' \cite{IFR2025Industrial} and
    \item \textit{Service robots} defined as robots that ``perform useful tasks for humans or equipment'' outside traditional industrial automation contexts \cite{IFR2025Service}.
\end{itemize}

Service robots are further classified as professional service robots, medical robots and consumer robots. \autoref{tab:ifr_industrial}, \autoref{tab:ifr_service_professional}, \autoref{tab:ifr_medical_prof} and \autoref{tab:ifr_service consumer} present the corresponding taxonomies of classes and subclasses for industrial, professional service, medical and consumer robots, respectively. \autoref{tab:ifr_to_ontology} presents our classification of each subclass onto our ontology.

{\small
\setlength{\extrarowheight}{5pt}
\begin{longtable}{p{2.0cm} p{1.9cm} p{1.9cm} p{1.7cm} p{2.1cm} p{2.1cm}}

\caption
[IFR taxonomy of industrial robots]
{\textbf{IFR taxonomy of industrial robots.} 
Column headings are classes and items within each column are subclasses.
\label{tab:ifr_industrial}} \\

\toprule
\textbf{Handling Operation} 
& \textbf{Welding} 
& \textbf{Dispensing} 
& \textbf{Cutting/ Processing} 
& \textbf{Assembling} 
& \textbf{Others} \\
\midrule
\endfirsthead

\caption[]{\textbf{IFR taxonomy of industrial robots.} (continued)} \\
\toprule
\textbf{Handling Operation} 
& \textbf{Welding} 
& \textbf{Dispensing} 
& \textbf{Cutting/ Processing} 
& \textbf{Assembling} 
& \textbf{Others} \\
\midrule
\endhead

\midrule
\multicolumn{6}{r}{\textit{Continued on next page}} \\
\midrule
\endfoot

\bottomrule
\endlastfoot


Metal casting
& Arc welding
& Painting, enamelling 
& Laser cutting
& Assembling
& Cleanroom for FPD \\

Plastic molding
& Spot welding
& Sealing material
& Water jet cutting
& Assembling unspecified
& Cleanroom for semiconductors \\

Stamping, forging, bending
& Laser welding
& Other dispensing, spraying
& Grinding
& Disassembling
& Cleanroom for others \\

Handling operation at machine tools
& Welding other
& Dispensing unspecified
& Processing other
&
& Service applications \\

Machine tending
& Soldering
&
& Processing unspecified
&
& Medical applications \\

Measurement, inspection, testing
& Welding unspecified
&
&
&
& Others \\

Palletizing
&
&
&
&
& Unspecified \\

Packaging, picking, placing
&
&
&
&
& \\

Material handling
&
&
&
&
& \\

Handling operations unspecified
&
&
&
&
& \\

\end{longtable}
}

{\small
\setlength{\extrarowheight}{5pt}
\begin{longtable}{p{3.0cm} p{3.0cm} p{3.0cm} p{3.0cm}}

\caption
[IFR taxonomy of service professional robots]
{\textbf{IFR taxonomy of service professional robots.}
\label{tab:ifr_service_professional}} \\

\toprule
\textbf{Agriculture} 
& \textbf{Cleaning} 
& \textbf{Inspection} 
& \textbf{Construction/ Demolition} \\
\midrule
\endfirsthead

\caption[]{\textbf{IFR taxonomy of service professional robots.} (continued)} \\
\toprule
\textbf{Agriculture} 
& \textbf{Cleaning} 
& \textbf{Inspection} 
& \textbf{Construction/ Demolition} \\
\midrule
\endhead

\midrule
\multicolumn{4}{r}{\textit{Continued on next page}} \\
\midrule
\endfoot

\bottomrule
\endlastfoot


Cultivation
& Clean floor
& Inspect building
& Construction demolition \\

Milking
& Disinfect
& Inspect other
& \\

Farming other
& Cleaning other
&
\\

\midrule

\addlinespace[24pt]


\textbf{Transportation/ Logistics}
& \textbf{Search/Rescue/ Security}
& \textbf{Hospitality}
& \textbf{Other} \\
\midrule


Indoors without public
& Search, rescue services
& Prepare food
& Other \\

Indoors with public
& Security
& Mobile guidance, telepresence
& \\

Outdoors without public
&
&
\\

Outdoors with public; other transportation and logistics
&
&
\\

\end{longtable}
}

\setlength{\extrarowheight}{5pt}
\begin{table}[ht]
\centering

\caption{\textbf{IFR taxonomy of medical robots.}}

\label{tab:ifr_medical_prof}

\begin{tabular}{p{6cm}}
\toprule
\textbf{Medical robots} \\
\midrule
Diagnostics and medical analysis \\

Perform surgery \\

Rehabilitation and non-invasive therapy \\

Other medical robots \\
\bottomrule
\end{tabular}

\end{table}

\setlength{\extrarowheight}{5pt}
\begin{table}[h]
    \centering
    \caption{
        \textbf{IFR taxonomy of service consumer robots.}
    }
    \label{tab:ifr_service consumer}
    \begin{tabular*}{\textwidth}{@{\extracolsep\fill}p{2.8cm}p{2.8cm}p{2.8cm}p{2.8cm}}
    \toprule
    \textbf{Domestic tasks} 
    & \textbf{Social interaction, education} 
    & \textbf{Care at home} 
    & \textbf{Consumer other} \\
    
    \midrule
    
    Cleaning floors indoor
    &  Social interaction, companions
    & Mobility assistants, assist eating
    & Other consumer robots
    \\

    Window cleaning, gardening and other domestic tasks
    & Education
    & Manipulation aids and other care robots
     \\

    Cleaning outdoors
    &  
    & 
    & 
    \\
    \bottomrule
    \end{tabular*}
    
\end{table}

\clearpage

\subsection{Classification of IFR subclasses onto the ontology}
\label{classification}

\begin{figure}[h]
    \centering
    \includegraphics[width=0.9\linewidth]{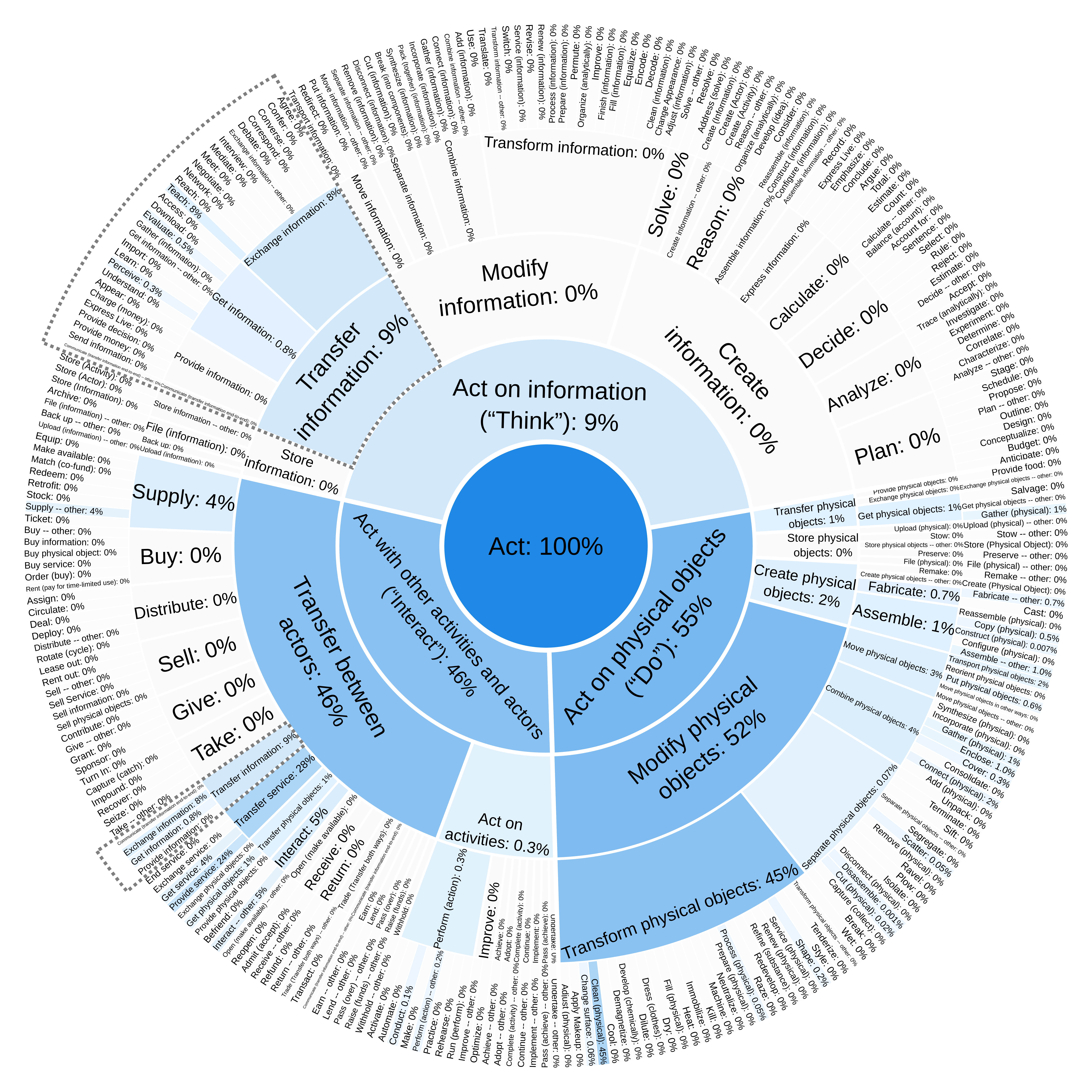}
    \caption{
        \textbf{Sunburst diagram showing the distribution of robotic systems deployed in 2024 excluding the node ``Clean floor''.}
    }
    \label{exclude_clean_floor_robot_inst_2024}
\end{figure}

\autoref{tab:ifr_to_ontology} shows the mapping of the IFR subclasses to the Ontology and~\autoref{exclude_clean_floor_robot_inst_2024} shows the distribution of robotic systems deployed in 2024 excluding the node ``Clean floor''.

{\small
\begin{longtable}{p{0.58\linewidth} p{0.42\linewidth}}
    \caption[Mapping of IFR subclasses to ontology work activities.]{
        \textbf{Mapping of IFR subclasses to ontology work activities.}
        \label{tab:ifr_to_ontology}
    } \\
    
    \toprule
    \textbf{IFR Subclass} & \textbf{Ontology Work Activity} \\
    \midrule
    \endfirsthead
    
    \caption[]{\textbf{Mapping of IFR Subclasses to Ontology Work Activities.} (continued)} \\
    \toprule
    \textbf{IFR Subclass} & \textbf{Ontology Work Activity} \\
    \midrule
    \endhead
    
    \midrule
    \multicolumn{2}{r}{\textit{Continued on next page}} \\
    \midrule
    \endfoot
    
    \bottomrule
    \endlastfoot

    \multicolumn{2}{l}{\textbf{Industrial robots}} \\
    \midrule
    
    Metal casting & Pour Metal \\
    Handling for plastic molding & Mold Material \\
    Handling for stamping/forging & Shape Material \\
    Handling operations at machine tools & Operate Tool \\
    Machine tending for other processes & Tend Machine \\
    Measurement, inspection, testing & Measure (Physical Object) \\
    Palletizing & Stack (Physical Object) \\
    Packaging, picking, placing & Pack (together) (Physical Object) \\
    Material handling & Handling Object \\
    Handling operations unspecified & Handling Object \\
    Arc welding & Weld Metal \\
    Spot welding & Weld Connection \\
    Laser welding & Weld Connection \\
    Other welding & Weld (Physical Object) \\
    Soldering & Solder (Physical Object) \\
    Welding unspecified & Weld (Physical Object) \\
    Painting and enamelling & Coat (Physical Object) \\
    Application of adhesive, sealing material & Apply Sealer \\
    Other dispensing/spraying & Spray (Physical Object) \\
    Dispensing unspecified & Spray (Physical Object) \\
    Laser cutting & Cut Material \\
    Water jet cutting & Cut Material \\
    Mechanical cutting/grinding/deburring & Smooth (Physical Object) \\
    Other processing & Process (Physical Object) \\
    Processing unspecified & Process (Physical Object) \\
    Assembling & Assemble (Physical Object) \\
    Assembling unspecified & Assemble (Physical Object) \\
    Disassembling & Disassemble (Physical Object) \\
    Cleanroom for FPD & Fabricate Panel \\
    Cleanroom for semiconductors & Fabricate Panel \\
    Cleanroom for others & Fabricate (Physical Object) \\
    Service applications (other) & Assist (Activity) \\
    Medical applications (industrial) & Treat (Actor) \\
    Other (industrial) & Act on Physical Objects (Do) \\
    Unspecified (industrial) & Act on Physical Objects (Do) \\ \\
    \midrule
    
    
    \multicolumn{2}{l}{\textbf{Professional service robots}} \\
    \midrule
    
    Cultivation (harvest) & Harvest Crop \\
    Milking & Collect Fluid \\
    Other (farming/agriculture) & Cultivate \\
    Professional clean floor & Clean Floor \\
    Disinfect & Disinfect (Physical Object) \\
    Professional clean other & Clean (Physical Object) \\
    Inspect building / construction & Inspect Building \\
    Inspect other & Inspect (Physical Object) \\
    Construction / demolition & Construct (Physical Object) \\
    Transportation and logistics indoors without public & Transport (Physical Object) \\
    Transportation and logistics indoors with public & Transport (Physical Object) \\
    Transportation and logistics outdoors without public & Transport (Physical Object) \\
    Outdoor environments with public traffic & Transport (Physical Object) \\
    Rescue services / firefighting & Rescue (Actor) \\
    Security services & Patrol Area \\
    Prepare food/drinks & Prepare Product \\
    Mobile guidance / information / telepresence & Guide (Actor) \\
    Other professional service robots & Perform Service \\
    
    \midrule
    
    
    \multicolumn{2}{l}{\textbf{Medical robots}} \\
    \midrule
    
    Diagnostics and medical laboratory analysis & Conduct Diagnostic \\
    Perform surgery & Perform Surgery \\
    Rehabilitation and non-invasive therapy & Provide Rehabilitation \\
    Other medical robots & Assist Patient \\
    
    \midrule
    
    
    \multicolumn{2}{l}{\textbf{Consumer service robots}} \\
    \midrule
    
    Cleaning floors indoor (domestic) & Clean Floor \\
    Window cleaning, other domestic tasks & Clean Window \\
    Gardening and other domestic tasks & Care for Lawn \\
    Cleaning outdoors (domestic) & Clean Yard \\
    Social interaction / companions & Provide Companionship \\
    Education & Teach (Actor) \\
    Mobility assistants / assist eating & Assist Individual \\
    Manipulation aids and other care robots & Provide Care \\
    Other consumer robots & Interact Client \\
        
\end{longtable}
}
\clearpage

\clearpage

\newpage
\subsection{Robotics Sunburst diagrams (2015--2024)}
\label{ap:robot:sunburst}

\begin{figure}[ht]
    \centering
    \includegraphics[width=0.9\linewidth]{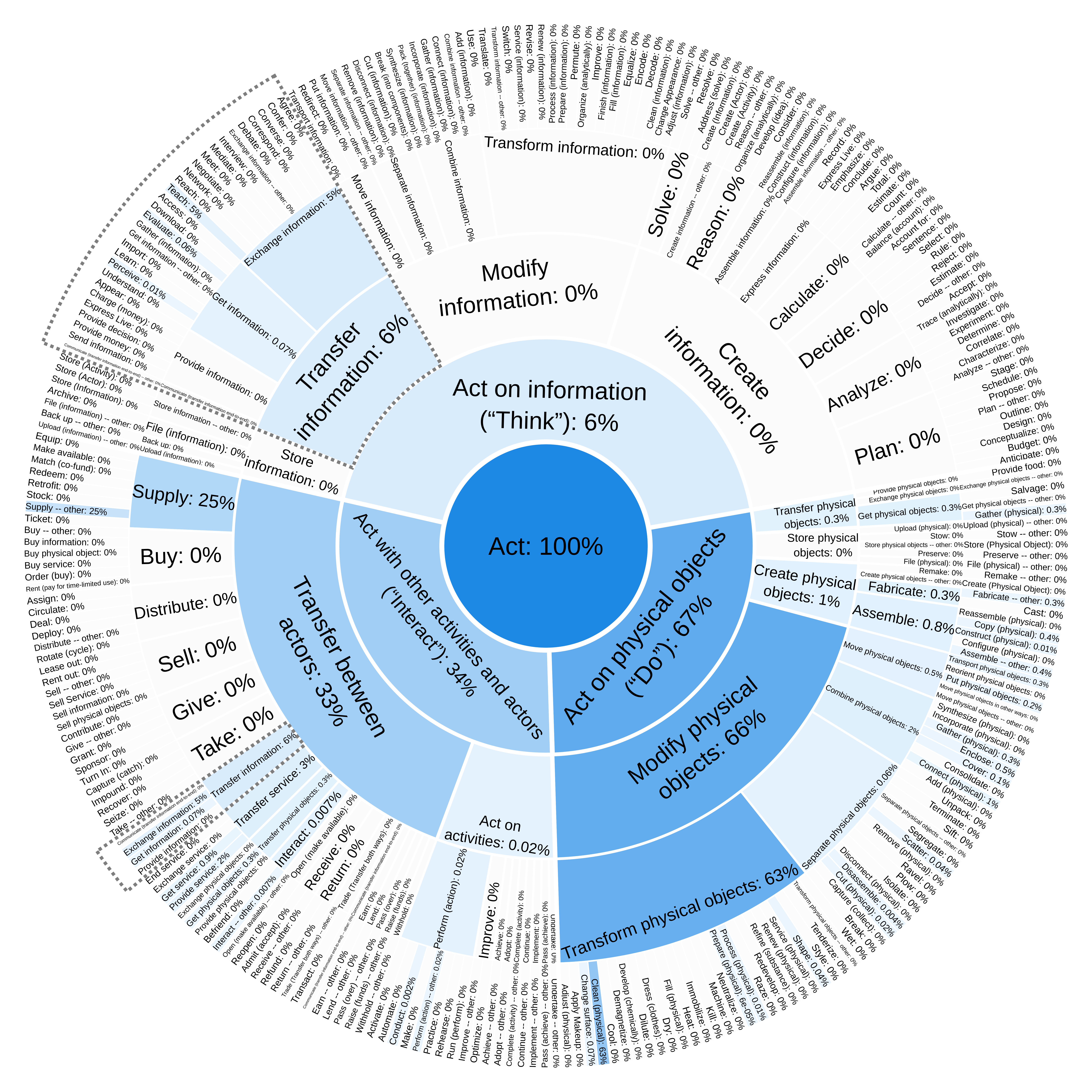}
    \caption{Distribution of robot units in 2015.}
    \label{fig:robotics_yearly_2015}
\end{figure}

\newpage
\begin{figure}[ht]
    \centering
    \includegraphics[width=0.9\linewidth]{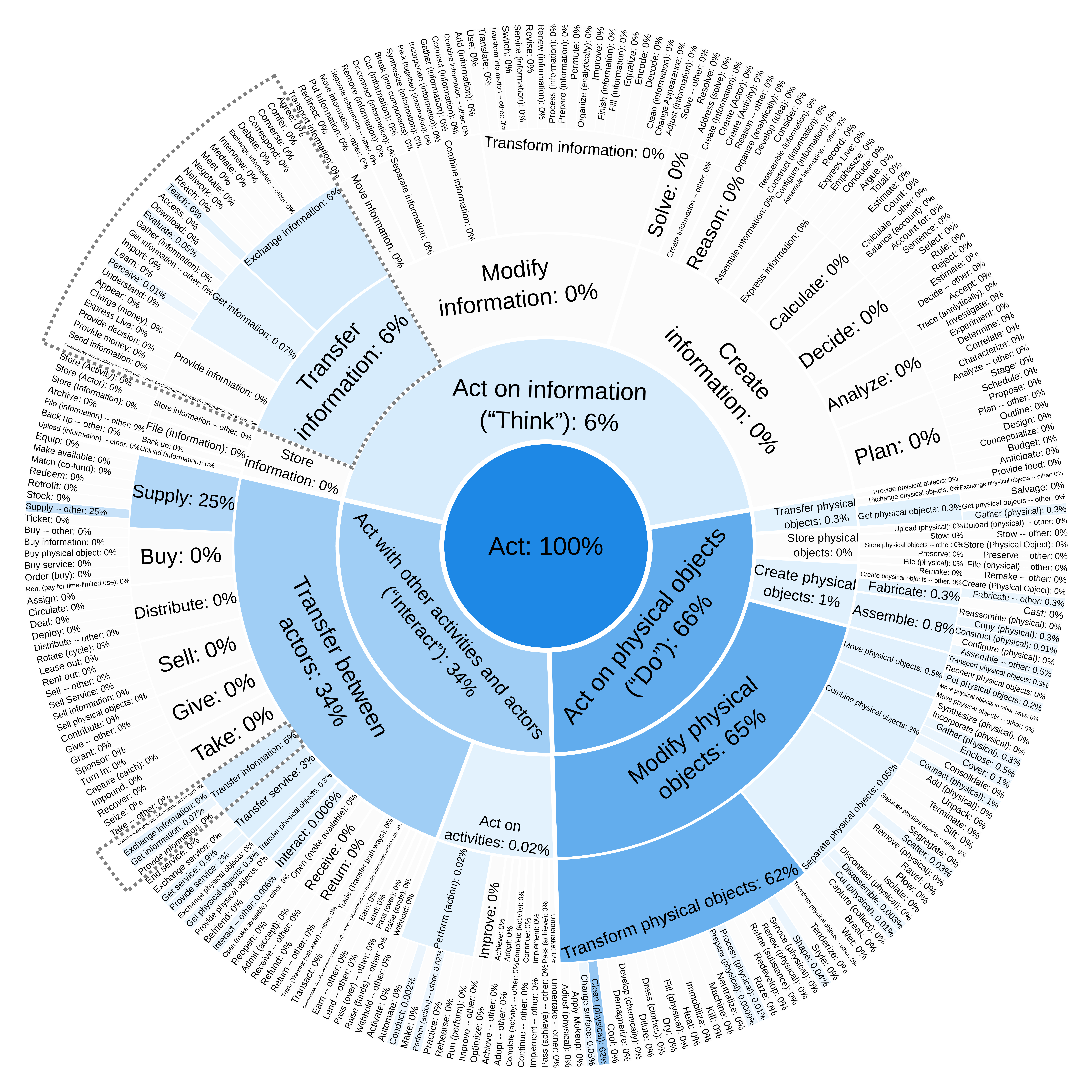}
    \caption{Distribution of robot units in 2016.}
    \label{fig:robotics_yearly_2016}
\end{figure}

\newpage
\begin{figure}[ht]
    \centering
    \includegraphics[width=0.9\linewidth]{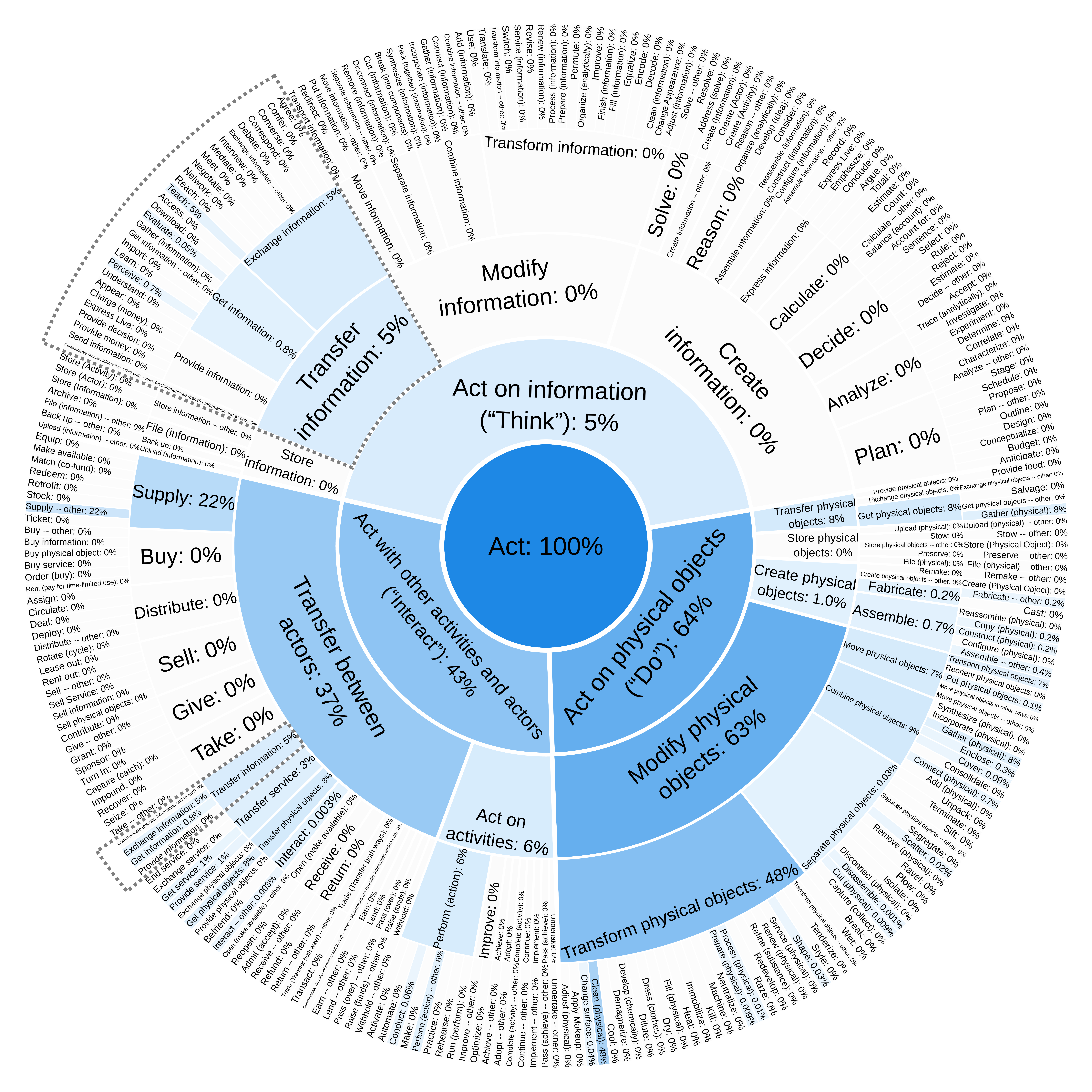}
    \caption{Distribution of robot units in 2017.}
    \label{fig:robotics_yearly_2017}
\end{figure}

\newpage
\begin{figure}[ht]
    \centering
    \includegraphics[width=0.9\linewidth]{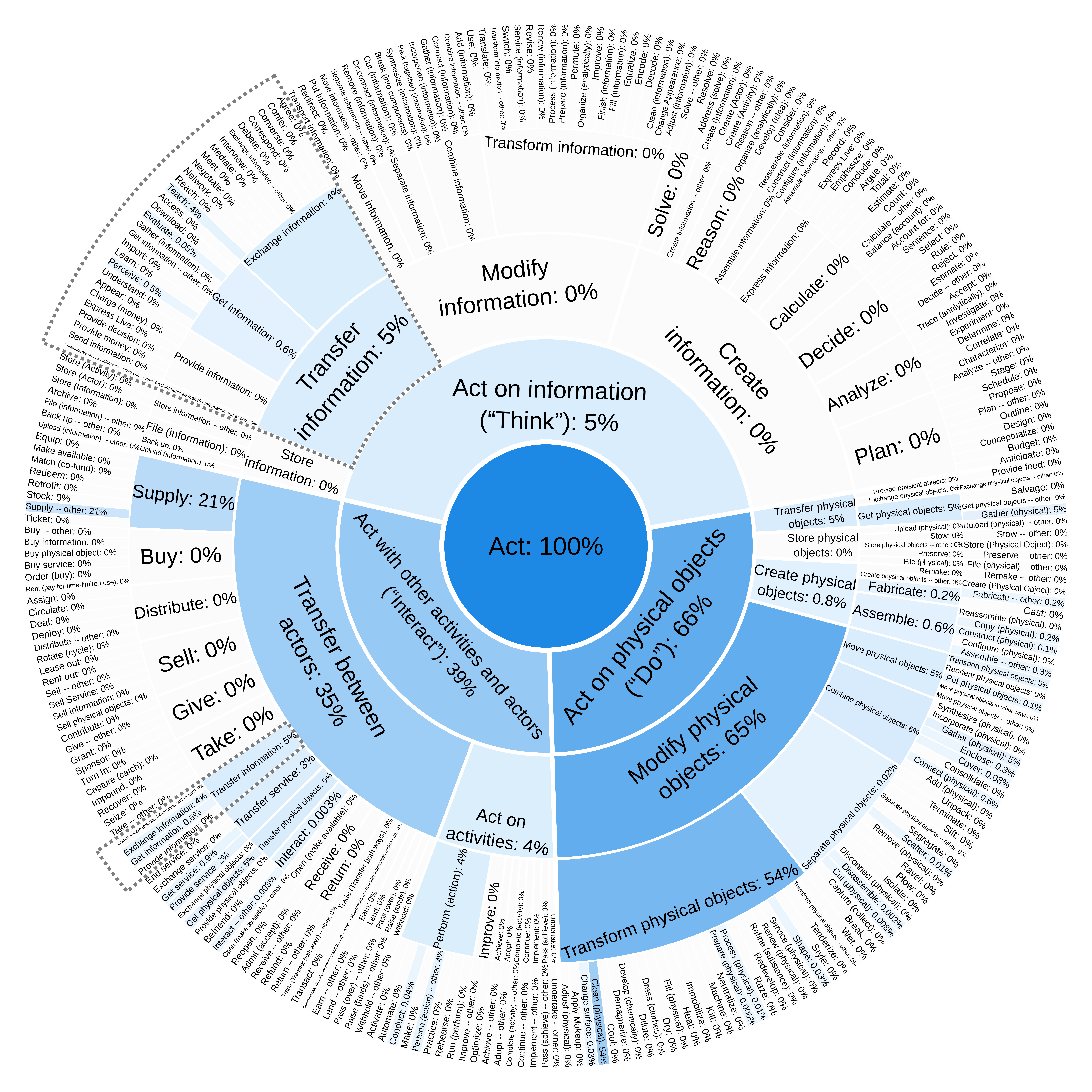}
    \caption{Distribution of robot units in 2018.}
    \label{fig:robotics_yearly_2018}
\end{figure}

\newpage
\begin{figure}[ht]
    \centering
    \includegraphics[width=0.9\linewidth]{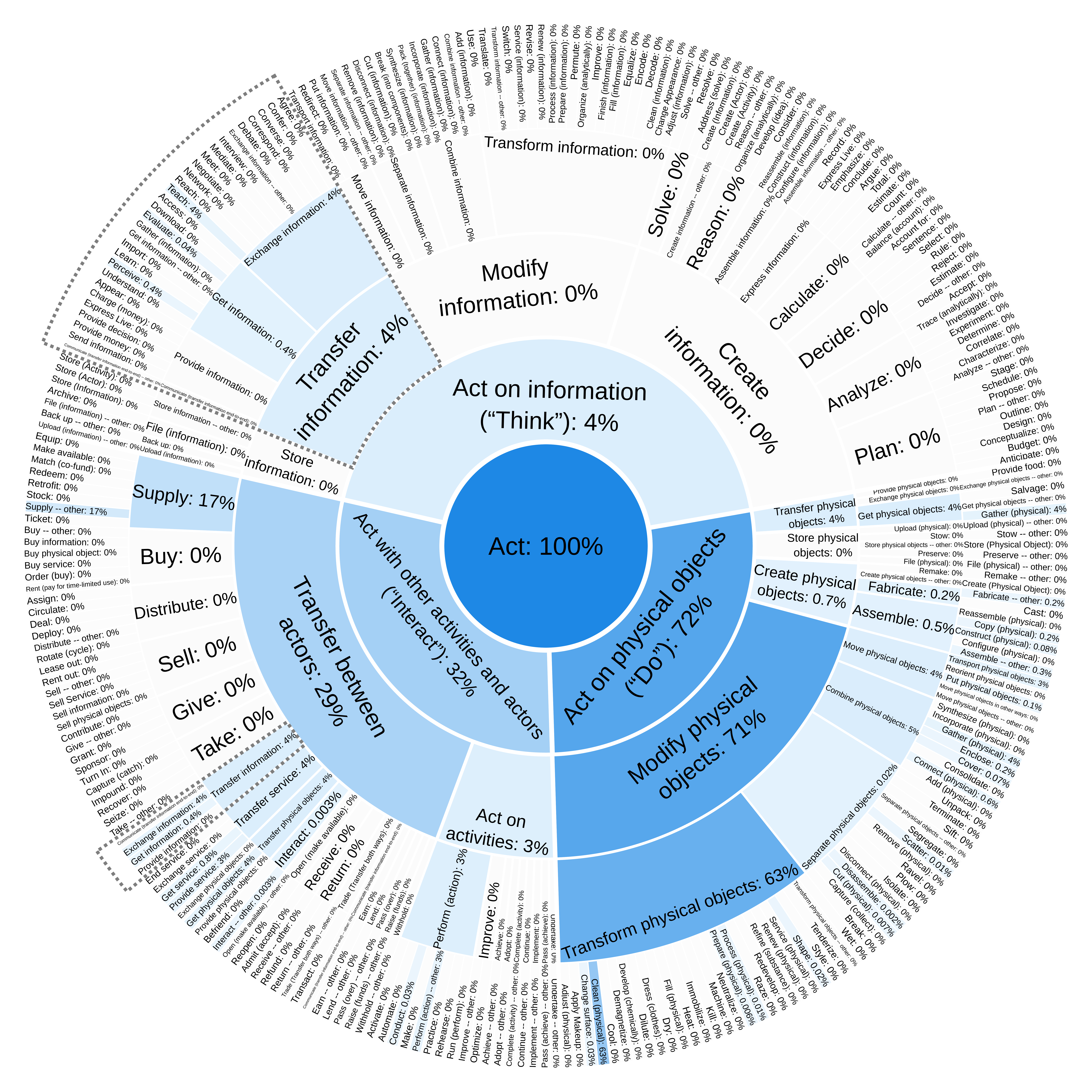}
    \caption{Distribution of robot units in 2019.}
    \label{fig:robotics_yearly_2019}
\end{figure}

\newpage
\begin{figure}[ht]
    \centering
    \includegraphics[width=0.9\linewidth]{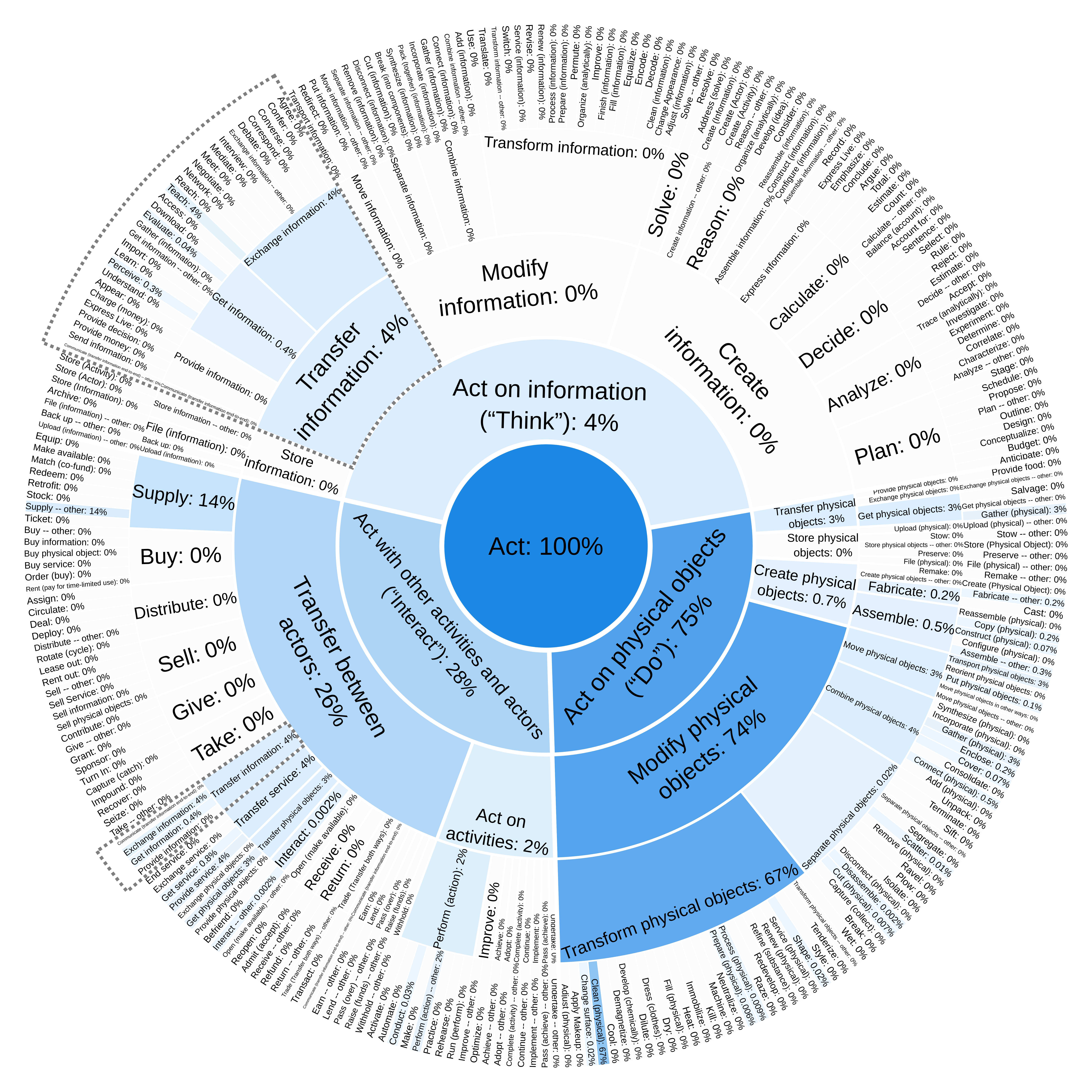}
    \caption{Distribution of robot units in 2020.}
    \label{fig:robotics_yearly_2020}
\end{figure}

\newpage
\begin{figure}[ht]
    \centering
    \includegraphics[width=0.9\linewidth]{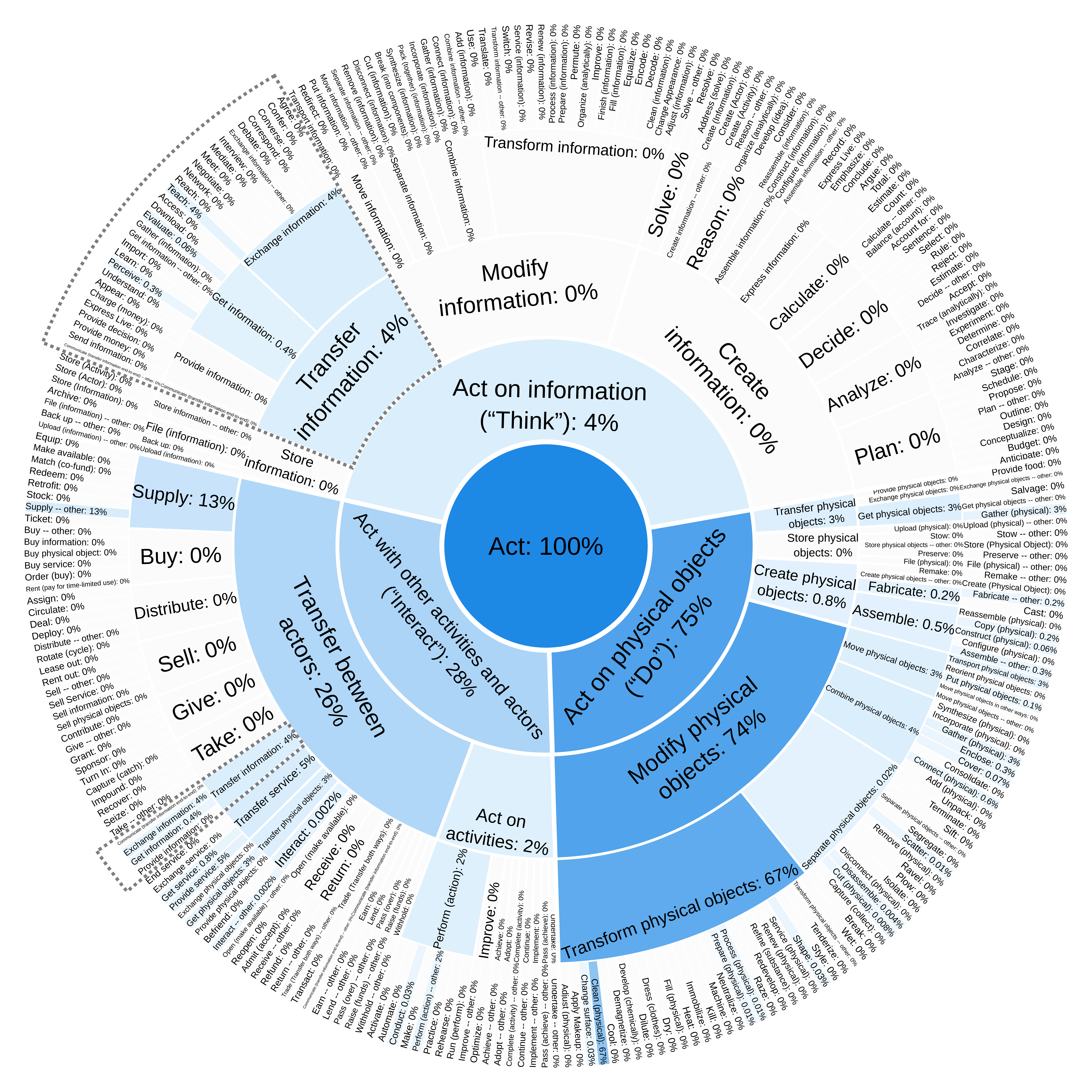}
    \caption{Distribution of robot units in 2021.}
    \label{fig:robotics_yearly_2021}
\end{figure}

\newpage
\begin{figure}[ht]
    \centering
    \includegraphics[width=0.9\linewidth]{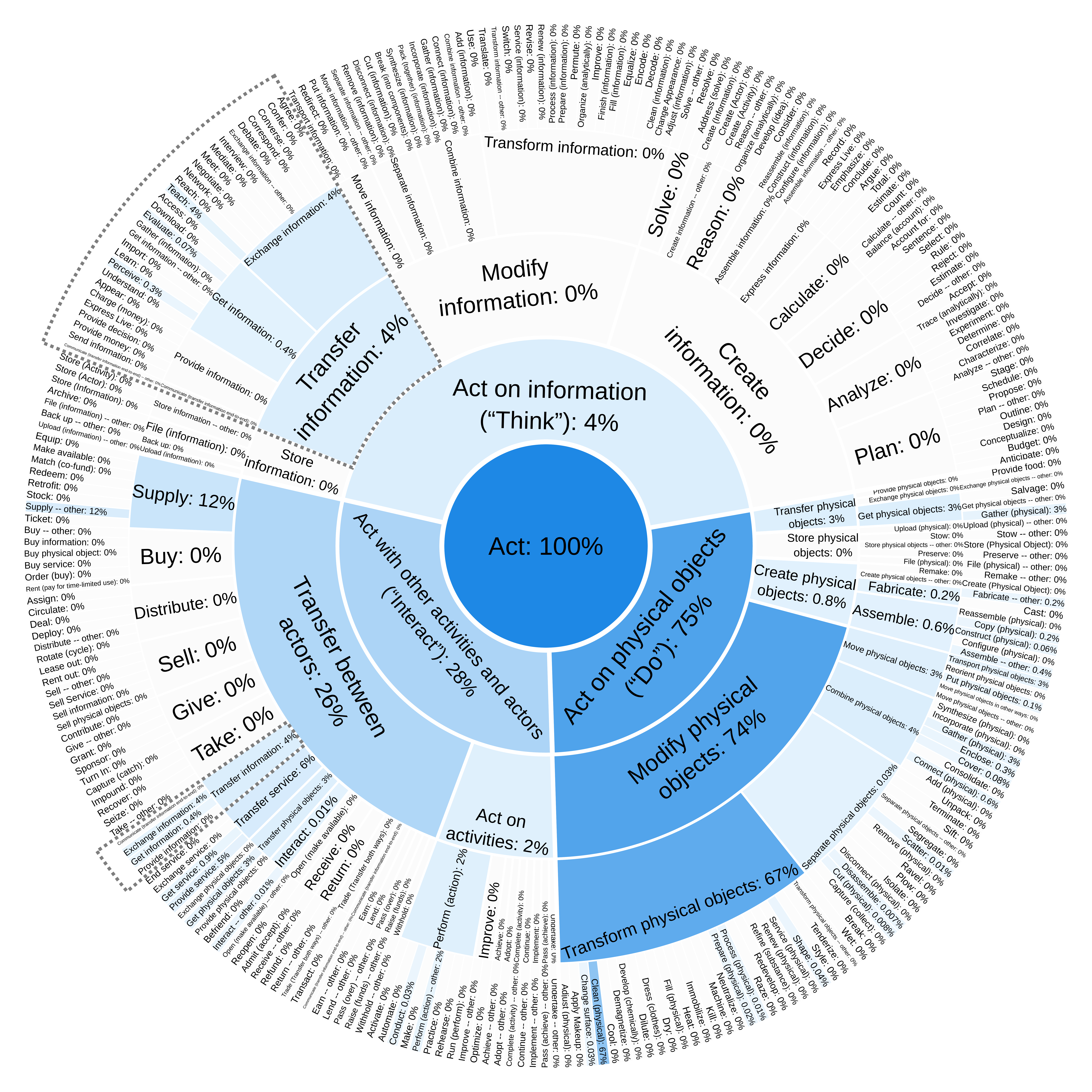}
    \caption{Distribution of robot units in 2022.}
    \label{fig:robotics_yearly_2022}
\end{figure}

\newpage
\begin{figure}[ht]
    \centering
    \includegraphics[width=0.9\linewidth]{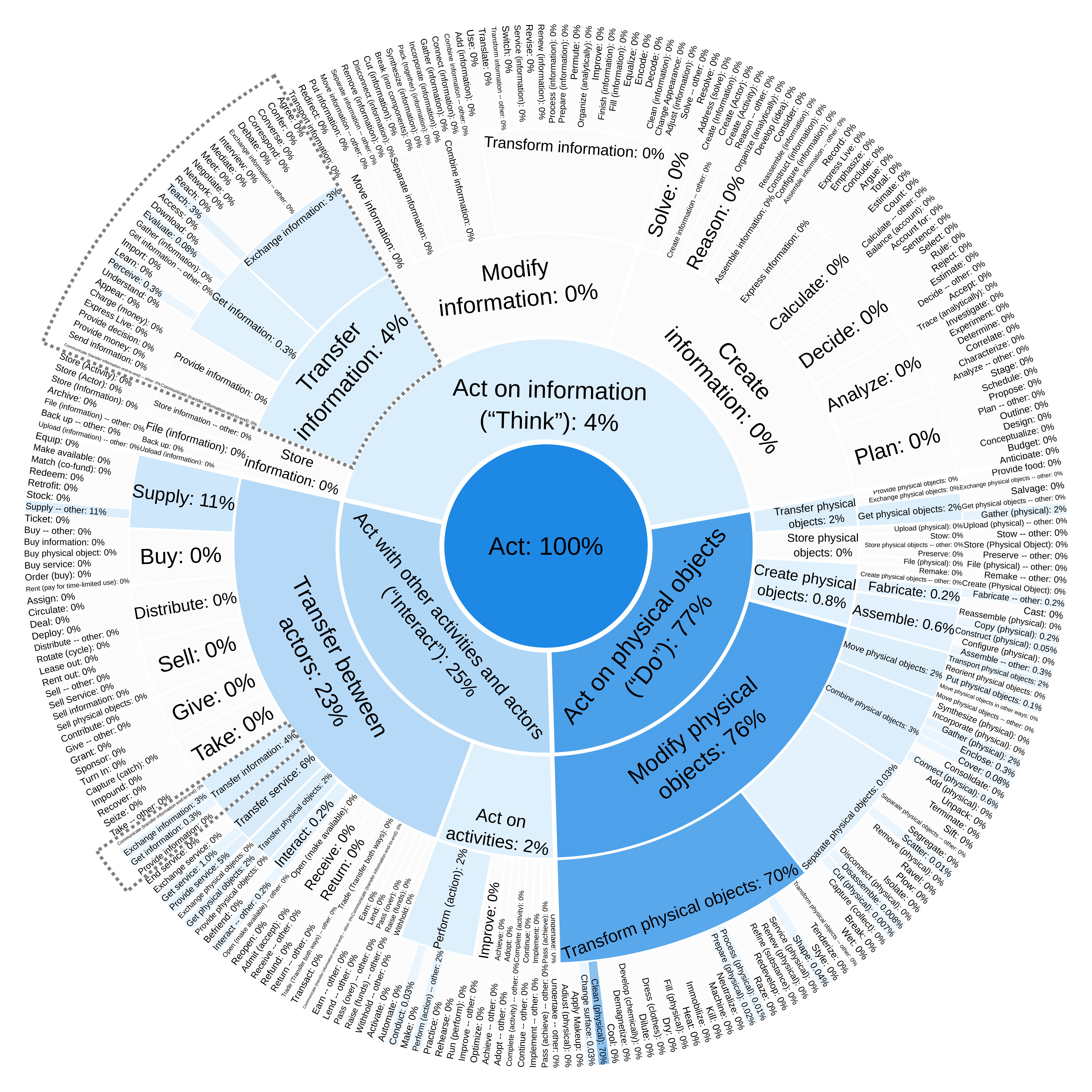}
    \caption{Distribution of robot units in 2023.}
    \label{fig:robotics_yearly_2023}
\end{figure}

\newpage
\begin{figure}[ht]
    \centering
    \includegraphics[width=0.9\linewidth]{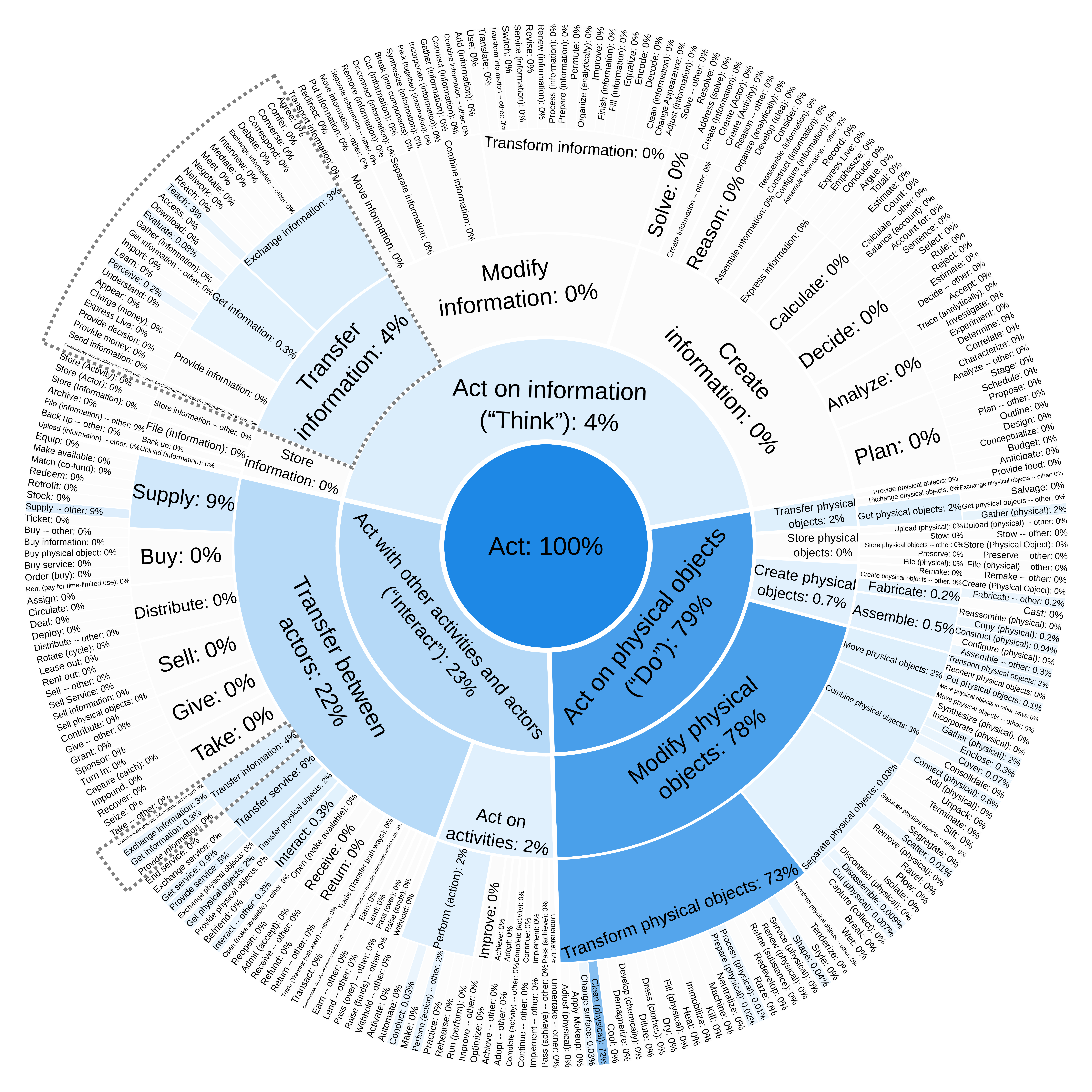}
    \caption{Distribution of robot units in 2024.}
    \label{fig:robotics_yearly_2024}
\end{figure}

\clearpage
\section{Combining usage measures for AI software applications and robotic systems}
\label{combination}
 \subsection{Assumptions}
\label{market_assumptions}

To combine the measures of AI usage from AI software applications and robotic systems, we made the assumptions and completed the calculations listed below. We believe that our approach is a plausible way to estimate the desired quantities. But it is important to realize that all the results are only approximate, and we hope that we and others can continue to develop better ways of estimating these numbers.

\begin{itemize}
    \item \textit{Assumption 1:} The \textit{size of the overall global AI market} in 2024 was USD 186.4 billion.
    \begin{itemize}
        \item Justification: \cite[p.~20]{statista2025ai}
    \end{itemize}
    \item \textit{Assumption 2:} The \textit{size of the global market for robotic systems} in 2024 was USD 46.1 billion. 
    \begin{itemize}
        \item Justification: \cite[p.~7]{statista2024robotics}
        \item Justification: Although there are other estimates available for the global robotics market size, this source is closely aligned with the IFR report that provided our primary data on robotics usage. There are close similarities between the two sources in both:
        \begin{itemize}
            \item market size estimates: USD 46.11 billion vs. USD 50.0 billion \cite[p.~65]{IFR2025Industrial}
            \item definitions of robots and categories of robots.
        \end{itemize} 
    \end{itemize}
    \item \textit{Assumption 3:} The \textit{size of global market for AI software applications} in 2024 was USD 140.29 billion. 
    \begin{itemize}
        \item Justification: The global AI market can be divided into two parts: (a) AI software applications and (b) robotic systems. Therefore, subtracting the robotics market size [Assumption 2] from the overall AI market size [Assumption 1] should give a reasonable estimate for the size of the global market for AI software. 
    \end{itemize}
    \item \textit{Assumption 4:} The \textit{mapping of IFR subclasses to robotics industry segments} described in Step 2 of Section \ref{estimate_robot_market_value} is a plausible basis for estimating market sizes of the different IFR subclasses.
    \begin{itemize}
        \item Justification: This mapping was done by three knowledgeable experts in the field of robotics who consulted a variety of reference materials in order to make plausible inferences.
    \end{itemize}
    \item \textit{Assumption 5:} The \textit{method for estimating relative prices for IFR subclasses} described in Step 3 of Section \ref{estimate_robot_market_value} is a plausible basis for estimating market sizes of the different IFR subclasses. 
    \begin{itemize}
        \item Justification: 
        The estimation of relative prices was done by a knowledgeable expert in the field of robotics who consulted publicly available price ranges for representative robotic systems within each IFR subclass, drawing on industry reports and manufacturer information. 
         \end{itemize}
         
        \item \textit{Assumption 6:} The \textit{mathematical procedure for calculating market values of the IFR subclasses based on their relative prices}, as described in Steps 4 and 5 of Section \ref{estimate_robot_market_value}, is a plausible method for estimating these numbers. 
  
    \begin{itemize}
        \item Justification: The mathematical equations shown are self-evident logical consequences of the meaning of the variables. But the method adjusts relative prices separately for each industry segment, and that means the same robotic subclass might receive different estimated prices in different industry segments. Even though these estimated prices are unlikely to be exactly correct, the total market value attributed to subclasses that occur in multiple industry segments is approximately 10\%. So even if \textit{all} these attributions of market value were completely incorrect, the changes in results would be no more than a total of 10\%  spread across all the activities in the ontology. Therefore, we assume that the resulting market values will still be generally useful.
\end{itemize}
  \end{itemize}


 

\subsection{Estimating the market value of AI software applications}
\label{estimate_AI_market_value}

To calculate the market value of the AI applications in the TAAFT dataset, we first estimated revenue for each application. Because direct revenue figures were not publicly available for most applications in the dataset, we approximated revenue through observable engagement and pricing signals.
Using this, we estimated the relative market share of each AI application within the TAAFT dataset using this formula:

\par\vspace{\abovedisplayskip}
\noindent\(\displaystyle
    \text{Market share}_{\text{application}} =
    \frac{
        \text{Number of saves} \times \text{Price} \times \text{Billing frequency}
    }{
        \sum_{}
        \left(
        \text{Number of saves} \times \text{Price} \times \text{Billing frequency}
        \right)
    }
\)
\par\vspace{\belowdisplayskip}



Then, using the ontology-based classification results described in Section~\ref{sec:taaft:current}, we attributed each application’s estimated market share to the ontology activity in which that application was classified. Aggregating across all applications produced an estimated market share for each activity node in the ontology.

Next, to estimate the size of the global market for AI software applications, we started with the estimated size of the overall AI market in 2024 (USD 186.4 billion) \cite[p.~20]{statista2025ai} \textit{[Assumption 1]}, and then subtracted the size of the robotics market (USD 46.11 billion) \cite[p.~7]{statista2024robotics} \textit{[Assumption 2]}. This yielded a estimated market value of USD 140.29 billion for the global AI software applications market [Assumption 3]. In other words, AI software applications account for 75\% of the overall AI market value, and robotic systems account for 25\%.


Finally, to compute the market value of AI systems in each ontology activity, we used the following formula: 
\par\vspace{\abovedisplayskip}
\noindent\(\displaystyle
    \text{Market value}_{\text{activity node}} = 140.29~\text{B USD} \times \text{Market share}_{\text{activity node}}
\)
\par\vspace{\belowdisplayskip}

This procedure provides an approximate, ontology-informed view of how the global market value of AI software applications is distributed across different categories of activities.

\subsection{Estimating the market value of robotic systems}
\label{estimate_robot_market_value}

To estimate the market value of each robot subclass, we performed the following steps:

\paragraph{Step 1: Estimate market sizes for different robotics industry segments}

We began with the global robotics market estimate, which Statista reports as USD 46.1 billion~\cite[p.~7]{statista2024robotics} in 2024 \textit{[Assumption 2]}. Statista also provides data on market shares for 12 robotics industry segments (e.g. medical, logistics, domestic services, and multiple industrial categories)~\cite[p.~10]{statista2024robotics}. We calculated market values in each industry segment by multiplying the shares by our market size estimate (USD 46.1 billion).

\autoref{tab:market_share_revenue_2024} presents the market share per segment as reported in Statista together with the corresponding market values.

\begin{table}[h]
    \centering
    \caption{
        \textbf{Market share and corresponding market value by robotic industry (2024).}
    }
    \label{tab:market_share_revenue_2024}
    \begin{tabular}{ p{4.0cm} p{3.0cm} p{3.0cm}}
    \toprule
      \textbf{Industry segment}& \textbf{2024 market share (\%)} & \textbf{2024 market value (USD B)} \\
    \midrule
    
    \multicolumn{3}{l}{\textbf{Industrial}} \\
    \midrule
     Automotive & 7\% & 3.2 \\
    Electrical/Electronic & 6\% & 2.8 \\
    Logistics & 5\% & 2.3\\

     Metal & 2\% & 0.9 \\
     Chemical & 2\% & 0.9 \\
     Food & 1\% & 0.5 \\
     Other industrial & 2\% & 0.9 \\
    \midrule
    
    \multicolumn{3}{l}{\textbf{Professional services}} \\
    \midrule
     Medical & 29\% & 13.2 \\
     Agriculture & 3\% & 1.4 \\
    Other professional services* & 9\% & 4.1 \\
    \midrule
    
    \multicolumn{3}{l}{\textbf{Consumer}} \\
    \midrule
     Domestic & 24\% & 11.2 \\
    Entertainment & 10\% & 4.6 \\
    \midrule
    
    \textbf{Total} &  \textbf{100\%} & \textbf{46.0} \\
    \bottomrule
    \end{tabular}
    
    \vspace{0.5em}
    \footnotesize{\textit{*Other professional services robots include cleaning, inspection, construction, search \& rescue, and hospitality.}}
\end{table}

\paragraph{Step 2: Map IFR subclasses to robotics industry segments}

In \autoref{tab:ifr_to_ontology}, we mapped the 66 IFR subclasses to activities in the ontology. Now, to connect market sizes to ontology activities, we also needed to map the 66 IFR subclasses to the 11 industry segments. In some cases, IFR subclasses correspond directly to industry segments (e.g., IFR's four subclasses of medical robots all correspond to the medical industry segment, and IFR's subclasses in transportation and logistics correspond to the logistics industry segment). 

In other cases, however, there are no obvious correspondences between IFR subclasses and industry segments. In these cases, three of the authors of this paper, with expertise in robotics, reviewed multiple external sources in order to make plausible inferences about which IFR subclasses would be included in which industry segments [\textit{Assumption 4}]. For example, within the ``automotive'' industry segment, we assigned the following IFR subclasses: arc welding, assembling, dispensing/spraying, disassembling, machine tending, material handling, painting and enameling, spot welding. Because published sources differ in how they categorize robot applications within specific industries, we made subjective judgments about how best to reconcile these differences. And these classifications should, therefore, be regarded as approximations, not as definitive classification.
In addition to the IFR report on industrial robots \cite{IFR2025Industrial}, sources were consulted to help make these mappings for various industry segments included:  
 Electric/Electronic~\cite{bogue_role_2023}, Automotive~\cite{noauthor_6_2020}, Metal~\cite{noauthor_machinery_nodate},
Food~\cite{robots_food_industry_2026}, and Chemical~\cite{noauthor_chemistry_nodate} segments.


\paragraph{Step 3: Estimate the *relative* prices for each IFR subclass in each industry}

Since there is no obvious source of pricing information for different IFR subclasses, we needed to use some form of expert estimates for these quantities. But, due to the large variability in robotic systems, we believed it would be difficult or impossible, even for experts, to directly estimate prices for each subclass in a way that would also be mathematically consistent with both (a) the actual units sold in each subclass and (b) the overall revenues in each industry segment. Therefore, we chose to start by estimating \textit{relative} prices for each IFR subclass in an industry segment, and then, in Step 4, to pick an adjustment factor to generate estimates of actual prices that were simultaneously consistent with (a) the estimated relative prices, (b) the units sold per subclass, and (c) the revenues in that industry segment.

To generate estimates of the relative prices, the same author of this paper who served as an expert in Step 2, also created estimates of \textit{relative} average unit prices in each industry segment \textit{[Assumption 5]}. This was done by first compiling estimates of high and low prices for each subclass, using publicly available sources, including manufacturer disclosures, system integrator listings, and industry reports. When sources reported differing price ranges, we adopted the widest reported range to avoid understating potential price dispersion. For subclasses appearing in multiple industry segments, we applied identical high and low price bounds across segments. Also, in these cases, we divided the IFR-reported number of installations evenly between the segments.

Then, for each subclass, we determined the midpoint between the estimated low and high unit prices and treated this as an estimate of the average unit price. Finally, for each industry segment, we rescaled the estimated prices to all be \textit{relative} to the lowest price. In other words, the lowest price would now be 1, and the other prices would be multiples of that lowest price. For example, if the lowest average price was $x$, a subclass whose average price was twice as high ($2x$) received a relative price of 2. This yielded a set of relative average unit prices for each subclass.

\paragraph{Step 4: Calculate an adjustment factor to make the estimated prices for each IFR subclass mathematically consistent with the relative prices and industry segment revenues}

To calculate the adjustment factor, we first calculate the total segment revenue that would be received if the \textit{relative} prices were used. Then we divide the \textit{actual} segment revenue by that amount to get the adjustment factor \textit{[Assumption 6]}. Mathematically, this can be expressed as follows:

For a given industry segment, let

\begin{flalign*}
R &= \text{revenue in the segment} \\
R_i &= \text{revenue in subclass } i \text{ of the segment} &\\
n &= \text{number of subclasses in the segment} &\\
U_i &= \text{units sold in subclass } i \text{ of the segment} &\\
P_i &= \text{average unit price in subclass } i \text{ of the segment} &\\
P'_i &= \textit{relative } \text{average unit price in subclass } i \text{ of the segment} &\\
x &= \text{price adjustment factor for the segment,} &
\end{flalign*}

\noindent where the relative prices are normalized such that the lowest relative price in the segment satisfies \(P'_1 = 1\), and all other
\(P'_i\) values are proportionally greater.
\\

\noindent Then:
\begin{flalign*}
P_i &= x * P'_i \\
R_i &= P_i * U_i &\\
    &= x * P'_i * U_i &
\end{flalign*}

\noindent Total segment revenue is given by:
\begin{flalign*}
R &= \sum_{i=1}^{n} R_i &\\
  &= \sum_{i=1}^{n} \left(x * P'_i * U_i\right) &\\
  &= x * \sum_{i=1}^{n} \left(P'_i * U_i\right) &
\end{flalign*}

\noindent Solving for the price adjustment factor: \\
\par\noindent\vspace{1.5\baselineskip}
\(\displaystyle
x = \frac{R}{\sum_{i=1}^{n} \left(P'_i * U_i\right)}
\)

\paragraph{Step 5: Calculate the estimated market value per subclass in each industry}
Once the adjustment factor is known, the estimated market value for each subclass can be calculated by multiplying the relative prices by the adjustment factor to get actual prices and then multiplying the actual prices by units to get revenue (which is equivalent to market value) in each subclass. 

\par\vspace{\baselineskip}
\noindent\begin{minipage}{\linewidth}

\noindent Using the mathematical notation from above, this is equivalent to calculating:
\par\vspace{1.0\baselineskip}
\noindent
\(\displaystyle
R_i = x * P'_i * U_i
\)
\end{minipage}
\par\vspace{\baselineskip}

\autoref{market_value_example} shows an example of this estimation process for the medical robots segment. This segment accounted for 29\% of global robotics revenue in 2024, which corresponded to approximately USD 13.2 billion. The medical segment has four subclasses: (i) diagnostics and laboratory analysis, (ii) surgical, (iii) rehabilitation and non-invasive therapy, and (iv) other. 
After  researching price ranges and completing the calculations, our analysis showed that surgical robots accounted for nearly 12 billion USD, more than 90\% of the 13.2 billion USD medical robots segment, while the other three medical subclasses each account for less than 1 billion USD.

\setlength{\extrarowheight}{5pt}
\begin{table}
\centering
\caption{
    \textbf{Estimated Market Value of Medical Robotics Subclasses (2024)}. \\The Medical Robotics industry represents 29\% of the global robotics market value (13.2bn USD). Average prices for each subclass are initially estimated as the midpoint of the reported price ranges and then converted to relative prices. The relative prices are multiplied by the price adjustment factor ($x = 110.1$) to obtain adjusted price estimates. Then, the adjusted price estimates are multiplied by the number of units to calculate the market value of each subclass. \\ 
}

\begin{tabular}{@{}p{1.7cm}rp{1.6cm}p{1.4cm}p{1.2cm}p{1.1cm}p{1.4cm}@{}}
\toprule
\textbf{Subclass} & \textbf{Units} & \textbf{Price range (USD)} & \textbf{Initial estimate of avg price (USD)} & \textbf{Relative avg price (ratio)} & \textbf{Adjusted avg price (USD)} & \textbf{Market value (USD)}\\
\midrule

Diagnostics/ Lab analysis & 3,293 & 100k--150k & 125k & 1.3 & 145k & 0.5 bn \\
Surgical & 6,612 & 600k--2,500k & 1550k & 16.4 & 1,800k & 11.9 bn \\
Rehab/ non-invasive therapy & 5,759 & 69k--120k & 94.5k & 1.0 & 110k & 0.6 bn \\
Other & 995 & 100k--150k & 125k & 1.3 & 145k & 0.2 bn \\

\cmidrule(l){1-7}
\textbf{Total (USD)} & & & & & & \textbf{13.2bn} \\

\bottomrule
\end{tabular}

\label{market_value_example}
\end{table}


\clearpage
\section{Results from market value analysis}
\label{Results_market_value_analysis}
\subsection{Results on market value of AI applications}
\label{Results_AI_market_value}

\begin{figure}[h]
    \centering
    \includegraphics[width=0.9\textwidth]{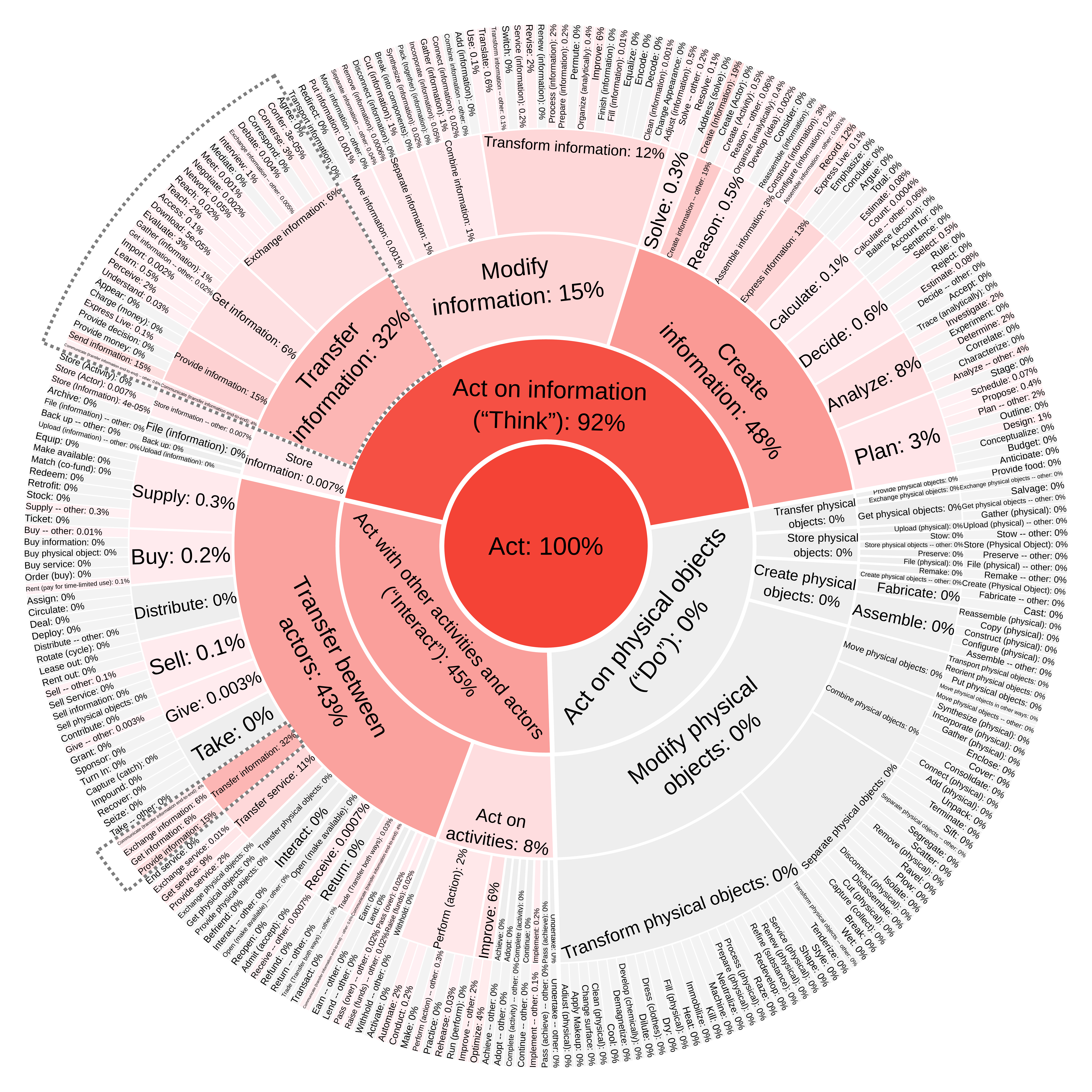}
    \caption{
        \textbf{The global market value distribution of AI software applications in 2024.}
    }
    \label{taaft_revenue}
\end{figure}

\autoref{taaft_revenue} presents a sunburst visualization of the global AI application market value projected onto our activity ontology. Among the three major categories of activities, ``Act on information (`Think')'' accounts for the largest share of revenue, covering approximately 92\% of the total. This is roughly twice the share of ``Act on activities or actors (`Interact')'', which accounts for about 45\%. Owing to the ontology’s multiple-inheritance structure, these percentages are not mutually exclusive. In particular, the activity ``Transfer information'', which is shared by both categories, represents approximately 32\% of the total AI application revenue.

Within ``Act on information (`Think')'', the most dominant activity is ``Create information'', which alone accounts for 48\% of total revenue. This concentration indicates that nearly half of the AI application market is driven by systems designed to generate new informational content, underscoring both the commercial success and relative maturity of generative AI technologies in current markets. Further decomposing ``Create information'' reveals that ``Create information -- other'' (19\%), ``Express information'' (13\%), and ``Transfer information'' (12\%) are the three largest contributing activities.

An examination of representative applications associated with these activities suggests that present-day AI revenue is primarily driven by products such as AI image and video generation and editing tools, conversational AI systems, writing assistants, and customer engagement platforms. These applications emphasize content creation, expression, and communication, reflecting strong demand for AI systems that operate directly on informational artifacts.

In contrast, activities that also inherit from ``Create information'', including ``Decide'', ``Plan'', and ``Analyze'', account for comparatively little market value. This imbalance suggests that, despite significant research progress in reasoning, planning, and analytical capabilities, these functions have not yet translated into widespread, monetized AI applications at scale. Possible explanations include higher integration costs, greater domain specificity, or the need for tighter coupling with organizational workflows and decision-making authority.

Overall, projecting AI application market value onto our ontology reveals a highly uneven distribution across activity types. A small number of content-centric activities account for the vast majority of market value, while many other activities account for minimal market value despite their potential importance. This imbalance highlights substantial untapped opportunities for AI deployment, particularly in activities that involve deeper reasoning, decision support, and complex task orchestration, areas that may represent the next frontier for economically impactful AI systems.

\subsection{Results on market value for robotic systems}

\label{Results_robot_market_value}

\begin{figure}[h]
    \centering
    \includegraphics[width=0.9\textwidth]{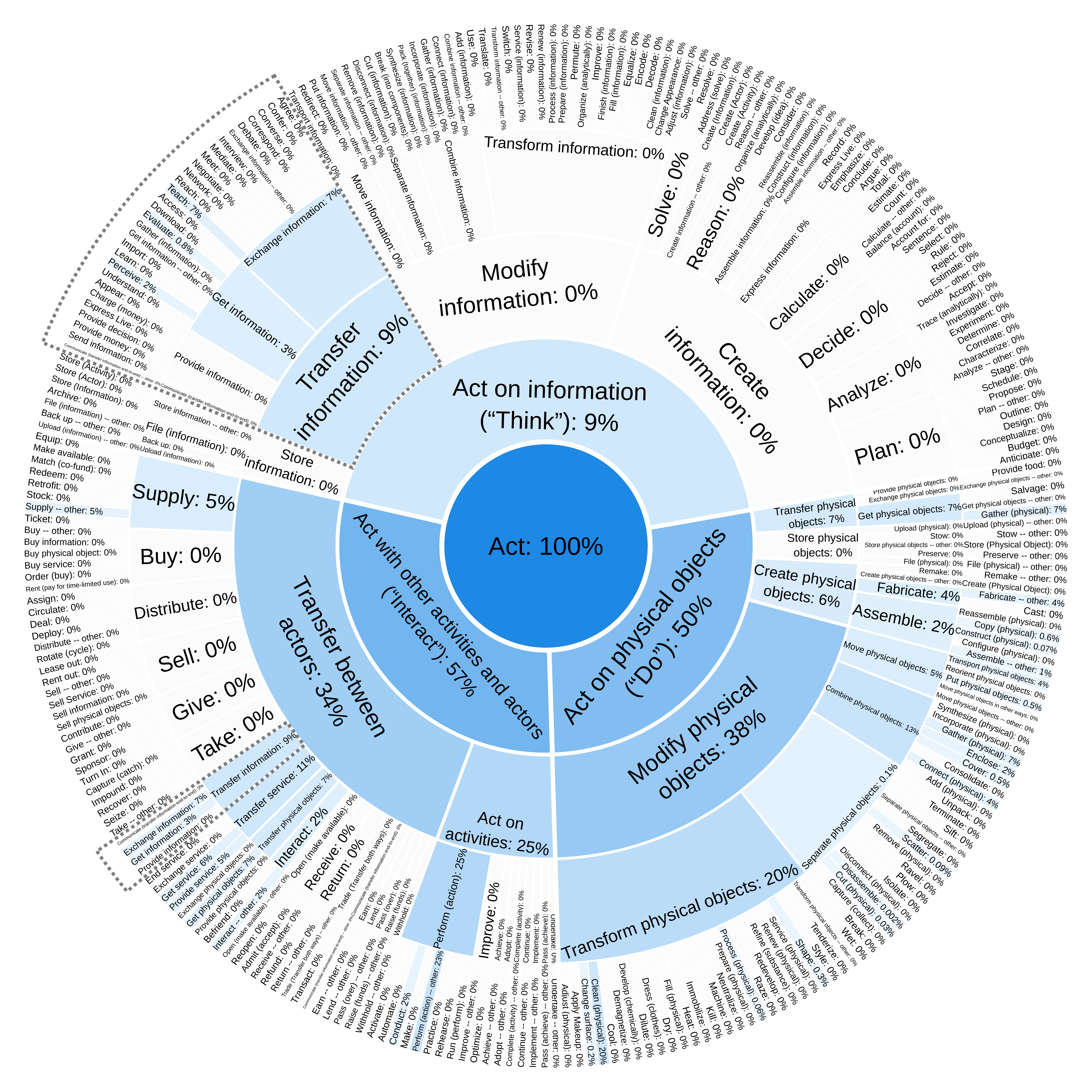}
    \caption{
        \textbf{Sunburst diagram showing the global market value of robotic systems (industrial and service robots) for the year 2024.}
    }
    \label{robot_revenue}
\end{figure}

\autoref{robot_revenue} presents the estimated market value of robotic applications at the subclass level. In contrast to the distribution of \textit{counts} of robot installations, where
activities classified under ``Do'' and
``Interact'' account for approximately 79\% and
22\% of deployed units, respectively, the corresponding distribution of the \textit{market values}
is substantially more balanced. Specifically, activities within the
``Do'' domain account for approximately 50\% of total
market value, while activities within the ``Interact'' domain contribute
approximately 57\%, rendering these two domains broadly comparable in economic terms.

A closer examination of the disaggregated data reveals a pronounced divergence between deployment volume and market value across activity subclasses. For instance, the ``Clean'' activity includes a very large number of deployed consumer robots (approximately 20 million units); however, due to the relatively low unit cost of these systems, it contributes only moderately to the total market revenue. By contrast, interactive robotic applications, particularly in the medical domain, are deployed in comparatively small numbers but generate disproportionately high revenue. This effect is driven by the high
capital cost of such systems, with prices for advanced medical robots frequently exceeding USD~2~million per unit. These findings underscore the importance of distinguishing between deployment scale and economic impact when assessing the structure and dynamics of the robotics market.

\subsection{Results on allocation of total market value to AI software applications and robotic systems}
\label{combined_percentages}

\begin{figure}[h]
    \centering
    \includegraphics[width=0.9\textwidth]{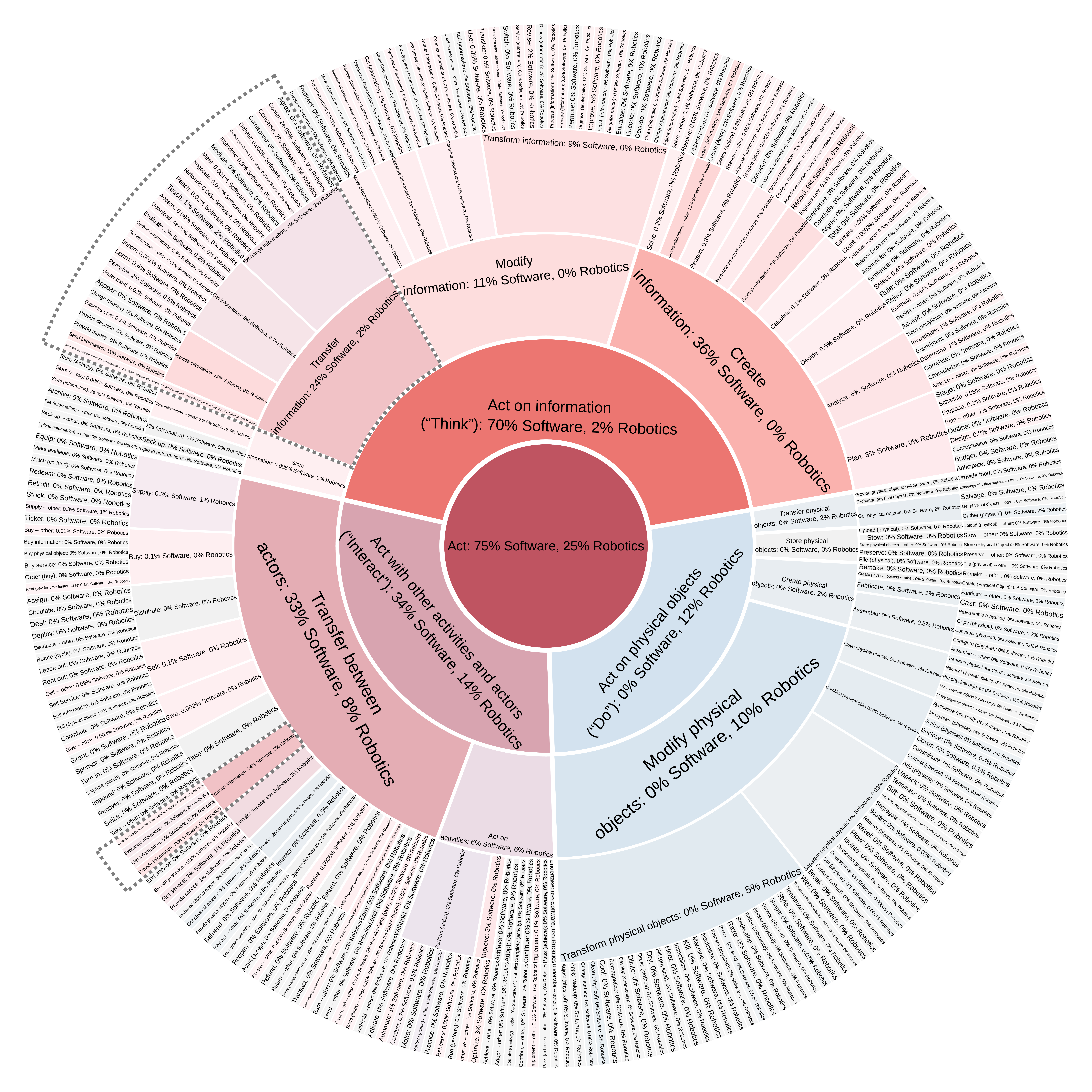}
    \caption{
        \textbf{Sunburst diagram showing the percentage contributions to market value of AI software applications and robotic systems for each work activity in 2024.} Red shading indicates activities performed by AI software applications. Blue shading indicates activities performed by robotic systems. And purple shading indicates activities performed by a combination of both.
    }
    \label{combined_revenue_percentage}
\end{figure}

\autoref{combined_revenue_percentage} shows the market value contributions of AI software applications and robotic systems for each activity in the ontology for the year 2024. Values are calculated using the methods described in previous sections of \autoref{combination}.

\clearpage




\end{appendices}


\bibliography{main}

\end{document}